\documentclass{article}
\usepackage[a4paper, total={6in, 8in}]{geometry}
\usepackage{amsmath}
\usepackage{amssymb}
\usepackage{amsthm}
\usepackage{multirow}
\usepackage{tcolorbox}
\usepackage{booktabs}
\usepackage{hyperref}
\usepackage{array}
\usepackage[all,pdf]{xy}
\usepackage{bm}
\usepackage{extarrows}
\usepackage{siunitx}
\usepackage{algorithm}%
\usepackage{algorithmicx}%
\usepackage{algpseudocode}%
\usepackage{listings}%
\usepackage{subcaption}
\usepackage{graphicx}
\graphicspath{{./}{./fig/}}
\usepackage{svg}
\usepackage{todonotes}

\usepackage{xcolor}

\newtheorem{theorem}{Theorem}

\newtheorem{remark}{Remark}

\usepackage{authblk}

\newcommand{\ourM}{UniGenX}

\title{\ourM{}: a unified generative foundation model that couples sequence, structure and function to accelerate scientific design across proteins, molecules and materials}

\author[1,2,9,*]{Gongbo Zhang}
\author[1,3,*]{Yanting Li}
\author[1,*,$\dagger$]{Renqian Luo}
\author[1,10,*,$\dagger$]{Pipi Hu}
\author[1,4,*]{Yang Yang}
\author[6]{\\Zeru Zhao}
\author[5]{Lingbo Li}
\author[1]{Guoqing Liu}
\author[1]{Zun Wang}
\author[1]{Ran Bi}
\author[6]{Kaiyuan Gao}
\author[7]{Liya Guo}
\author[1]{Yu Xie}
\author[1]{Chang Liu}
\author[1]{Jia Zhang}
\author[1]{Tian Xie}
\author[1]{Robert Pinsler}
\author[1]{Claudio Zeni}
\author[1]{Ziheng Lu}
\author[1]{Hongxia Hao}
\author[1]{Yingce Xia}
\author[1]{Marwin Segler}
\author[1]{Maik Riechert}
\author[4,6]{Wei Yang}
\author[8]{Hao Jiang}
\author[8,9]{Wen-Bin Zhang}
\author[13]{Zhijun Zeng}
\author[7,10]{Yi Zhu}
\author[11]{Li Dong}
\author[12]{Xiuyuan Hu}
\author[2,9]{Li Yuan}
\author[3]{Lei Chen}
\author[1]{Haiguang Liu}
\author[1]{Tao Qin}

\affil[1]{Microsoft Research AI for Science}
\affil[2]{School of Electronic and Computer Engineering, Peking University}
\affil[3]{DSA, The Hong Kong University of Science and Technology (Guangzhou)}
\affil[4]{School of Computer Science and Technology, Huazhong University of Science and Technology}
\affil[5]{Department of Automation, Tsinghua University}
\affil[6]{School of Artificial Intelligence and Automation, Huazhong University of Science and Technology}
\affil[7]{Yau Mathematical Sciences Center, Tsinghua University}
\affil[8]{Beijing National Laboratory for Molecular Sciences; Key Laboratory of Polymer Chemistry \& Physics of Ministry of Education; Center for Soft Matter Science and Engineering; College of Chemistry and Molecular Engineering, Peking University, Beijing}
\affil[9]{School of AI4S, Shenzhen Graduate School, Peking University, Shenzhen 518055}
\affil[10]{Beijing Institute of Mathematical Sciences and Applications, Beijing 101408}
\affil[11]{Microsoft Research Asia}
\affil[12]{Department of Electronic Engineering, Tsinghua University}
\affil[13]{Department of Mathematical Sciences, Tsinghua University} 

\affil[*]{Co-first authors in random order. Work done while Gongbo and Yanting were interns at Microsoft Research AI for Science}
\affil[$\dagger$]{Corresponding authors in random order, Email: renqianluo@gmail.com;, hpp@bimsa.cn}

\date{}
\begin{document}

\maketitle

\begin{abstract}
Function in natural systems arises from one-dimensional sequences forming three-dimensional structures with specific properties. However, current generative models suffer from critical limitations: training objectives seldom target function directly, discrete sequences and continuous coordinates are optimized in isolation, and conformational ensembles are under-modeled. We present UniGenX, a unified generative foundation model that addresses these gaps by co-generating sequences and coordinates under direct functional and property objectives across proteins, molecules, and materials. UniGenX represents heterogeneous inputs as a mixed stream of symbolic and numeric tokens, where a decoder-only autoregressive transformer provides global context and a conditional diffusion head generates numeric fields steered by task-specific tokens. Besides the new high SOTAs on structure prediction tasks, the model demonstrates state-of-the-art or competitive performance for the function-aware generation across domains: in materials, it achieves ``conflicted" multi-property conditional generation, yielding 436 crystal candidates meeting triple constraints, including 11 with novel compositions and 4 confirmed thermodynamically stable by DFT; in chemistry, it sets new benchmarks on five property targets and conformer ensemble generation on GEOM; and in biology, it improves success in modeling protein induced fit (RMSD $<$ 2 Å) by over 23-fold and enhances EC-conditioned enzyme design. Ablation studies and cross-domain transfer substantiate the benefits of joint discrete–continuous training, establishing UniGenX as a significant advance from prediction to controllable, function-aware generation.
\end{abstract}


\section{Introduction}

The ability of one-dimensional sequences to form three-dimensional structures with specific functions is a fundamental principle in natural systems. Replicating this generative process is a central goal for artificial intelligence in scientific discovery, promising to unlock transformative applications in materials design, drug discovery, and protein engineering~\cite{butler2018machine,vamathevan2019applications}. The intricate interplay between a system's discrete sequence (e.g., chemical formula, amino acid chain) and its continuous coordinates (e.g., atomic positions, protein fold) dictates its essential properties and functions. For instance, a crystal's atomic arrangement governs its electronic characteristics~\cite{hoffmann1987chemistry}, a molecule's spatial conformation determines its pharmacological efficacy~\cite{verdonk2003improved}, and a protein's 3D shape enables its biological role~\cite{branden2012introduction}. Given this sensitivity, where even minute deviations in atomic coordinates can nullify function, the challenge is not just to predict structures, but to generate them with high numerical precision under desired functional constraints.

However, current generative models face critical limitations that hinder this goal. First, most training objectives target structural accuracy in isolation, rather than the ultimate function or property of interest. Second, the generation of discrete sequences and continuous coordinates are often treated as separate tasks, failing to capture their co-dependent relationship. Finally, models typically generate a single static structure, under-modeling the dynamic conformational ensembles that are crucial for the function of many molecular and biological systems. These gaps have created a significant bottleneck, preventing a unified approach to controllable, function-aware generation across scientific domains.

The current modeling landscape is dominated by two powerful yet incomplete paradigms. On one hand, autoregressive language models (LLMs) excel at processing and generating discrete token sequences, demonstrating remarkable flexibility, scalability, and long-context reasoning~\cite{achiam2023gpt, touvron2023llama}. However, their token-based nature struggles to enforce the strict numerical precision required for generating physically plausible 3D atomic coordinates. On the other hand, diffusion models have achieved state-of-the-art performance in generating high-fidelity continuous data, such as images and 3D structures~\cite{ho2020denoising, jiao2023crystal, miller2024flowmm}. Yet, they are less adept at handling discrete sequence information and lack the flexible, token-by-token generative framework of autoregressive models, making the joint generation of sequence and structure challenging.

To address these shortcomings, we introduce UniGenX, a unified generative foundation model that co-generates sequences and coordinates under direct functional and property objectives. UniGenX bridges the gap between discrete and continuous generation by integrating an autoregressive transformer with a conditional diffusion head. In our framework, heterogeneous inputs are represented as a mixed stream of symbolic and numeric tokens. A decoder-only transformer processes this stream to capture global context, while the conditional diffusion head generates precise numeric fields (i.e., coordinates) steered by task-specific functional tokens. This hybrid architecture synergistically combines the contextual power and flexibility of autoregressive models with the numerical precision of diffusion models, enabling a single framework to operate across proteins, molecules, and materials.

UniGenX marks a significant advance from prediction to controllable, function-aware generation. Beyond setting new state-of-the-art benchmarks on structure prediction tasks, our model demonstrates unparalleled capabilities in conditional generation across domains. In materials science, it successfully navigates ``conflicted" multi-property objectives to design 436 novel crystal candidates, including 11 with new compositions and 4 confirmed as thermodynamically stable by DFT calculations. In chemistry, it establishes new performance records on five property-targeting tasks and conformer ensemble generation. For biology, it improves the success rate of modeling protein induced fit (RMSD $<$ 2 Å) by over 23-fold and enhances enzyme design conditioned on Enzyme Commission (EC) numbers. Through extensive ablation studies and cross-domain transfer experiments, we substantiate the benefits of our joint discrete–continuous training strategy, establishing UniGenX as a versatile and powerful foundation for AI-driven scientific discovery.

\section{Methods}
\ourM{} leverages the strengths of both autoregressive (AR) and diffusion models, achieving a natural unification. The diffusion head addresses the numerical precision limitations of general AR models by operating in a continuous vector space for numerical data. Conversely, the AR next-token prediction provides effective conditioning for the diffusion head, which diffuses only a low-dimensional variable (e.g., three dimensions for molecular 3D structures), significantly simplifying the diffusion training process. Our design seamlessly integrates symbolic (words) and numerical (numbers) data across scientific domains without sacrificing scientific precision. This section details the key features of the \ourM{} architecture.

\subsection{Sequentialize All for Multi-Domain/Task Compatibility}

\begin{figure}[htbp]
    \centering
    \includegraphics[width=1.0\linewidth]{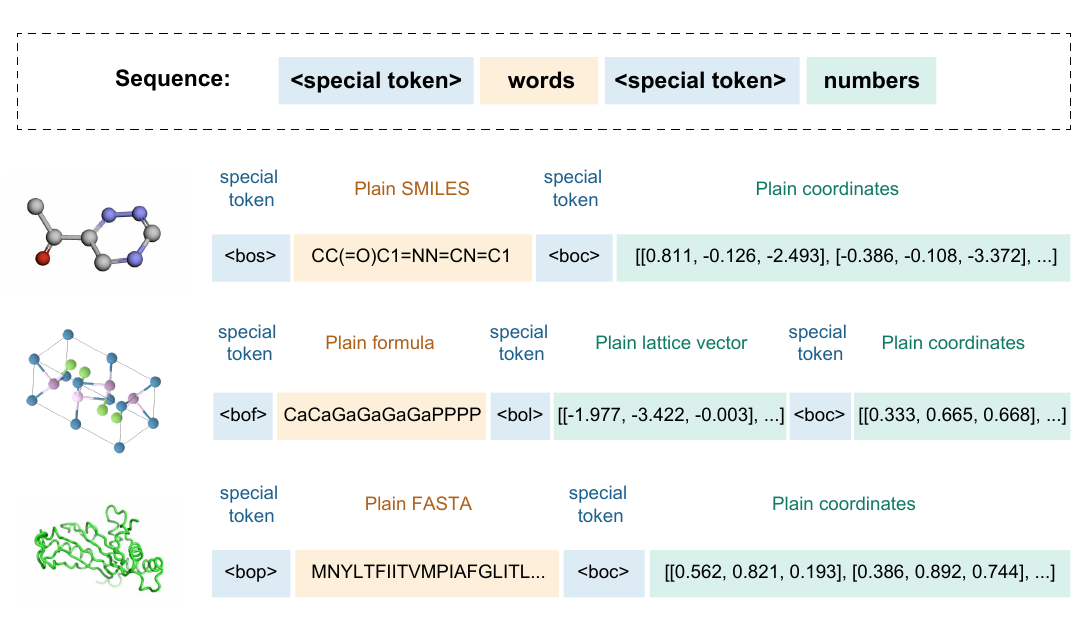}
    \caption{Sequentialization approach combines symbolic (words) and numerical data (numbers), separated by special tokens that identify the category of each data element. As illustrated in the figure for small molecules and materials, SMILES strings or chemical formulas are treated as words and delimited by special tokens such as $<$bos$>$ (beginning of SMILES) or $<$bof$>$ (beginning of formula). Corresponding coordinates are treated as numbers, delimited by tokens such as $<$boc$>$ (beginning of coordinates). This method provides a general framework applicable to diverse scientific data formats.}
    \label{fig:sequence}
\end{figure}

The success of multimodal GPT-like models in integrating language and images demonstrates the potential of sequence alignment for addressing diverse problems across various domains. To achieve alignment with language—a naturally sequential modality—images are encoded into a latent space using a pretrained encoder and subsequently aligned with language embeddings within this shared latent space. This process enables general comprehension and facilitates generation in both language and image modalities. This approach highlights the flexibility of sequentialization across diverse domains and tasks, a principle validated by previous research and applications of GPT models.

However, scientific data presents distinct challenges in sequentialization compared to data from general domains like language, images, and videos. Scientific data exhibits significantly greater complexity and diverse standards across different disciplines. For instance, small molecules are frequently represented using SMILES strings, encoding atoms and chemical bonds~\cite{toropov2005simplified}, while periodic materials are often described by chemical formulas~\cite{grosso2013solid},  accompanied by atomic coordinates. Proteins, DNA, and data types like energies and forces also utilize specialized, compact representations. This inherent domain specificity and the deep integration of expert knowledge within these representations make establishing a unified standard exceptionally challenging. Consequently, developing a universal representation capable of aligning structures, energies, forces, and other numerical properties with language-like atomic or molecular descriptions is exceedingly difficult. Furthermore, scientific data inherently contains a large amount of numerical data that is highly sensitive to numerical precision. The aforementioned challenges preclude the direct transfer of sequentialization methods from general domains to scientific applications.

To fully leverage the flexibility of multimodality, we propose a simple yet effective sequentialization approach: representing all data—formulas, coordinates, energies, forces, and so on—as a single sequence segmented by special tokens. This sequentialization enables framing all generation tasks as next-token prediction problems, thereby harnessing the scalability and flexibility inherent in autoregressive models for scientific applications. And we will show that this approach can handle various tasks across different scientific domains.

Figure~\ref{fig:sequence} depicts our sequentialization scheme, which unifies symbolic (words) and numerical data (numbers) into a single sequence delineated by special tokens. For small molecules, the $<$bos$>$ token marks the onset of the SMILES string, while $<$boc$>$ signals the commencement of its coordinate sequence. A parallel approach is applied to materials, with $<$bof$>$ initiating the chemical formula and $<$boc$>$ the coordinate sequence. The $<$eos$>$ token is appended to signify the sequence's termination. This methodology facilitates the encoding of diverse scientific data into a unified sequence format and is adaptable to other data types, including protein and DNA sequences. The explanation of the special tokens is provided in the Table~\ref{tab:special_tokens}.

This sequentialization method offers a simple and effective way to represent diverse scientific data across various domains and tasks. However, traditional sequence models struggle to handle numerical data (numbers) with the same efficacy as symbolic data (words), particularly when numerical precision is crucial. A diffusion strategy combined with a sequence model effectively handles both numerical and symbolic data within a sequence modeling framework. This approach will be detailed in the following subsection.

\subsection{Bridge the Gap: Unified Sequence Modeling of Words and Numbers}
\begin{figure}[htbp]
    \centering
    \includegraphics[width=0.9\linewidth]{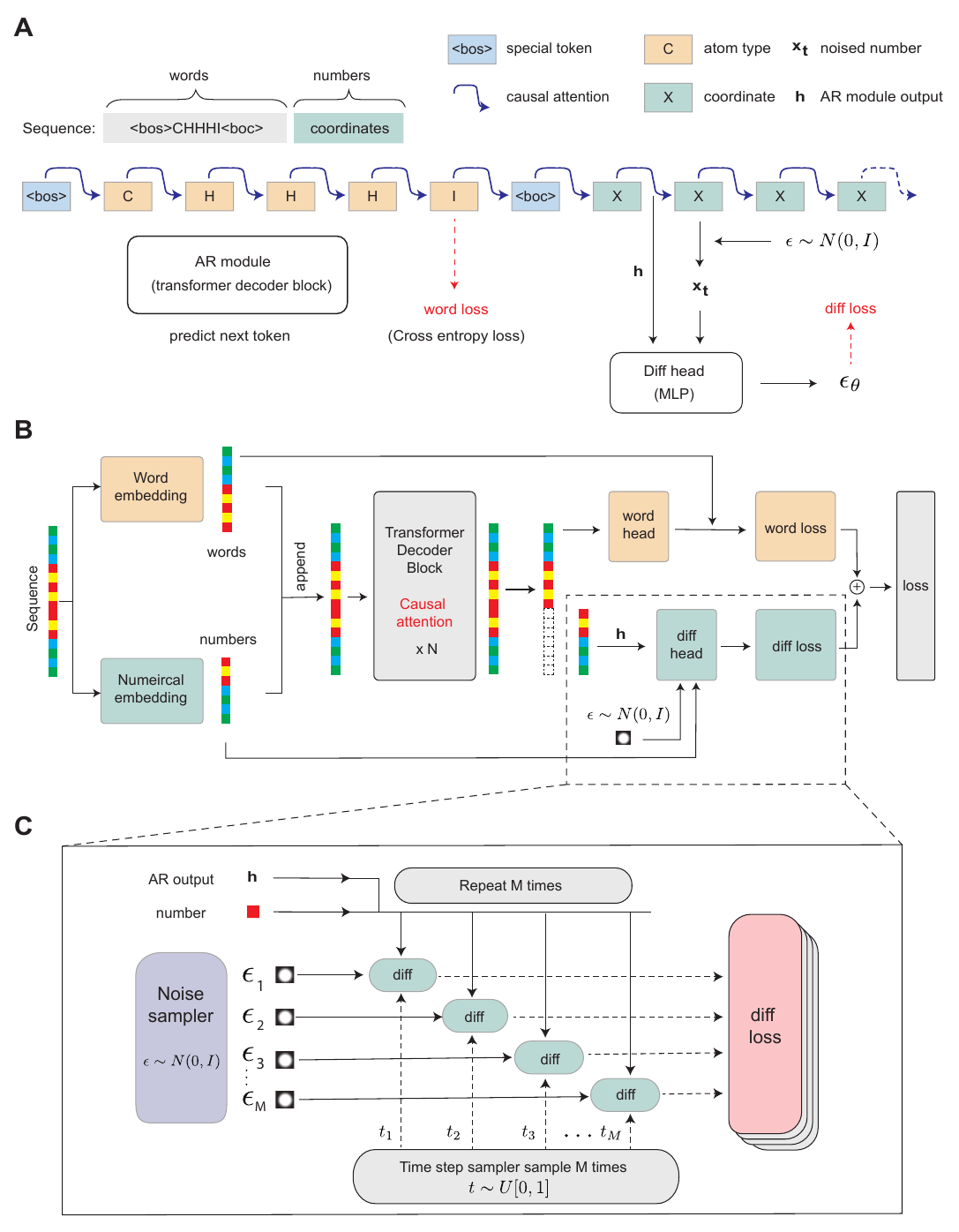}
    \caption{The scheme of the \ourM{} framework for a unified generation. We present a novel autoregressive diffusion framework for science generation, surpassing traditional LLMs and structural prediction models. By combining next-token prediction and conditional diffusion, our approach offers enhanced flexibility, overcomes the difficulties of training of the traditional jointly diffusion, and the numerical accuracy issues that have limited the application of LLMs in high-precision scientific fields.}
    \label{fig:scheme}
\end{figure}


The key principle of this design is ``word-to-word, number-to-number'' prediction. Specifically, a causal attention transformer decoder (AR module) performs next-token prediction. If the predicted token is a word, a standard cross-entropy loss is computed. In contrast, if the predicted token represents a number, the diffusion head is activated to predict the diffusion target conditioned on the output of AR module, and a diffusion loss is calculated. To improve training efficiency, losses from several diffusion time steps are aggregated and combined with the word loss to jointly train the transformer decoder and the diffusion head. This process is illustrated in Figure~\ref{fig:scheme}(a), where the sequence ``$\langle$bos$\rangle$CHHHI$\langle$boc$\rangle$coordinates" (a mixture of words and numbers) is fed into the AR module for next-token prediction using causal attention. Finally, the total loss, computed as the sum of the word loss and the aggregated diffusion loss, is used to jointly train the AR module and the diffusion head (Figure~\ref{fig:scheme}(b)). To fully leverage the conditioning $h$ from the autoregressive next-token prediction, we sample multiple diffusion time steps $t_1, t_2, \cdots, t_M$ for the diffusion head and sum the corresponding $M$ diffusion losses to obtain the final diffusion loss (Figure~\ref{fig:scheme}(c)).

This design treats words and numbers equally within the autoregressive framework but employs distinct heads to model the discrete and continuous spaces, respectively. The majority of parameters reside within the autoregressive module, responsible for next-token prediction, while the diffusion head is significantly lighter. This is because the autoregressive module captures most of the contextual information in the predicted token, leaving only a low-dimensional variable for the diffusion process. This approach differs substantially from traditional joint diffusion training and consequently requires far fewer parameters and time steps for training the diffusion head. The diffusion head operates on a single variable at a time, for example, the three-dimensional coordinates or forces of a specific atom. Consequently, the diffusion training is agnostic to the semantic meaning of the variable; whether it represents coordinates, energy, or force is irrelevant to the diffusion process itself. Therefore, a single diffusion head can be used for training all numerical data without loss of precision. Therefore, our model exhibits significant flexibility across diverse domains and tasks that can be represented as sequences of words and numbers.

Our model does not incorporate explicit inductive biases for equivariance or invariance. This design choice grants significant flexibility in handling diverse tasks and scientific domains with varying symmetry requirements. We demonstrate through our results that learning specific equivariant or invariant properties can be effectively achieved through data augmentation, thus preserving the scalability of the underlying transformer architecture.

\begin{remark}\textbf{Benefits of Unified training of Sequence and Structure.}
Beyond the advantages for generation capabilities derived from the unified generation of sequence and structure, we emphasize the training perspective. Unified training allows the model to more comprehensively capture the inherent knowledge within scientific data. To illustrate, let $s$ denote the sequence, $x$ the coordinates of the structure, and $f_\theta$ the model with training parameters $\theta$. The learned mappings for different models are as follows:
$$
\begin{cases}
    f_\theta^{\text{diffusion}}: s \mapsto x, & \text{Sequence-to-structure diffusion models,} \\
    f_\theta^{\text{sequence}}: s \mapsto s, & \text{Sequence-to-sequence autoregressive models,} \\
    f_\theta^{\text{\ourM{}}}: (s, x) \mapsto (s, x), & \text{\ourM{} model.}
\end{cases}
$$

The joint mapping $f_\theta^{\text{\ourM{}}}$ from $(s, x)$ to $(s, x)$ significantly enhances knowledge distillation from scientific data compared to the other two mappings. In scientific domains, the properties or functions of a molecule are determined by both its sequence and structural coordinates, which are richly interconnected. The joint training strategy enables \ourM{} to uncover these interdependencies, leading to improved performance in understanding and generation tasks. Consider the co-evolution of amino acids in a protein for example: two amino acids, though not adjacent in the sequence, may be spatially close in the 3D structure. Consequently, a mutation in one amino acid can induce a related mutation in the other. A sequence-only model would entirely miss this information, and a sequence-to-structure model, while capturing some information from extensive sequence-structure data, lacks explicit sequence modeling due to the absence of a sequence-based loss.
\end{remark}

\section{Experiments}
To evaluate \ourM{}, we conduct experiments across various tasks in the materials and small molecule domains. In this work, we present two model configurations of different sizes: \ourM{}(100M) and \ourM{}(400M), where the transformer decoder backbones contain 100M and 400M parameters, respectively. See Table~\ref{tab:llama_configurations} for detailed model configurations. In general, \ourM{}(400M) outperfroms \ourM{}(100M) reflecting scaling law in this model architecture.

\subsection{Material}
For material domain, we evaluate \ourM{} on crystal structure prediction and de novo generation task.

\subsubsection{Crystal Structure Prediction}
In this task, we performed finetuning and evaluations on three mainstream benchmarks.
The MP-20 dataset is a subset of the Materials Project~\cite{Jain2013}. 
It encompasses all materials with a unit cell atomic count less than 20, and most properties can achieve DFT accuracy. 
Carbon-24~\cite{carbon2020data} contains 10k carbon materials, which share the same composition, but have different structures. 
Similar to the MP-20 dataset, it covers materials composed only of carbon atoms with a unit cell atomic count less than 24. 
The MPTS-52 dataset includes time-based splits for cross-validation and benchmarking generative models, allowing up to 52 atoms in the unit cells.

The performance is compared with three baselines in Table~\ref{tab:csp baseline} which are all previous state-of-the-art methods. Notably, In contrast to them, our model does not incorporate explicit inductive biases for equivariance or invariance, we demonstrate through our results that these properties can be effectively learned through data augmentation, preserving the scalability of the underlying transformer architecture and providing flexibility across diverse tasks and scientific domains with varying symmetry requirements.
 CDVAE~\cite{xie2021crystal} is modified in~\cite{jiao2023crystal} to perform the CSP task by replacing its original normal prior for generation with a parametric prior conditioned on the encoding of the specified composition.
\begin{table}[h]
\centering
\begin{tabular}{|l|l|}
\hline
\textbf{Method} & \textbf{Model Type} \\
\hline
CDVAE\cite{xie2021crystal}(modified by \cite{jiao2023crystal}) & Variational Autoencoder (VAE)\\
\hline
DiffCSP\cite{jiao2023crystal} & Diffusion Model \\
\hline
FlowMM\cite{miller2024flowmm} & Riemannian Flow Matching \\
\hline
\end{tabular}
\caption{Comparison of generative models for crystal structure prediction.}
\label{tab:csp baseline}
\end{table}

We pretrained our model on a comprehensive dataset that we constructed, which includes the training sets of MP-20, Carbon-24, MPTS-52 ,and a subset of data from NOMAD, which are disjoint from all test sets. Subsequently, we finetuned the pretrained model on MP-20, Carbon-24, and MPTS-52 respectively. For evaluation, following \cite{xie2021crystal}, we employed \texttt{StructureMatcher}~\cite{ong2013python} to compute metrics including the Match Rate and the Root-Mean-Square Derivation (RMSD) with thresholds stol$=0.5\textrm{\r{A}}$, angle\_tol$=10^\circ$, ltol=$0.3\textrm{\r{A}}$. The Match Rate represents the percentage of matched structures within the test set. The RMSD is computed between the ground truth and the matched structure. 

\subsubsection{De Novo Generation}

As a unified model, \ourM{} inherently possesses the capability of de novo generation. 
Following \cite{miller2024flowmm}, we let the model finetuned on MP-20 generate 10,000 materials. Different from CSP task, It start with only begin token $\texttt{<bos>}$ and generating the compositions and structures through sampling and diffusion. 

We assess the generation performance using three metrics: Validity, Coverage, and Property Statistics. Validity measures the correctness of the predicted crystals. Coverage evaluates the similarity between the test set and the generated samples. Property Statistics calculates properties such as density and the number of elements.

\subsubsection{Conditional Generation}

As an autoregressive model, one of its key features is the ability to perform controllable conditional generation with ease. Therefore, we evaluate \ourM{} on conditional generation tasks, where the goal is to generate stable crystal structures that satisfy specific target conditions. This setting reflects real-world design scenarios in which materials are sought with desired physical or chemical properties.

To enable conditional generation, we prepend a special token to the input sequence, which includes both the material composition and structure, to indicate the target property type. Sequentially, we embed the target property value using a multilayer perceptron (MLP) and append it to the sequence, in front of the material's composition and structure. We finetune our pretrained model using datasets with property labels. We focus on scalar property conditioning, targeting single-valued material properties such as magnetic density and bulk modulus. For each property, we train a separate model using a small-to-moderate-sized labeled dataset.

During training, we apply the loss only on the material itself, without directly supervising the property value. To better handle the numerical range of the target properties, we apply a nonlinear transformation of the form $\mathrm{sign}(x) \cdot \log(|x| + 1)$ to the property values before embedding.

To demonstrate our model's ability to generate out-of-distribution (OOD) samples, following \cite{zeni2025generative}, we consider two scalar property targets:
\begin{itemize}
\item \textbf{Magnetic Density:} We finetune on 600k structures labeled with magnetic density (in $\text{\AA}^{-3}$), and generate samples targeting values equal to 0.20~$\text{\AA}^{-3}$. 
\item \textbf{Bulk Modulus:} We finetune on a small dataset of 5k samples, targeting high-stiffness materials with a bulk modulus equal to 400~GPa. 
\end{itemize}

During inference, inspired by \cite{karras2024guiding}, we employed an earlier checkpoint of our own model—trained for fewer steps—to guide the generation process of the final model. This approach serves as a substitute for the standard classifier-free guidance (CFG), effectively enabling a form of “temperature reduction” during sampling. For the generated materials, property values were evaluated using the MatterSim \cite{yang2024mattersim}.

\subsubsection{Multi-conditional Generation}

We further evaluated \ourM{} under multi-conditional generation, where the target is to generate materials satisfying multiple properties simultaneously. This is a classic materials design problem involving trade-offs between multiple physical properties. The core significance of this task lies in the fact that these three properties are typically in conflict with one another, making simultaneous optimization highly challenging—providing an ideal benchmark to evaluate the capability of a material generation model. Ultimately, these materials can potentially provide implications in many high-value engineering applications such as aerospace materials, thermal energy storage and management materials, and protective materials etc. 

In this setting, we prepend three special tokens to the input sequence, each indicating a specific property type: heat capacity at 300 K (J/g·K), Hill's elastic modulus(GPa), and density (kg/m³), followed by their corresponding signed-log-transformed values.

Training and inference procedures remain consistent with the single-condition setup, including the use of classifier-free guidance during sampling.

The multi-labeled dataset is constructed from the Materials Project containing no more than 16 atoms, and employed the pre-trained MatterSim-v1.0.0-5M machine-learning interatomic potential via the ASE and Phonopy interfaces to compute, under the harmonic approximation, each crystal’s Hill's elastic modulus, its thermal conductivity at 300 K, and its density—these quantities serving as the training labels. For each data item containing three distinct labels, we partition it into $C_3^1 + C_3^2 + C_3^3$ subsets, where each subset contains 1, 2, or 3 labels, respectively. These subsets are then used to fine-tune our model. This strategy is designed to enable \ourM{} to flexibly generate structures under varying numbers of conditioning labels, thereby enhancing its versatility in both under- and fully-conditioned generation scenarios.

As a representative multi-conditional generation task, we aim to generate crystal structures that satisfy the following target properties: density of $2.5\pm0.2kg/m^3$, Hill's elastic modulus $E > 200$GPa, and specific heat capacity $C_v > 1$J/g·K.

\subsection{Molecule}
For small molecule domain, we evaluate \ourM{} on conformation generation and conditional generation tasks.

\subsubsection{Conformation Generation}
Adopting the data splits defined by DMCG~\cite{zhu2022direct}, we conducted training and evaluation on the large-scale GEOM-QM9 and GEOM-Drugs datasets. The GEOM dataset~\cite{axelrod2022geom} comprises three-dimensional conformations for over 37 million molecules, each annotated with experimental data and high-precision Density Functional Theory (DFT) energy values. GEOM-QM9 and GEOM-Drugs are subsets of this comprehensive dataset. Notably, GEOM-QM9 offers a richer conformational diversity compared to the traditional QM9 dataset~\cite{ramakrishnan2014quantum}, incorporating intermediate states in addition to ground states, which is advantageous for downstream tasks such as molecular and protein-molecule docking.

Among the comparative methods, GEOMOL~\cite{ganea2021geomol} employs message passing neural networks (MPNNs) and SE(3)-invariance to predict local atomic 3D structures and torsion angles, facilitating deterministic assembly of complete conformers. ConfGF~\cite{shi2021learning} is a gradient-based model that optimizes molecular 3D structures by learning interatomic distance gradient fields and utilizing score matching for training. DMCG~\cite{zhu2022direct}, the current state-of-the-art, directly predicts atomic 3D coordinates through a rotation, translation, and permutation-invariant loss function and an iterative refinement architecture.

For evaluation, given molecule $x$ with $N_x$ conformations in the test set, we generate $2N_x$ conformations, following \cite{shi2021learning}. Let $\mathbb{S}_g$ and $\mathbb{S}_r$ represent the sets of generated and ground truth conformations, respectively. Conformation deviation is quantified using $\texttt{GetBestRMS}$ from the $\texttt{RDKit}$ package, with root-mean-square deviation denoted as $\mathrm{RMSD}(R, \hat{R})$. The recall-based coverage (COV) and matching (MAT) scores are defined as follows:

\begin{equation}
	     \mathrm{COV}(\mathbb{S}_g ,\mathbb{S}_r) = \frac{1}{\vert \mathbb{S}_r \vert} \begin{vmatrix} \begin{Bmatrix} R \in \mathbb{S}_r \vert \mathrm{RMSD}(R, \hat{R}) < \delta, \exists \hat{R} \in \mathbb{S}_g \end{Bmatrix} \end{vmatrix}
\label{cov}
\end{equation}

\begin{equation}
    \mathrm{MAT}(\mathbb{S}_g ,\mathbb{S}_r) = \frac{1}{\vert \mathbb{S}_r \vert} \sum_{R \in \mathbb{S}_r} {min}_{\hat{R} \in \mathbb{S}_g} \mathrm{RMSD}(R, \hat{R})
\label{mat}
\end{equation}

An effective method should exhibit a high coverage (COV) score and a low matching (MAT) score. Following \cite{shi2021learning,xu2023prediction}, the thresholds ($\delta$) are set to $0.5\textrm{\r{A}}$ and $1.25\textrm{\r{A}}$ for GEOM-QM9 and GEOM-Drugs, respectively. Precision-based COV and MAT scores are also computed by interchanging $\mathbb{S}_g$ and $\mathbb{S}_r$ in equations \ref{cov} and \ref{mat}. The precision-based coverage and matching scores are defined as follows:

\begin{equation}
	     \mathrm{COV}\text{-}\mathrm{P}(\mathbb{S}_g ,\mathbb{S}_r) = \frac{1}{\vert \mathbb{S}_g \vert} \begin{vmatrix} \begin{Bmatrix} {\hat{R} \in \mathbb{S}_g} \vert \mathrm{RMSD}(R, \hat{R}) < \delta, \exists R \in \mathbb{S}_r \end{Bmatrix} \end{vmatrix}
\label{cov-p}
\end{equation}

\begin{equation}
    \mathrm{MAT}\text{-}\mathrm{P}(\mathbb{S}_g ,\mathbb{S}_r) = \frac{1}{\vert \mathbb{S}_g \vert} \sum_{{\hat{R} \in \mathbb{S}_g}} {min}_{R \in \mathbb{S}_r} \mathrm{RMSD}(R, \hat{R})
\label{mat-p}
\end{equation}

\subsubsection{Property Evaluation}
We extended our evaluation to include molecular property prediction alongside conformation generation, a task that involves predicting properties from generated conformations~\cite{axelrod2022geom}. Following \cite{zhu2022direct}, we randomly selected 30 molecular SMILES from the test set of the GEOM-QM9 dataset. For each molecule, we generated 50 distinct conformations. Using the Psi4 quantum chemistry software package~\cite{smith2020psi4}, we calculated energy levels, HOMO, and LUMO characteristics for both generated and reference conformations. We then derived ensemble averages for energy (\(\overline{E}\)), minimum energy (\(E_{\text{min}}\)), average HOMO-LUMO gap (\(\Delta \overline\epsilon\)), and the gap extremities (\(\Delta \epsilon_{\text{min}}\) and \(\Delta \epsilon_{\text{max}}\)) based on the conformational attributes of each molecule. Mean absolute error (MAE) was used to quantify property discrepancies between generated and reference conformations. Given our model's superior performance in predicting these statistical quantities, depicted in Sec.~\ref{sec:molecule}, such as average energy and average HOMO-LUMO gap, we further compared the distribution of sampled results from \ourM{} with the ground-truth conformations.

\subsubsection{Conditional Generation}

We further explored our model's conditional generation capabilities. Instead of employing guidance in the diffusion model, \ourM{} utilizes a straightforward approach: prepending the training data sequence with the condition property and its value, marked by a special token like $<$bobulk$>$ for the bulk property. This enables simple prompting at inference, where conditional generation is achieved by providing the property-value pair and the sequence generation special token. Data processing examples are detailed in Appendix~\ref{app:cond}. In this section, using conditional molecule generation on the QM9 benchmark as an example, we demonstrate \ourM{}'s strong performance and note its easy extensibility to different domains.

Following \cite{hoogeboom2022equivariant}, the QM9 dataset is divided into two halves, the first half is used to train the classifier, while the second half is used to train our model. For the classifier, we use the training weights provided by \cite{you2024latent} to achieve a fair comparison. Unlike other diffusion models, we wrap the condition into the sequence. Only one special token is needed for wrapping. Refer to Appendix for more details.

The properties include volume $\alpha(Bohr^3)$, the ability of a molecule to become polarized under an external electric field; HOMO $\varepsilon_{H}(meV)$, the energy level of the highest occupied molecular orbital; LUMO $\varepsilon_{L}(meV)$, the lowest unoccupied molecular orbital energy; energy gap $\Delta\varepsilon(meV)$, the energy difference between HOMO and LUMO; dipole moment $\mu(D)$, the separation of positive and negative charges within the molecule.

For evaluation, we employed two different sampling methods. The first, introduced in \cite{hoogeboom2022equivariant}, models the distribution of conditions in the training set based on equation~\ref{edm}. It first samples the number of atoms, $N$ , and then samples the property values corresponding to $N$. The second method, proposed by \cite{you2024latent}, addresses a potential inductive bias in the original method from \cite{hoogeboom2022equivariant}, where an implicit correlation between molecule size and properties may arise during evaluation. To mitigate this, they propose uniformly sampling property values directly between the minimum and maximum values, as described in equation~\ref{ldm}.

We evaluated our model's performance aligning both sampling methods. Notably, unlike prior approaches that trained separate models for each property and relied on extensive preprocessing, our model is trained as a unified, all-in-one model and not require pre-generation of the number of atoms, N.

\begin{equation}
    p(\mathcal{M} | \mathbf{x}_{\text{cond}}) \propto \sum_N p(\mathcal{M}_N | N, \mathbf{x}_{\text{cond}}) p(N)
\label{ldm}
\end{equation}

\begin{equation}
    p(\mathcal{M} | \mathbf{x}_{\text{cond}}) \propto \sum_N p(\mathcal{M}_N | N, \mathbf{x}_{\text{cond}}) p(N) p(\mathbf{x}_{\text{cond}} | N)
\label{edm}
\end{equation}

\subsubsection{Multi-conditional Generation}
In the domain of small molecule conditional generation, another crucial task is multi-condition generation. Since our model is a sequence model, it easily accommodates multiple condition properties by appending them to the beginning of the sequence for conditional generation.

We have observed that previous works typically focus on single-condition generation and train models using only one condition. Here, we introduce a multi-condition generation setting. Consistent with \cite{hoogeboom2022equivariant}, we divide the QM9 dataset equally into two halves, with the first half used to train the classifier and the second half to train our model. The QM9 dataset contains six comparable properties, so we select $n$ properties (where $1 \leq n \leq 6$) for splitting the data. This results in a total of $\binom{6}{1} + \binom{6}{2} + \binom{6}{3} + \binom{6}{4} + \binom{6}{5} + \binom{6}{6} = 63 $ data splits. We continue to use the training weights provided by \cite{you2024latent} for our classifier. Similar to single-condition generation, we embed conditions into the sequence, requiring only one special token per condition for embedding. For multi-condition generation, we determine the order of conditions based on the sorted string representation of them.

During the evaluation phase, we randomly selected 1k values from any of the 63 condition combinations in the divided test set for generation, and subsequently assessed them using a classifier. We believe that calculating the Mean Absolute Error (MAE) for each property in multi-condition generation lacks significant meaning. The primary focus is for our generator to produce distributions that meet real-world requirements. Therefore, for different condition combinations, we plotted the distribution of both the actual data and the generated data.

\subsection{Protein}
We pre-trained a 400M-parameter model using approximately 78 million protein sequence–structure pairs from the AFDB. These proteins were selected based on sequence identity below 90\% and pLDDT scores above 70. During training, sequences longer than 256 residues were randomly truncated, enabling the model to effectively learn the relationship between protein sequences and their 3D structures, and to support downstream tasks.

\subsubsection{Emulating MD equilibrium distributions}
We evaluated \ourM{}'s ability to model long-timescale molecular dynamics (MD) equilibrium distributions using simulations of 12 fast-folding proteins generated by \cite{lindorff2011fast} on the special-purpose supercomputer Anton.

From the DESRES fast-folding MD trajectories, we sampled one snapshot every 10 frames to construct the dataset. For each snapshot, we extracted the Cartesian coordinates of the alpha carbon atoms (C$\alpha$) as a representation of the amino acid positions, along with the corresponding one-letter amino acid sequence. To mitigate overfitting, we incorporated a 5\% subset of the AFDB training data containing sequences shorter than 80 residues into the training set.

We then trained 12 “DESRES-finetuned models,” each fine-tuned on 11 fast folders using our base model pretrained on AFDB data, and tested on the remaining held-out fast folder. Across these 12 tasks, our model successfully recovered native and unfolded states that are structurally consistent with the expected free energy landscapes of the proteins.

For evaluation, we employed Time-lagged Independent Component Analysis (TICA) to identify low-dimensional projections associated with the system's slowest dynamical processes. Using pairwise distance matrices of protein structures as input, TICA extracts directions corresponding to the most kinetically relevant degrees of freedom, which effectively serve as latent axes of the free energy landscape. We then projected the trajectories onto the first two TICA components to visualize the system in a 2D plane, where each point represents a protein conformation. High-density regions in this space correspond to low free energy (i.e., highly probable) states, while sparse regions indicate high free energy (i.e., less probable) states. This analysis enables us to assess whether \ourM{} has successfully learned to reproduce the true long-timescale MD equilibrium distributions.

\subsubsection{EC Number-guided Protein Generation}
The Enzyme Commission (EC) number is a numerical classification scheme for enzymes based on the chemical reactions they catalyze. Each EC number consists of four hierarchical levels (e.g., 2.6.1.1), providing increasingly specific information about enzyme function. In our work, we use only the first three levels (e.g., 2.6.1) as generation conditions, which is sufficient to define the primary functional class of the target enzyme while maintaining generalization.

Our approach to EC number-guided protein generation follows the same design as the molecular and material property-conditional generation described earlier. Specifically, the EC number is converted into a sequence of tokens and special tokens, which are prepended to the input sequence. For example, an EC number 2.6.1.x is encoded as \texttt{<ec1>2<ec2>6<ec3>1<prot>} at the beginning of the input sequence and structure tokens.

For training data, we used all protein sequences in UniProt that have an assigned EC number and corresponding structural information available in the AlphaFold Database (AFDB), and with sequence lengths below 512 residues. This results in a dataset of approximately 20 million samples.

To evaluate the effectiveness of EC number conditioning, we selected three representative EC numbers and generated 400 protein sequences for each. The validity of the generated sequences was first checked using CLEAN \cite{yu2023enzyme} to ensure they conformed to the assigned EC number. To further assess the structural quality and functional relevance, we predicted the 3D structures of the generated sequences using ESMFold \cite{lin2023evolutionary}, performed sequence alignment against the UniProt database using BLAST, and conducted structural comparisons against AFDB and PDB using Foldseek \cite{van2022foldseek}. Additionally, to evaluate the contribution of structure information during training, we trained a sequence-only version of the model as a baseline for comparison.

\subsection{Protein-Ligand Docking}
A central challenge in structural biology is the accurate prediction of small-molecule ligands that bind to a given protein pocket\cite{hu20253dmolformer}. Most existing approaches rely on static, co-crystallized protein-ligand complexes, failing to account for the conformational flexibility that proteins often exhibit upon ligand binding. In this work, we formulate two related tasks that explicitly incorporate protein dynamics into the modeling process.

The first task involves predicting the bound ligand conformation given both the unbound (apo) and bound (holo) pocket structures. The second, more challenging task requires predicting both the holo pocket conformation and the bound ligand coordinates from the apo structure alone. Together, these tasks aim to model the dynamic nature of molecular recognition and capture induced-fit effects that are often neglected in traditional structure-based docking pipelines. To this end, we leverage the MISATO dataset\cite{siebenmorgen2024misato}, which comprises approximately 20,000 experimentally resolved protein-ligand complexes along with over 170 microseconds of molecular dynamics (MD) trajectories—offering a unique view into the conformational transitions involved in binding events.

To construct inputs for these tasks, we use AlphaFold2-predicted structures\cite{jumper2021highly} as proxies for the unbound (apo) state of proteins. Because the MISATO dataset does not retain original chain identifiers and includes preprocessing steps such as hydrogen addition, we use the PDBbind dataset\cite{liu2015pdb} as a reference. We first align each MISATO complex to its corresponding entry in PDBbind, then align the relevant chains to their AlphaFold2-predicted structures from the AlphaFold Protein Structure Database (AFDB) using UniProt identifiers. This two-step alignment allows us to establish residue-level correspondences between experimental and predicted structures. Binding pockets are subsequently extracted using the MISATO-provided script, which selects all residues within 10 Å of the ligand.

Following this preprocessing pipeline, we obtain 9,744 protein-ligand complexes, each with 100 MD snapshots. We adopt the CASF-2016 benchmark split\cite{su2018comparative} to partition the data into 9,587 training samples and 157 test samples. Each small molecule is represented using its canonical SMILES string after hydrogen removal, while protein pockets are represented at the all-atom level.

Each test sample is evaluated using 2$N_f$ = 200 generated conformational frames. The two prediction tasks are assessed using distinct evaluation protocols. In the ligand-only prediction task, where both apo and holo pocket structures are provided as input, performance is measured by computing the root-mean-square deviation (RMSD) between the predicted ligand coordinates and the ground truth ligand in the corresponding MD frame. In the joint prediction task, which requires generating both the holo pocket and the ligand from the apo structure alone, evaluation is conducted separately for the ligand, the pocket, and the full protein-ligand complex. For each of these components, RMSD is computed against all MD frames corresponding to the same PDB ID, and the minimum RMSD across frames is reported. This flexible evaluation accounts for the conformational ambiguity inherent in dynamic binding processes, recognizing that multiple structurally distinct solutions may be equally valid.

\subsection{Unification training on Material and Molecule}
We demonstrated our model's ability to perform unified training and generation across diverse scientific domains. Specifically, we trained a single model on both molecule and material datasets. As detailed in the methods, \ourM{} not only accommodates pure sequences, such as crystal formulas in materials or SMILES strings in small molecules, but also effectively handles numerical tokens, such as atomic coordinates, through its diffusion heads. This showcases \ourM{}'s potential for unifying various scientific domains by training a single model across different tasks and datasets.

To differentiate domains, we introduced special tokens for materials and molecules as indicators during training. We trained a 100M parameter model on a combined dataset of 5M material data and 1M QM9 molecule data, denoted as \ourM{}(100M). We then finetuned this model on material datasets including MP-20, Carbon-24, and MPTS-52, as well as the 1M GEOM-QM9 molecule dataset.

\subsection{Combing with Large Language Model}

To facilitate more effective integration with the LLM, we performed two tasks: (1) fine-tuning the language model to endow it with the capability of predicting crystal structures; and (2) incorporating a gating module that enables the model to dynamically switch between generating textual and numerical outputs.

Building upon our approach, which leverages an autoregressive sequence model backbone and joint training on symbolic and numerical data, \ourM{} naturally lends itself to language capabilities. We explored this by loading a pretrained language model and performing instruction tuning on the structure prediction task using language prompts. For example, The MP-20 dataset was reconstructed into an instruction tuning dataset, with the data format detailed in Appendix~\ref{app:instruct}.

To enable the model to seamlessly switch between generating textual and numerical outputs — such as producing a set of structural coordinates followed by explanatory text — it is crucial for the model to autonomously predict whether the feature vector should be routed through the lm head or the diffusion head prior to entering them, rather than relying on this information as a given.

\begin{figure}[!htp]
    \centering
    \includegraphics[width=0.8\textwidth, keepaspectratio]{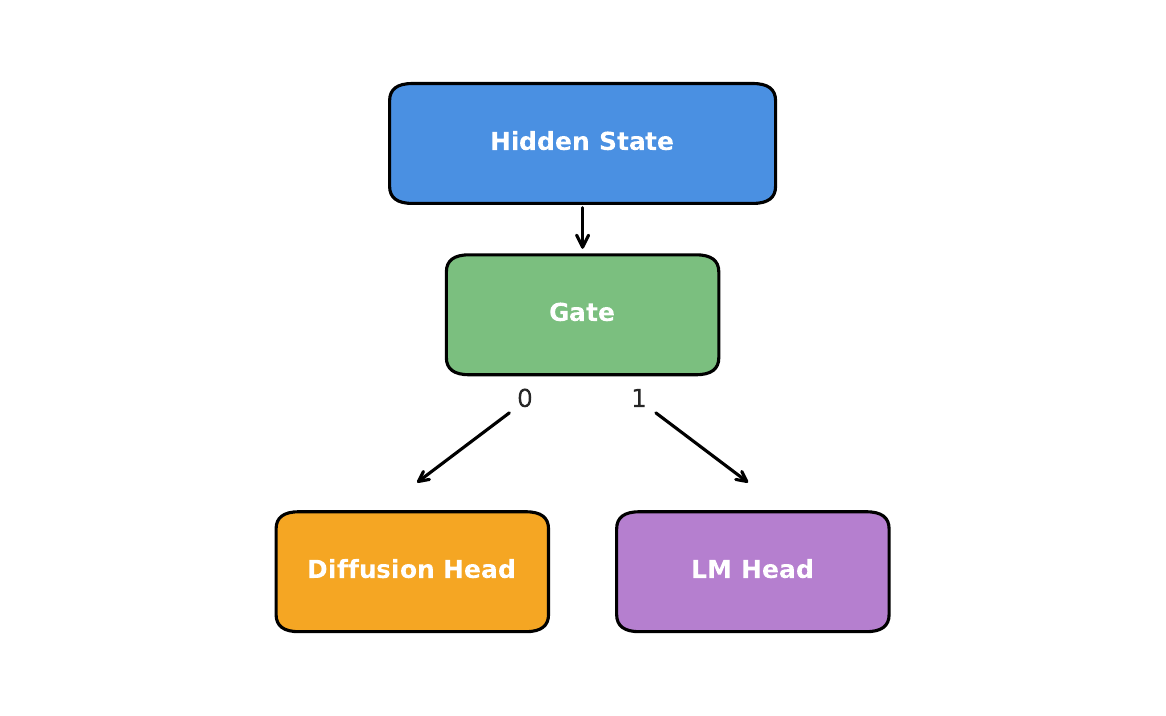} 
    \caption{Diagram of the hidden state, gate, and heads.}
    \label{fig:gate}
\end{figure}

For this purpose, we introduced a gating module shown in Figure~\ref{fig:gate} that performs a binary classification on the model's final hidden state to determine which head it should pass through. 

We then finetuned the pretrained model, which had approximately 1B parameters from NatureLM~\cite{xia2025naturelm}, on MP-20 and QM9, demonstrating its superior performance.

\begin{figure}[!htp]
    \centering
    \includegraphics[width=0.8\textwidth, keepaspectratio]{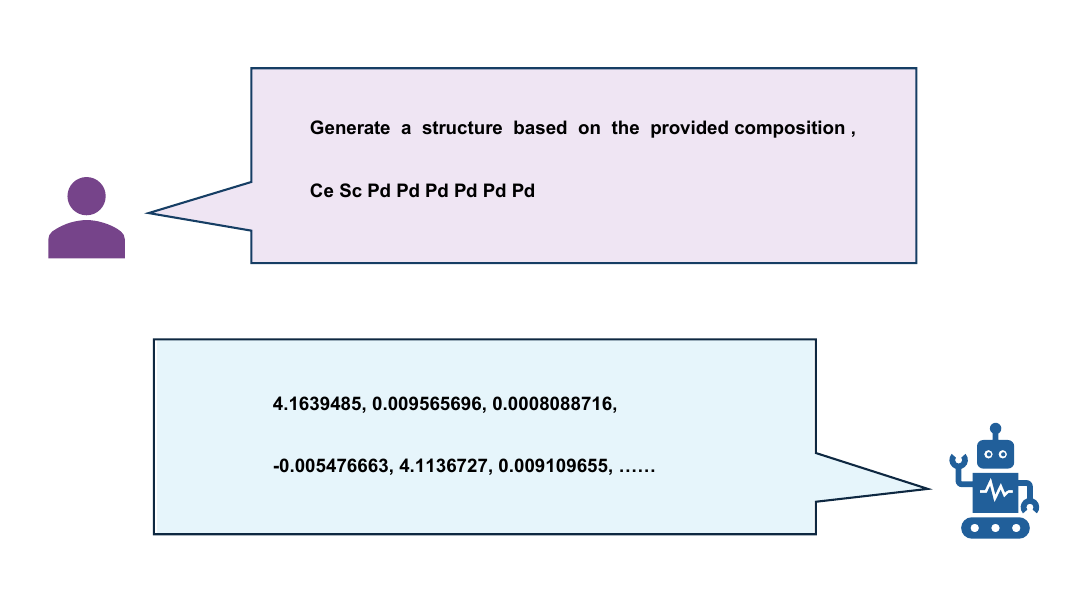} 
    \caption{An example for instruction tunning for the Crystal Structure Prediction (CSP) task.}
    \label{fig:chat}
\end{figure}

\section{Results}
We evaluated \ourM{} on crystal structure prediction and de novo generation in the material domain, conformation generation and conditional generation in the molecular domain, unification and language capacity.

\subsection{Material Results}
\begin{table}[htbp]
\centering
\small
\begin{tabular}{lllllll}
\hline & \multicolumn{2}{c}{ MP-20 } & \multicolumn{2}{c}{ Carbon-24 } & \multicolumn{2}{c}{ MPTS-52 } \\
\hline Methods & MR(\%)$\uparrow$ & RMSD(\textrm{\r{A}}) $\downarrow$ & MR(\%)$\uparrow$ & RMSD(\textrm{\r{A}}) $\downarrow$ & MR(\%)$\uparrow$ & RMSD(\textrm{\r{A}}) $\downarrow$ \\ 
\hline CDVAE\cite{xie2021crystal} & 33.90 & 0.1045 & 17.09 & 0.2969 & 5.34 & 0.2106 \\
\hline DiffCSP\cite{jiao2023crystal} & 51.49 & 0.063 & 17.54 & 0.2759 & 12.19 & 0.1786 \\
\hline FlowMM\cite{miller2024flowmm} & 61.39 & 0.057 & 23.47 & 0.4122 & 17.54 & 0.1726 \\
\hline NatureLM\cite{xia2025naturelm} & 61.78&0.044& \quad \textendash&\quad \textendash&30.20&0.084\\
\hline \ourM{}(400M) & $\mathbf{6 7 . 0 1}$ & $\mathbf{0 . 0 3 7}$ & $\mathbf{3 0 . 0 5}$ & $\mathbf{0 . 2 2 8 6}$ & $\mathbf{3 8 . 6 5}$ & $\mathbf{0 . 0 6 5 7}$ \\
\hline
\end{tabular}
\caption{Crystal structure prediction results on material benchmarks. ``-" indicates unreported results.  Compared to the latest, our model achieves a 10\% improvement in match rate on MP-20, and a 28\% improvement on Carbon-24 and MPTS-52. Our model significantly outperforms the state-of-the-art, with increased performance on longer sequences(materials with larger unit cells).}
\label{tab:csp}
\end{table}

\begin{table}[htbp]
\centering
\small
\begin{tabular}{llllllll}
\hline Methods & InSteps & VS (\%)$\uparrow$ & VC (\%)$\uparrow$ & CR (\%)$\uparrow$ & CP(\%)$\uparrow$ & PW($\rho$) $\downarrow$ & PW(Nel) $\downarrow$\\
\hline CDVAE\cite{xie2021crystal} & 5000 & $\mathbf{100}$ & 86.7 & 99.15 & 99.49 & 0.688 & 0.278 \\
\hline DiffCSP\cite{jiao2023crystal}  & 1000 & $\mathbf{100}$ & 83.25 & $\mathbf{99.71}$ & 99.76 & 0.35 & 0.125 \\
\hline FlowMM\cite{miller2024flowmm} & 250 & 96.58 & 83.47 & 99.48 & 99.65 & 0.261 & 0.107 \\
\hline FlowMM\cite{miller2024flowmm} & 1000 & 96.85 & 83.19 & 99.49 & 99.58 & 0.239 & 0.083 \\
\hline \ourM{}(100M) & $\mathbf{200}$ & 99.08 & $\mathbf{90.12}$ & 99.27 & $\mathbf{99.95}$ & $\mathbf{0.065}$ & $\mathbf{0.04}$ \\
\hline
\end{tabular}
\caption{De novo generation tasks benchmark. The abbreviations in the table refer to: Integration steps(InSteps), Validity Structural(VS), Validity Composition(VC), Coverage Recall(CR), Coverage Precision(CP) and Property Wdist(PW).}
\label{tab:dem}
\end{table}

The evaluation metrics are summarized in Table~\ref{tab:csp} and Table~\ref{tab:dem} for baseline methods and our model in Crystal Structure Prediction and de novo generation, respectively. As shown in Table~\ref{tab:csp}, our model significantly outperforms the previous diffusion/flow matching state-of-the-art, FlowMM, achieving match rate increases of \textbf{10\%}, \textbf{28\%}, and \textbf{120\%} on MP-20, Carbon-24, and MPTS-52, respectively. In terms of RMSD, we observe improvements of \textbf{35\%}, \textbf{45\%}, and \textbf{62\%}. Compared to NatureLM~\cite{xia2025naturelm}, a recently published autoregressive model that surpassed FlowMM on the MP-20 and MPTS-52 benchmarks, our model also demonstrates superior performance, establishing a new state-of-the-art across MP-20, Carbon-24, and MPTS-52.

The significant improvement on MPTS-52 is particularly noteworthy. We speculate that this is due to our model's inherent long-range perception capabilities, a feature often lacking in graph-based representations. The results in Table~\ref{tab:csp}, compared with both diffusion/flow matching and autoregressive models, underscore the importance and effectiveness of combining diffusion and autoregressive models in our approach.

Table~\ref{tab:dem} shows that \ourM{} can generate materials attributed to more reasonable, more precise, and more realistic under the condition of fewer steps.
Specifically, on the Property Wdist metric, which is an index on which previous models did not perform well, \ourM{} can achieve scores of 0.065 for $\rho$ (density) and 0.04 for Nel (number of electrons) in just 200 steps. 
Compared to the previously most advanced model, this represents an improvement of \textbf{73\%} and \textbf{52\%}, respectively. 

\begin{figure}[htbp]
\centering
\includegraphics[width=0.8\textwidth, keepaspectratio]{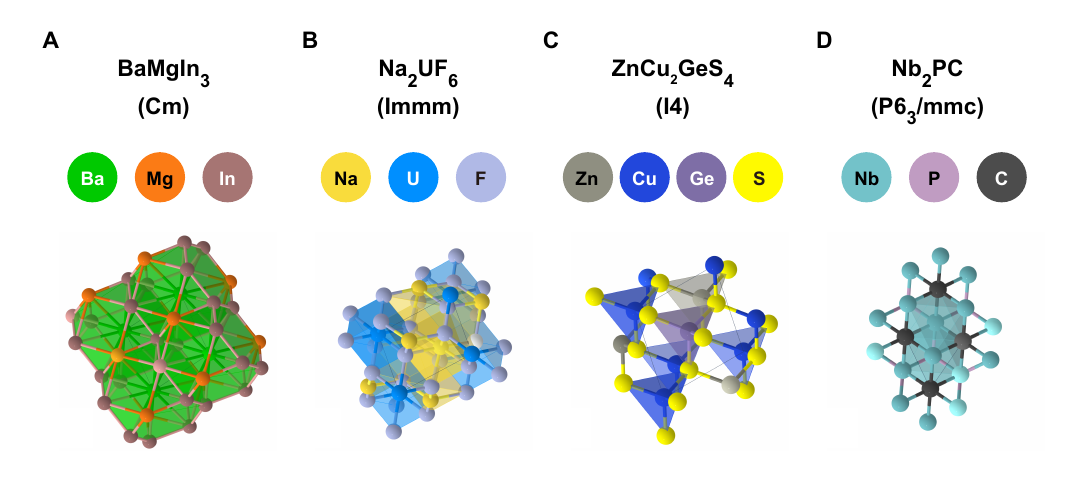} 
\caption{Examples of generated material structures from the crystal structure prediction task. The generated structures closely resemble the ground truth structures; therefore, the latter are omitted for clarity and conciseness.}
\label{fig:material_structure}
\end{figure}

\begin{figure}[htbp]
\centering
\includegraphics[width=0.8\textwidth, keepaspectratio]{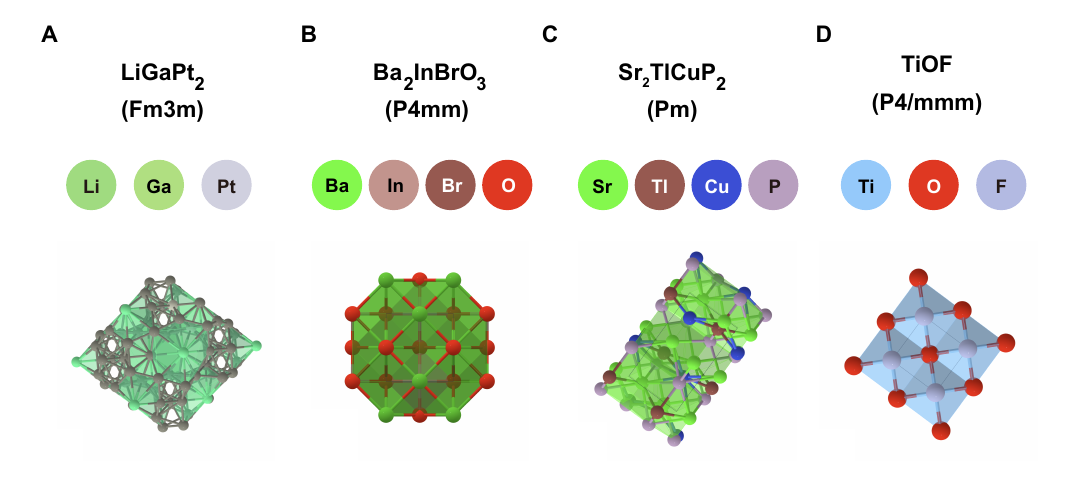} 
\caption{Examples of generated material structure in de novo generation task}
\label{fig:denovo}
\end{figure}

Some examples of crystal structure prediction results are shown in Figure~\ref{fig:material_structure}, and Figure~\ref{fig:denovo} displays some examples of de novo generation, in terms of both material formula and crystal structures.

\begin{figure}[htbp]
\centering
\includegraphics[width=0.8\textwidth, keepaspectratio]{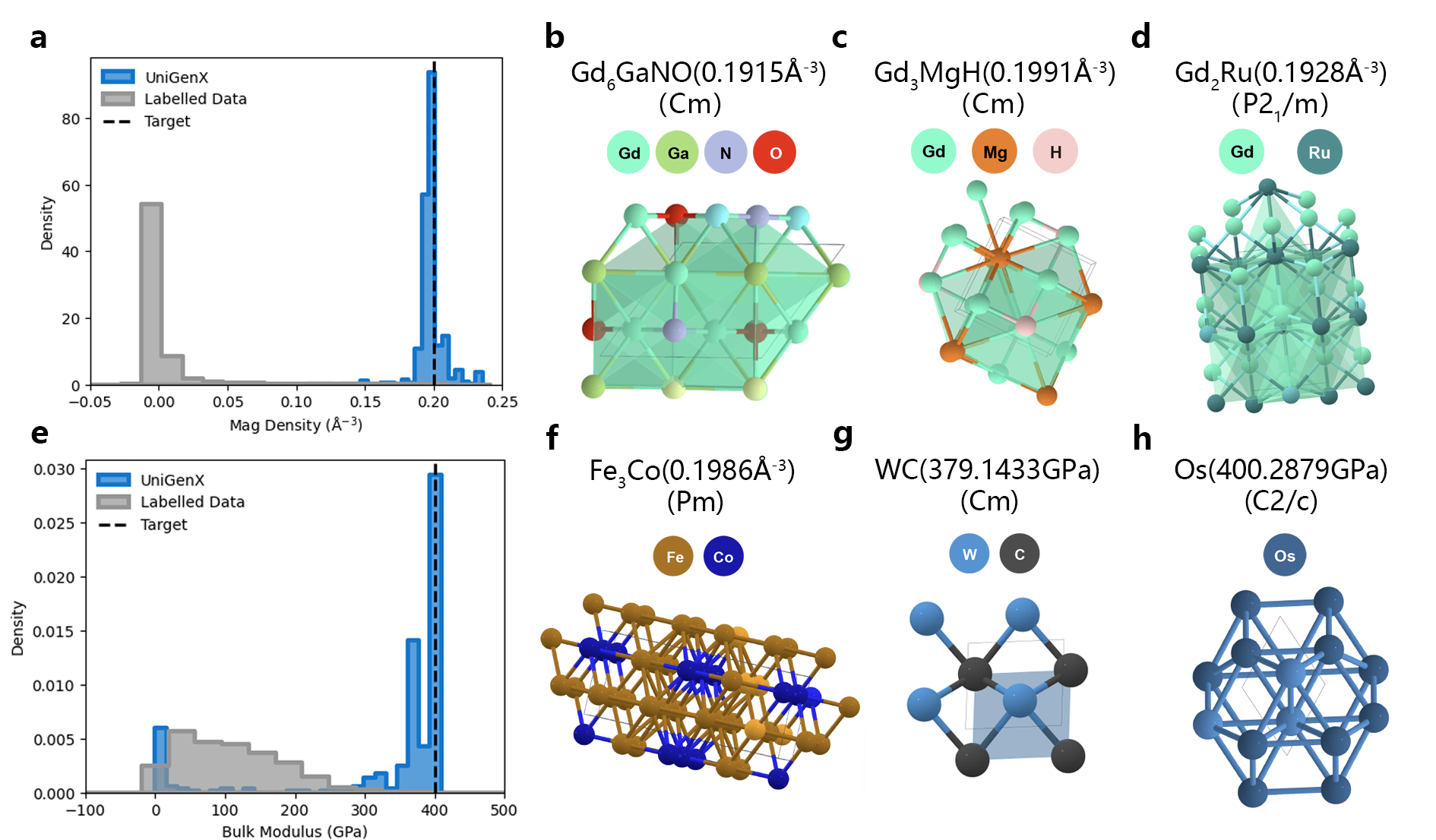} 
\caption{Generated materials with target magnetic and mechanical properties. Panels (a) and (e) show the density distributions of the target properties for the stable samples generated by \ourM{}, alongside the corresponding label distributions in the fine-tuning dataset for magnetic (a) and mechanical (e) properties. The target value specified for \ourM{} is indicated by a black dashed line. Panels (b–d) and (f–h) visualize representative S.U.N. structures generated under magnetic (b–d) and mechanical (f–h) property conditions, each annotated with the reduced formula, space group, and the predicted property value. The model demonstrates the ability to generate samples beyond the support of the training data, particularly in out-of-distribution (OOD) regions, highlighting its strong generalization capability.}
\label{fig:mag&bulk}
\end{figure}

Figure~\ref{fig:mag&bulk} presents the results of the material generation task conditioned on a single property. Panels (a) and (e) show the density distributions of the target properties for the stable samples generated by \ourM{}, alongside the corresponding label distributions in the fine-tuning dataset for magnetic (a) and mechanical (e) properties. The target value specified for \ourM{} is indicated by a black dashed line. Panels (b–d) and (f–h) visualize representative S.U.N. structures generated under magnetic (b–d) and mechanical (f–h) property conditions, each annotated with the reduced formula, space group, and the predicted property value. Notably, \ourM{} demonstrates the ability to generate samples beyond the support of the training data, particularly in out-of-distribution (OOD) regions. This highlights the model’s strong generalization capability—its ability to push forward from one distribution to another.

During experiments, we observed that the temperature and top-p value of \ourM{} have a significant impact on the quality of material generation. Higher temperature and top-p values lead to increased diversity among generated samples and a greater proportion of novel materials. In contrast, lower values result in property distributions that are more sharply concentrated around the target. Under the optimal setting, \ourM{} achieves an inference speed of 0.315 seconds per S.U.N. structure, demonstrating its practical efficiency for structural generation.

\begin{figure}[htbp]
\centering
\includegraphics[width=0.8\textwidth, keepaspectratio]{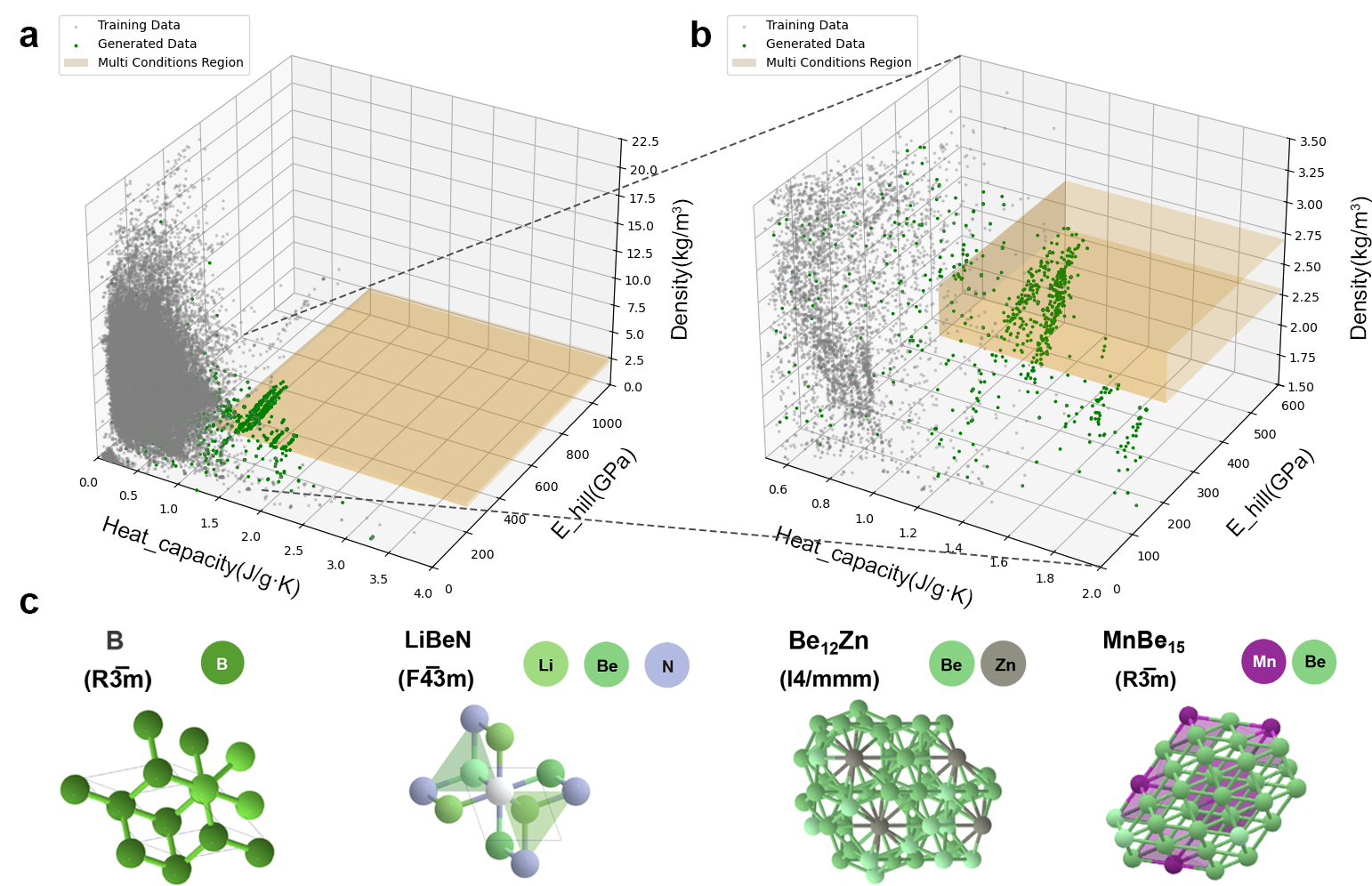} 
\caption{Generating materials with multi-objective constraints. This figure illustrates the performance of \ourM{} on the task of material generation under multiple property constraints. In panel (a), gray dots represent the distribution of labeled training data, while green dots denote the distribution of samples generated by our model. The yellow-shaded region indicates the domain defined by the target constraints for this task ($2.3\,\text{g/cm³} \leq \text{density} \leq 2.7\,\text{g/cm³}$, $\text{heat capacity} > 1\,\text{J/g·K}$, $E_{\text{Hill}} > 200\,\text{GPa}$). Panel (b) provides a detailed view of how the generated samples populate the constrained property space. Panel (c) presents selected examples of SUN (Stable, Unique, and Novel) structures among the generated samples. Out of 1,262 generated candidates, our model successfully identified 436 stable structures that satisfy all specified property constraints. Four of these were confirmed by DFT to be novel, thermodynamically stable compounds not present in existing databases.}
\label{fig:multigen}
\end{figure}

Figure~\ref{fig:multigen} illustrates the performance of \ourM{} on the task of multi-property-constrained material generation. In panel (a), gray dots represent the distribution of labeled training data, while green dots denote the distribution of structures generated by our model. The yellow-shaded region defines the desired target domain: $2.3,\text{g/cm³} \leq \text{density} \leq 2.7,\text{g/cm³}$, $\text{heat capacity} > 1,\text{J/g·K}$, and $E_{\text{Hill}} > 200,\text{GPa}$. Panel (b) zooms in on the coverage of the generated samples within this constrained property space, and panel (c) showcases selected SUN structures among the successful generations. Out of \num{1262} generated candidates, our model identified \num{436} structures predicted to be both stable and compliant with all target property constraints.

To further evaluate novelty, we conducted a rapid screening procedure based solely on elemental composition, filtering out structures present in the training set. This process yielded 11 novel candidates, of which 4 were confirmed through DFT calculations to satisfy all property requirements while also exhibiting thermodynamic stability. None of these four materials were found in the Materials Project database, suggesting they may represent previously unreported compounds.

Although this screening was intentionally limited in scope for computational efficiency, it demonstrates the model’s ability to generate valid, constraint-satisfying, and potentially novel materials. A more exhaustive comparison across the full set of generated structures would likely uncover additional unique candidates, offering further evidence of \ourM{}’s potential in accelerating inverse material design under multi-objective constraints.



\subsection{Molecular Results}\label{sec:molecule}

\begin{table}[htbp]
\centering
\small
\begin{tabular}{lllllllll}
\hline Dataset & \multicolumn{4}{c}{ Large-scale QM9 } & \multicolumn{4}{c}{ Large-scale Drugs } \\
\hline Methods & \multicolumn{2}{c}{COV-P $(\%) \uparrow$} & \multicolumn{2}{c}{MAT-P $(\textrm{\AA}) \downarrow$} & \multicolumn{2}{c}{ COV-P $(\%) \uparrow$} & \multicolumn{2}{c}{MAT-P $(\textrm{\AA}) \downarrow$} \\
 & Mean & Median & Mean & Median & Mean & Median & Mean & Median \\
\hline ConfGF\cite{shi2021learning} & 46.23 & 44.87 & 0.5171 & 0.5133 & 28.23 & 20.71 & 1.6317 & 1.6155 \\
\hline GeoMol\cite{ganea2021geomol} & 78.28 & 81.03 & 0.3790 & 0.3861 & 41.46 & 36.79 & 1.5120 & 1.5107 \\
\hline DMCG\cite{zhu2022direct} & 90.86 & 95.36 & $\mathbf{0.2305}$ & 0.2258 & 74.57 & 81.80 & $\mathbf{0.9940}$ & 0.9454 \\
\hline \ourM{}(400M) & $\mathbf{91.40}$ & $\mathbf{100.00}$ & 0.2516 & $\mathbf{0.1070}$ & $\mathbf{76.94}$ & $\mathbf{87.50}$ & 1.3394 & $\mathbf{0.8631}$ \\
\hline
\end{tabular}
\caption{Conformation generation: precision-based coverage and matching score. Our model surpasses the state-of-the-art on most metrics.}
\label{tab:qm9p}
\end{table}

\begin{table}[htbp]
\centering
\small
\begin{tabular}{lllllllll}
\hline Dataset & \multicolumn{4}{c}{ Large-scale QM9 } & \multicolumn{4}{c}{ Large-scale Drugs } \\
\hline Methods & \multicolumn{2}{c}{COV $(\%) \uparrow$} & \multicolumn{2}{c}{MAT $(\textrm{\AA}) \downarrow$} & \multicolumn{2}{c}{ COV $(\%) \uparrow$} & \multicolumn{2}{c}{MAT $(\textrm{\AA}) \downarrow$} \\
 & Mean & Median & Mean & Median & Mean & Median & Mean & Median\\
\hline ConfGF\cite{shi2021learning} & 89.21 & 95.12 & 0.2809 & 0.2837 & 70.92 & 85.71 & 1.0940 & 1.0917 \\
\hline GeoMol\cite{ganea2021geomol} & 91.05 & 95.55 & 0.2970 & 0.2993 & 69.74 & 83.56 & 1.1110 & 1.0864 \\
\hline DMCG\cite{zhu2022direct} & $\mathbf{98.34}$ & $\mathbf{100}$ & 0.1486 & 0.1340 & $\mathbf{96.22}$ & $\mathbf{100.00}$ & 0.6967 & 0.6552 \\
\hline \ourM{}(400M) & 92.67 & $\mathbf{100}$ & $\mathbf{0.1466}$ & $\mathbf{0.0856}$ & 92.09 & $\mathbf{100.00}$ & $\mathbf{0.6536}$ & $\mathbf{0.6120}$ \\
\hline
\end{tabular}
\caption{Conformation generation: Recall-based coverage and matching score}
\label{tab:qm9}
\end{table}

As shown in Table~\ref{tab:qm9p} and Table~\ref{tab:qm9}, the conformations generated by our model are able to achieve good Coverage and Match score whether based on recall or precision. Overall, the results are comparable to DMCG~\cite{zhu2022direct}. It is noticeable that for every metric's median we outperform the baselines; the remaining ones are only a little lower than DMCG.

\begin{table}[htbp]
\centering
\small
\begin{tabular}{llllll}
\hline
\textbf{Methods} & $\overline{E}$ & $E_{\text{min}}$ & $\overline{\Delta \epsilon}$ & $\Delta \epsilon_{\text{min}}$ & $\Delta \epsilon_{\text{max}}$ \\ \hline
RDKit& 0.8875 & 0.6530  & 0.3484  & 0.5570 & 0.2399 \\ \hline
ConfGF\cite{shi2021learning}  & 2.8349 & 0.2012 & 0.6903  & 4.9221  & 0.1820  \\ \hline
GeoMol\cite{ganea2021geomol} & 4.5700  & 0.5096  & 0.5616   & 3.5083 & 0.2650    \\ \hline
DMCG\cite{zhu2022direct}  & 0.4324 & 0.1364 & 0.2057 & 1.3229 &0.1509         \\ \hline
\ourM{}(400M) & \textbf{0.1464} & \textbf{0.0746}  & \textbf{0.1004} & \textbf{0.2241} & \textbf{0.1360} \\ \hline
\end{tabular}
\caption{Mean absolute error of predicted ensemble properties (Unit: eV)}
\label{tab:prop}
\end{table}

\begin{figure}[htbp]
\centering
\small
\includegraphics[width=0.9\textwidth, keepaspectratio]{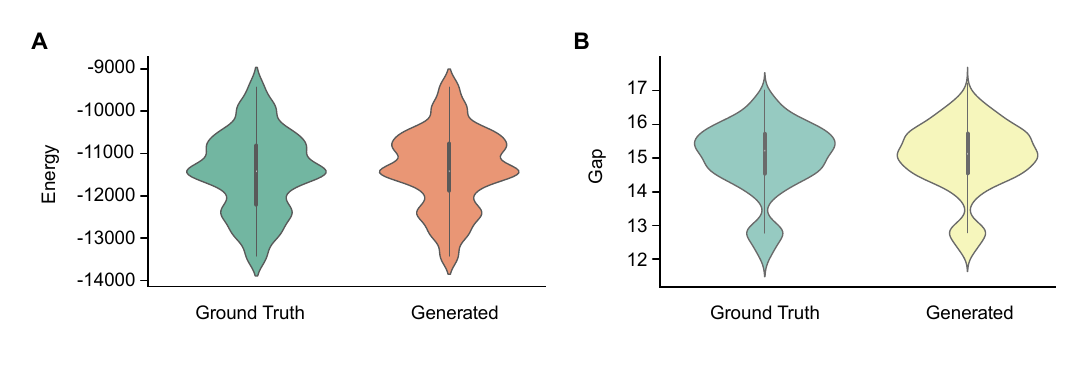} 
\caption{Comparison of energy (A) and gap (B) property distributions between ground truth and generated conformations.}
\label{fig:molecule_property}
\end{figure}

The results of property evaluation are shown in Table~\ref{tab:prop}. Our method significantly outperforms SOTA and baseline methods listed in  the Table, which shows the effectiveness of our method. Comparison between the distribution of energies and gaps of ground truth and generated conformations is in Figure~\ref{fig:molecule_property}. 

\begin{table}[htbp]
\centering
\small
\begin{tabular}{lccccc}
\toprule
\textbf{Model} & $\alpha(Bohr^3)$ & $\epsilon_{H}(mHa)$  & $\epsilon_{L}(mHa)$  & $\Delta \epsilon(mHa)$  & $\mu(D)$ \\
\midrule
Random & 41.00 & 103.30 & 121.83 & 193.36 & 8.40 \\
EDM\cite{hoogeboom2022equivariant} & 20.15 & 158.70 & 166.20 & 287.00 & 7.01 \\
LDM-3DG\cite{you2024latent} & 15.56 & 54.62 & 63.08 & 107.14 & 6.33 \\
LDM-3DG-GSSL\cite{you2024latent} & 16.43 & 55.03 & 66.53 & 113.15 & 9.22 \\
DiGress\cite{vignac2022digress} & 9.23 & 31.98 & 105.06 & 90.57 & 1.49 \\
SeaDAG\cite{zhou2024seadagsemiautoregressivediffusionconditional} & 8.85 & 30.91 & 103.03 & 89.70 & 1.33 \\
\ourM{}(100M) & \textbf{2.46} & \textbf{18.40} & \textbf{21.47} & \textbf{55.77} & \textbf{0.99} \\
\bottomrule
\end{tabular}
\caption{Conditional generation on five quantum properties(unit) evaluation following LDM setting in \cite{you2024latent}. Numbers represent the MAE between conditional and oracle-predicted properties~\cite{garcia2021n}.}
\label{tab:cond_mol_ldm}
\end{table}

\begin{table}[htbp]
\centering
\small
\begin{tabular}{l | c c c c c c}
\toprule
Property & $\alpha$& $\Delta \varepsilon$ & $\varepsilon_{\mathrm{HOMO}}$ & $\varepsilon_{\mathrm{LUMO}}$ & $\mu$ & $C_v$\\
Units & Bohr$^3$ & meV & meV & meV & D & $\frac{\text{cal}}{\text{mol}}$K  \\
\midrule
QM9 & 0.10  & 64 & 39 & 36 & 0.043 & 0.040  \\
\midrule
Random & 9.01 &  1470 & 645 & 1457  & 1.616  & 6.857   \\
EDM\cite{hoogeboom2022equivariant} & 2.76 & 655 & 356 & 584 & 1.111 & 1.101 \\
GeoLDM\cite{xu2023geometric} & 2.37 & 587  & 340 & 522 & 1.108 & 1.025 \\
GeoBFN\cite{song2024unified} & 2.34 & 577 & 328 & 516 & 0.998 & 0.949 \\
Geo2Seq with Mamba\cite{li2024geometry} & 0.46 & 98 & 57 & 71 & 0.164 & 0.275 \\
Geo2Seq with GPT\cite{li2024geometry} & 0.53 & 102 & 48 & 53 & 0.097 & 0.325 \\
MOL-STRUCTOK\cite{gao2024tokenizing3dmoleculestructure} & \textbf{0.33} & 89 & 64 & 62 & 0.285 & \textbf{0.169}  \\
\ourM{}(100M) & 0.38 & \textbf{65} & \textbf{39} & \textbf{35} & \textbf{0.045} & 0.194  \\
\bottomrule
\end{tabular}
\caption{Conditional generation results for five quantum properties (units specified), evaluated using the EDM setting from \cite{hoogeboom2022equivariant}. Numbers represent the mean absolute error (MAE) between conditional and oracle-predicted properties~\cite{garcia2021n}. The QM9 row indicates the baseline MAE of the EGNN model, used to obtain the oracle-predicted properties for generated samples, compared to QM9 property labels.}
\label{tab:cond_mol_edm}
\end{table}

Table~\ref{tab:cond_mol_ldm} presents sampling results according to equation~\ref{ldm} for conditional generation as introduced in \cite{you2024latent}, where \ourM{} achieves state-of-the-art (SOTA) performance across all five properties. Compared with the latest state-of-the-art methods, \ourM{} achieves up to a 260\% improvement and at least a 34\% improvement across various evaluation metrics.

\begin{remark} \textbf{Corrected the typo from meV to Hain in Table \ref{tab:cond_mol_ldm}.}
There is a typo with the evaluation results provided by \cite{you2024latent}. Following \cite{hoogeboom2022equivariant}, when we used another portion of the QM9 dataset to assess the performance of the classifier from \cite{you2024latent}, we found that the values for $\epsilon_{H}$, $\epsilon_{L}$ and $\Delta\epsilon$ were approximately 27.21 times larger than those reported in \cite{hoogeboom2022equivariant}. Given that $1 Ha = 27.2114 eV$, we suspect there is a unit discrepancy in \cite{you2024latent}. While the unit in their paper is listed as meV, we believe the correct unit should be mHa. Confirmed with the authors in the papers and we corrected it in our Table \ref{tab:cond_mol_ldm}.
\end{remark}

Table~\ref{tab:cond_mol_edm} presents the results sampling as given in equation~\ref{edm} , as introduced in \cite{hoogeboom2022equivariant}. \ourM{} achieves state-of-the-art (SOTA) performance on four out of six properties. Notably, for the property $\mu$, our model demonstrates a significant improvement of $53.6\%$ over the previous SOTA~\cite{li2024geometryinformedtokenizationmolecules}. However, performance on $\alpha$ and $C_v$ is slightly lower compared to \cite{gao2024tokenizing3dmoleculestructure}. This can be attributed to two factors: first, \cite{gao2024tokenizing3dmoleculestructure} employs separate models for each property, whereas \ourM{} uses a unified model, potentially leading to performance trade-offs. Second, \cite{gao2024tokenizing3dmoleculestructure} incorporates post-generation molecular force field optimization, following \cite{you2024latent}, which leverages domain-specific expert knowledge. Our approach evaluates the model's direct output without such optimization. While these differences may impact direct comparison, \ourM{} still excels in key performance metrics, highlighting its versatility and robustness.

The properties in Table~\ref{tab:cond_mol_edm} were calculated for generated samples using an EGNN model, consistent with the EDM settings from \cite{hoogeboom2022equivariant}. The QM9 row indicates the EGNN model's error estimation relative to the QM9 property labels. Notably, \ourM{}'s generated results exhibit error closely approximating the validation model's error. For instance, the MAE of HOMO energy level is identical to QM9 (39meV vs. 39meV), and the MAE of LUMO is lower (35meV vs. 36meV). This suggests that \ourM{}'s performance may exceed the validation model's capacity for accurate error assessment. A detailed discussion of this discrepancy is reserved for future work.

\begin{figure}[htbp]
\centering
\begin{subfigure}[t]{0.48\textwidth}
    \includegraphics[width=\textwidth, keepaspectratio]{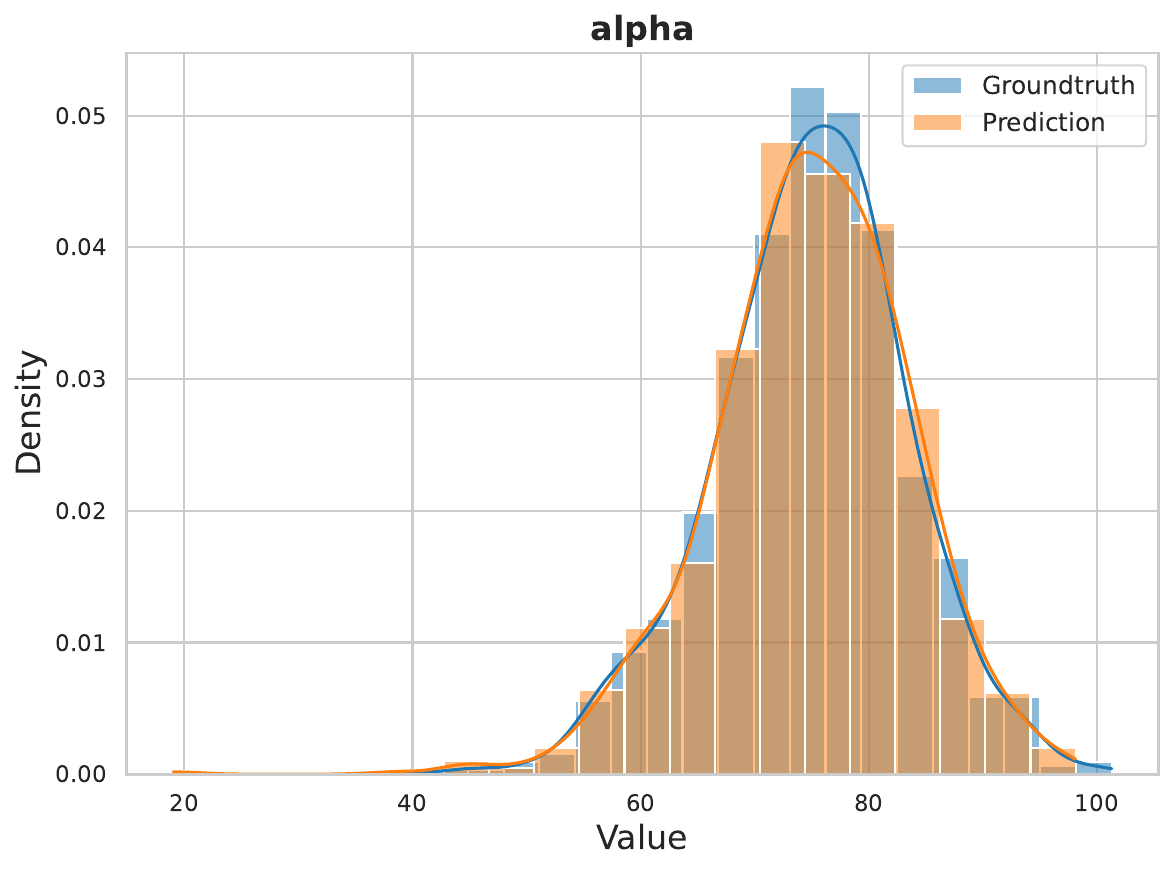}
    \caption{Kernel density estimate (KDE) plot comparing the distribution of the given target property and the predicted property values of generated molecules.}
    \label{fig:single_prop}
\end{subfigure}
\hfill
\begin{subfigure}[t]{0.48\textwidth}
    \includegraphics[width=\textwidth, keepaspectratio]{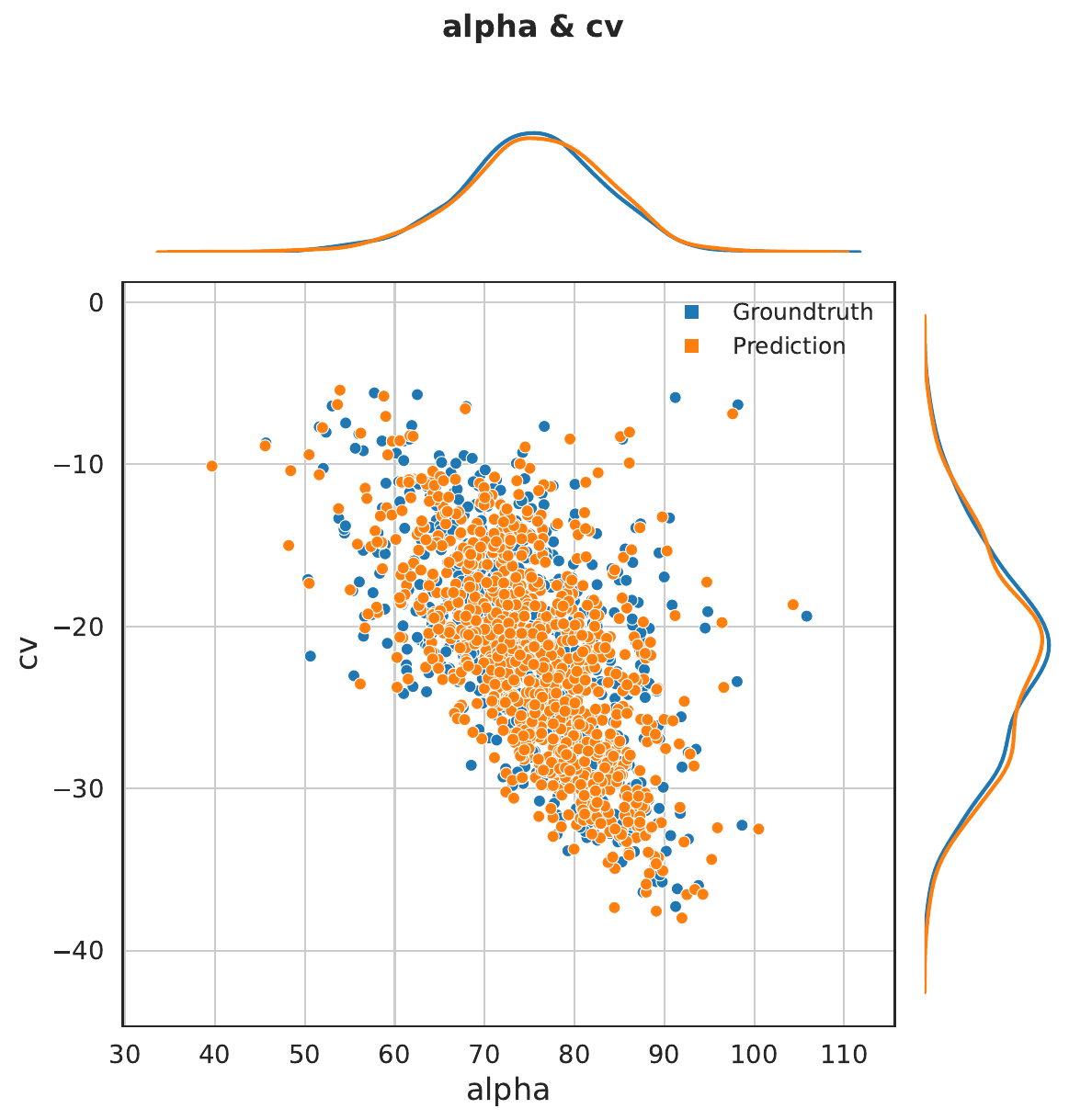}
    \caption{Scatter plot showing the joint distribution of two predicted properties for generated molecules conditioned on given target properties.}
    \label{fig:two_prop}
\end{subfigure}
\label{fig:multi_props}
\end{figure}

\begin{figure}[htbp]
\centering
\includegraphics[width=0.8\textwidth, keepaspectratio]{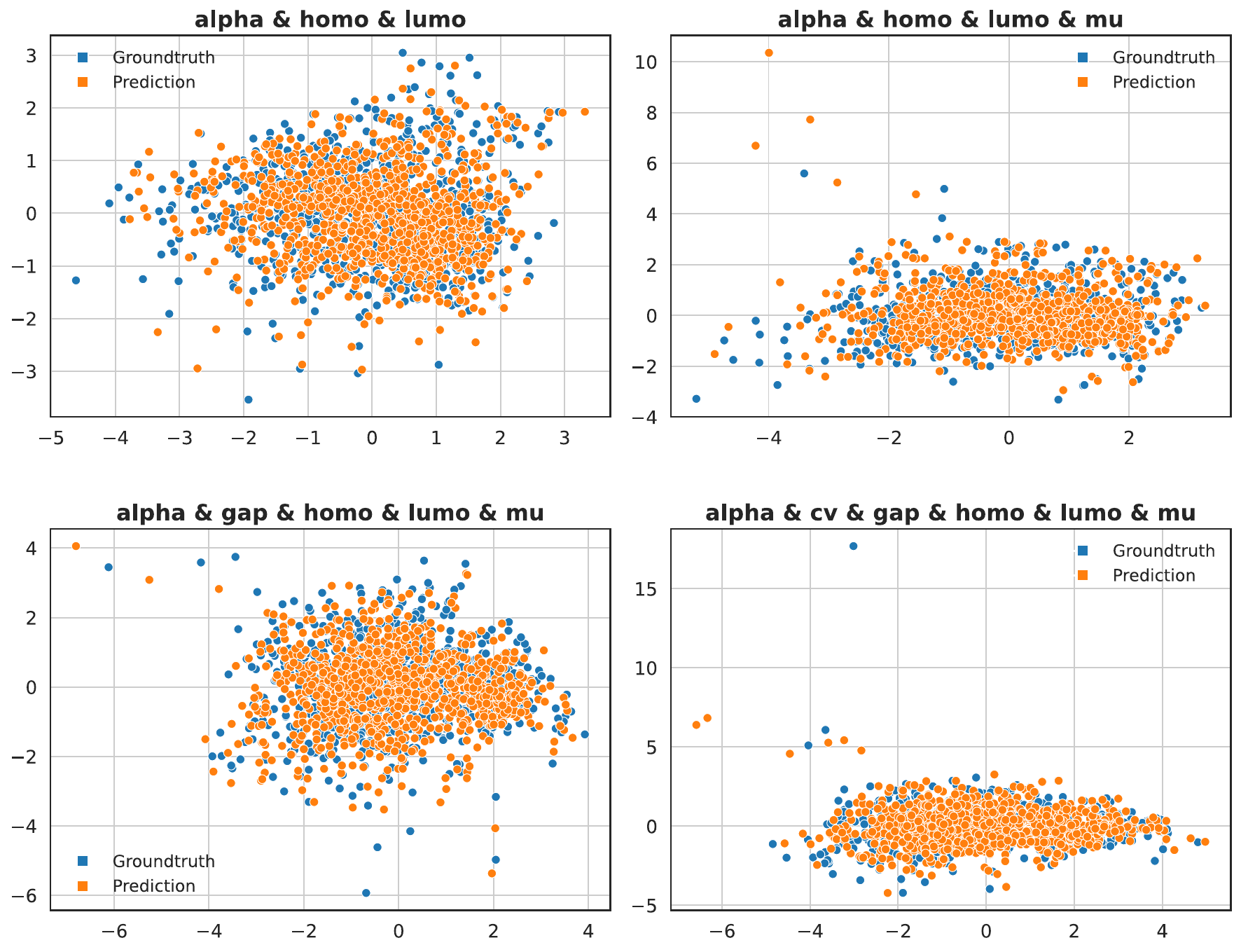} 
\caption{PCA projection of predicted properties of generated molecules conditioned on three or more target properties.}
\label{fig:pca4}
\end{figure}

For the multiple-property generation task, we visualize the results in the following way: for a single property, we plot a kernel density histogram~\ref{fig:single_prop}; for two properties, we use a scatter plot~\ref{fig:two_prop}; and for three or more properties, we apply principal component analysis (PCA) to reduce the dimensions and plot a 2D scatter plot~\ref{fig:pca4}. These visualizations demonstrate that our model can flexibly generate molecules conditioned on varying numbers of properties, while maintaining stability and reliability.


\begin{figure}[htbp]
\centering
\includegraphics[width=0.8\textwidth, keepaspectratio]{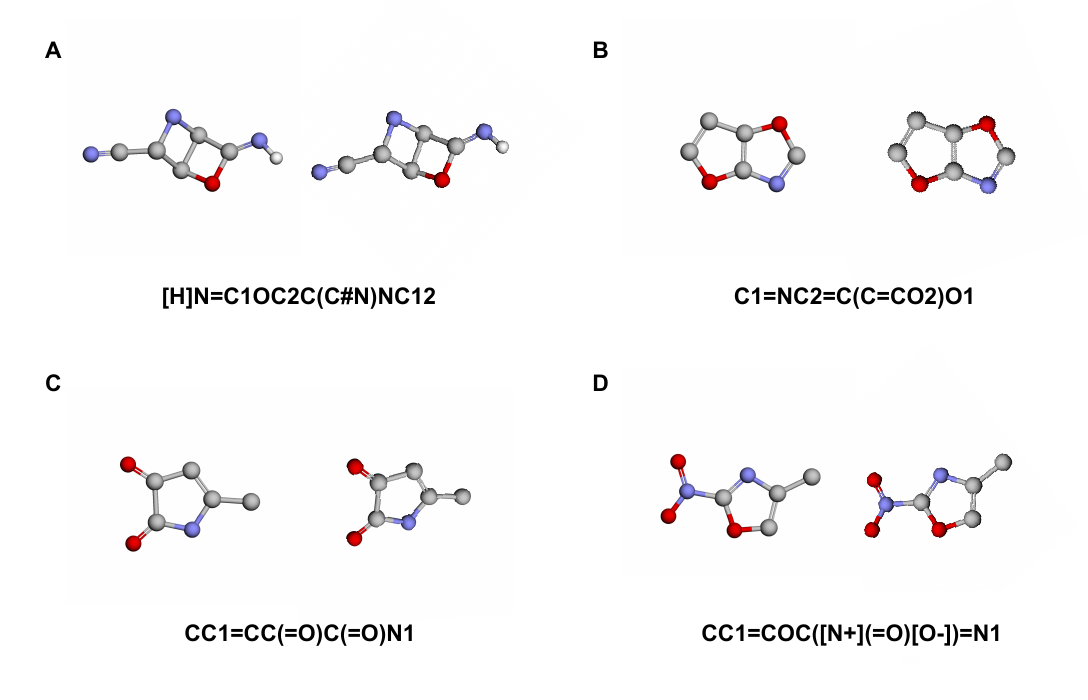} 
\caption{Examples of predicted structures. The groundtruth and predicted structures are shown side-by-side, with groundtruth on the left, generated structures on the right.}
\label{fig:molecule_structure}
\end{figure}

Furthermore, four examples are displayed in Figure \ref{fig:molecule_structure}, in each example on the left is ground truth and on the right is the generated conformation.

\subsection{Protein Results}

\begin{figure}[htbp]
\centering
\includegraphics[width=1.0\textwidth, keepaspectratio]{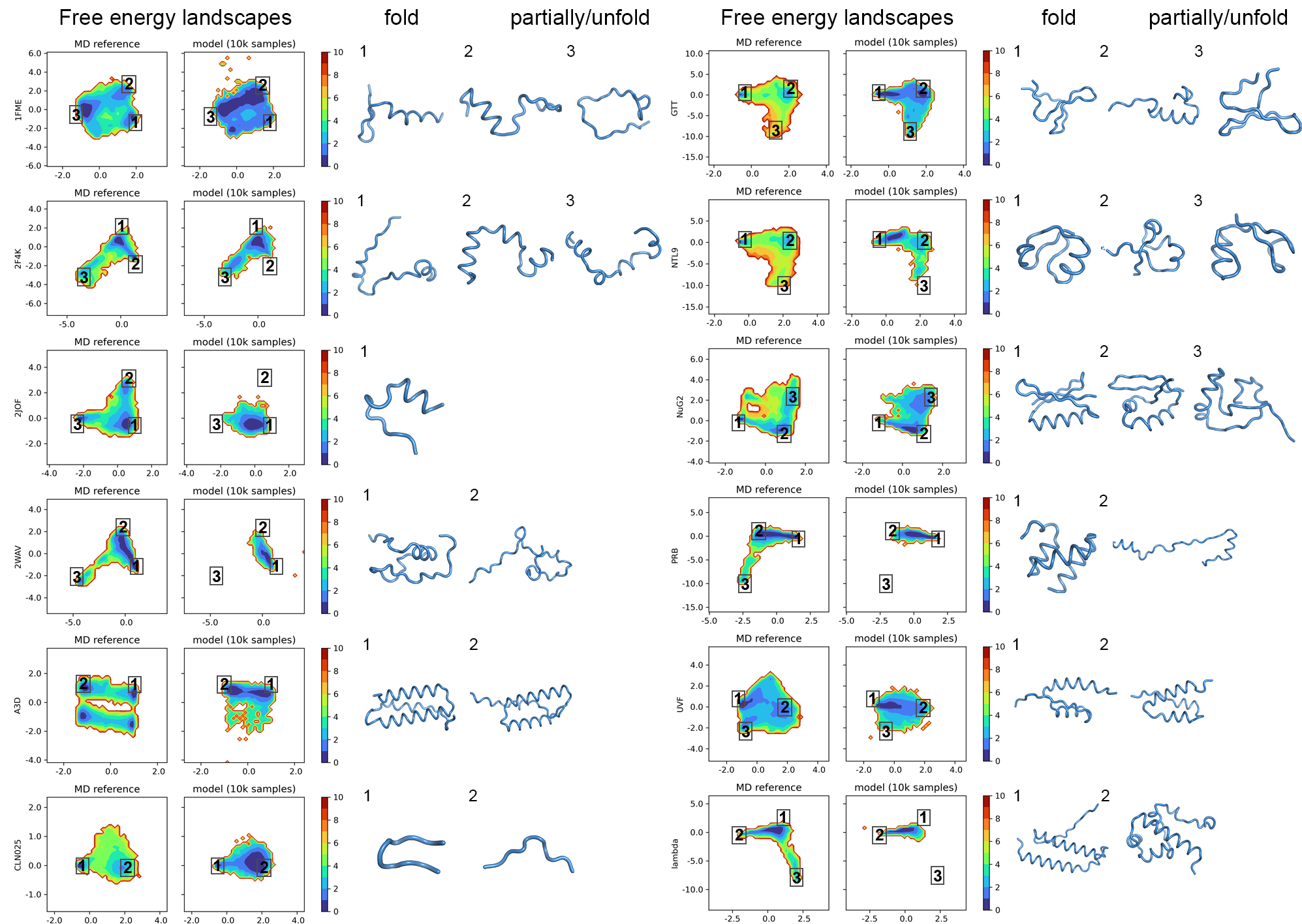} 
\caption{Free energy surfaces of DESRES fast-folding proteins. The landscapes on the left are obtained from long-timescale molecular dynamics (MD) simulations, while the landscapes on the right are generated by \ourM{}. The model successfully captures the key features of the energy landscapes, including native and unfolded states, with an average deviation of only 0.91 kcal/mol from the MD reference---a discrepancy comparable to that between different classical force fields~\cite{best2011free,hahn2024current}. Representative conformational states predicted by \ourM{} are also shown.}
\label{fig:protein_md}
\end{figure}

As shown in Figure~\ref{fig:protein_md}, for all proteins, our model successfully predicts native and unfolded states with similar topological features on the free energy landscape. The predicted two-dimensional free energy surfaces closely resemble those obtained from MD simulations. Notably, our model captures partial or complete folding intermediates that are visible on the landscape, demonstrating its capability to approximate the equilibrium distribution. Quantitatively, the average discrepancy between our model and molecular dynamics (MD) simulations on the 2D free energy landscapes is 0.91 kcal/mol, ranging from 0.64 kcal/mol for 1FME to 1.20 kcal/mol for A3D. Notably, this level of deviation is comparable to the expected differences between two classical MD force fields \cite{best2011free,hahn2024current}, suggesting that \ourM{} achieves near-physically consistent accuracy.

As shown in Table~\ref{tab:ecnumtab}, \ourM{} is able to generate protein sequences that not only match the target EC number but also exhibit strong structural quality and functional relevance. When comparing \ourM{} with and without structural conditioning during training, we observe consistent improvements across both sequence-level and structure-level metrics. Specifically, models trained with structural information yield higher pLDDT and pTM scores, along with significantly better TM-scores and lower RMSDs when benchmarked against both AFDB and PDB reference structures.

Moreover, comparison with structures predicted by ESMFold (Table~\ref{tab:ecnum_struct}) reveals a high degree of structural agreement: generated sequences consistently achieve TM-scores above 0.85 and low RMSD values, indicating that \ourM{} reliably produces proteins that fold into plausible and stable 3D structures aligned with the intended enzymatic function.

\begin{figure}[htbp]
\centering
\includegraphics[width=1.0\textwidth, keepaspectratio]{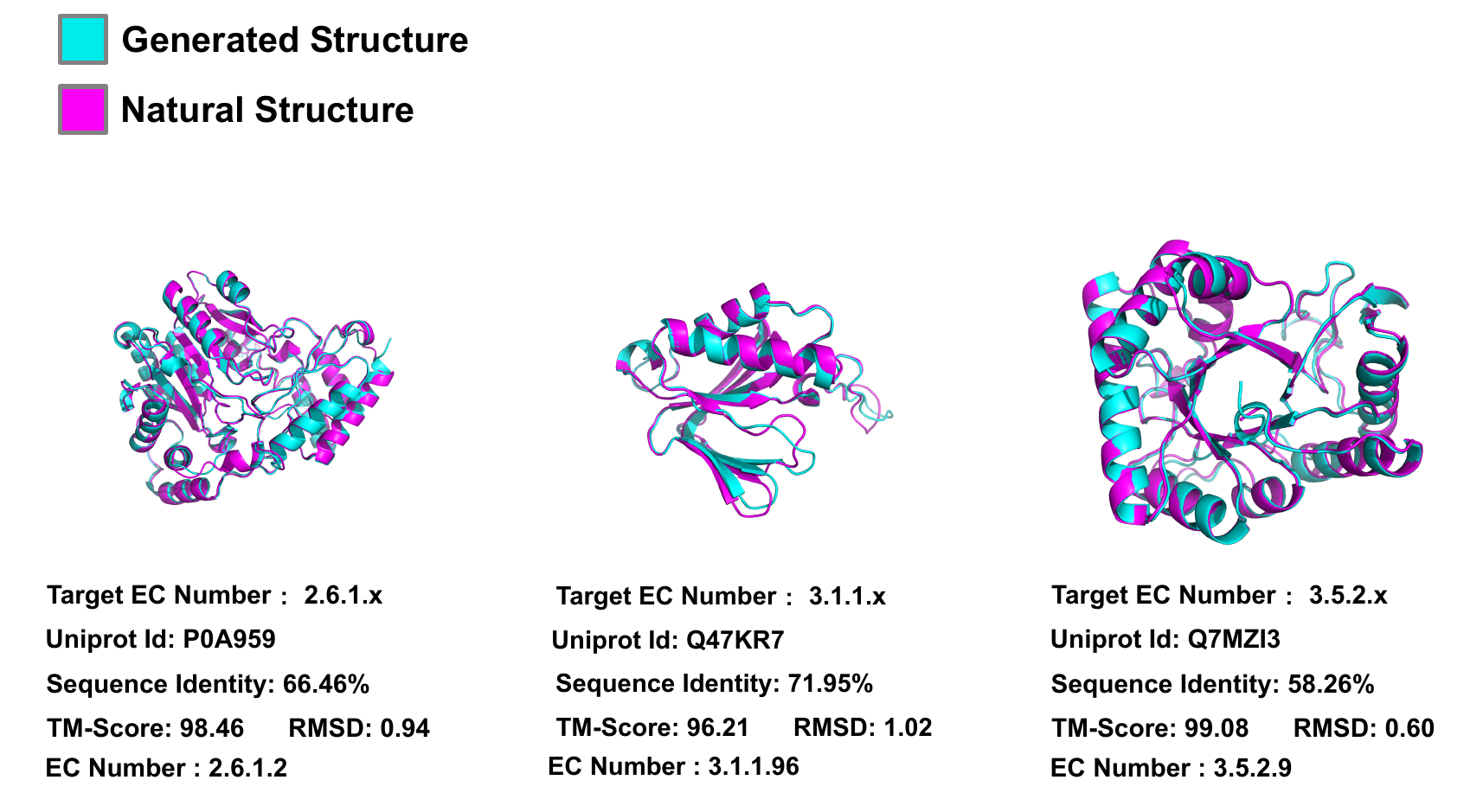} 
\caption{For all three cases, the proteins generated by \ourM{} exhibit the close structural alignment to naturally occurring proteins with relative small sequence identity.}
\label{fig:ecnum_case}
\end{figure}

Figure~\ref{fig:ecnum_case} presents one representative case for each of the three tested EC numbers. Notably, despite relatively low sequence identity to naturally occurring proteins, the generated sequences fold into structures with strong similarity to experimentally resolved counterparts. This demonstrates that \ourM{} can generalize beyond sequence similarity, capturing structural patterns that are functionally meaningful and biophysically consistent.


\begin{table*}[ht]
\centering
\small
\begin{tabular}{cl|ccc|cc}
\hline
\multicolumn{2}{c|}{\textbf{Tested EC number}} & \multicolumn{3}{c|}{\textbf{Overview}} & \multicolumn{2}{c}{\textbf{Sequence}} \\
\cline{1-2} \cline{3-5} \cline{6-7}
 & & pLDDT & pTM & Time(s) & Accuracy (\%) & Avg. Top Hit Identity (\%) \\
\hline
\multirow{3}{*}{\textbf{\ourM{}}} 
& 2.6.1.x & 80.11 & 0.821 & 9.38 & 63.87 & 77.35 \\
& 3.1.1.x & 87.25 & 0.883 & 3.77 & 76.42 & 75.31 \\
& 3.5.2.x & 91.59 & 0.938 & 8.28 & 84.95 & 77.49 \\
\hline
\multirow{3}{*}{\textbf{w/o struct.}} 
& 2.6.1.x & 74.96 & 0.766 & 9.20 & 69.48 & 71.58 \\
& 3.1.1.x & 85.83 & 0.878 & 3.37 & 74.28 & 70.07 \\
& 3.5.2.x & 87.46 & 0.911 & 9.13 & 79.91 & 71.84 \\
\hline
\end{tabular}
\caption{Evaluation of function-guided enzyme generation. The performance of \ourM{} trained with both sequence and structure is compared to a sequence-only baseline. Including structural information during training leads to generated proteins with higher predicted structural quality (pLDDT, pTM) and functional accuracy.}
\label{tab:ecnumtab}
\end{table*}

\begin{table}[ht]
\centering
\small
\begin{tabular}{ccc}
\hline
\textbf{Tested EC Number} & \textbf{RMSD (\AA)} & \textbf{TM-score} \\
\hline
2.6.1.x & 4.398 & 0.916 \\
3.1.1.x & 6.047 & 0.867 \\
3.5.2.x & 3.206 & 0.936 \\
\hline
\end{tabular}
\caption{Structural comparison between \ourM{} generated structures and ESMFold-predicted structures across three EC numbers.}
\label{tab:ecnum_struct}
\end{table}

\subsection{Protein-Ligand Docking Results}
We benchmark our proposed \ourM{} (100M) model against a modified version of 3DMolformer\cite{hu20253dmolformer}, in which the diffusion module is replaced with a deterministic linear transformation. To ensure a fair comparison, both models adopt the same Transformer backbone and are trained under identical hyperparameters for 1,000 epochs using a cosine learning rate schedule. Quantitative results are presented in Table~\ref{tab:unigenx_vs_linear_eval}.

\begin{figure}[htbp]
    \centering
    \includegraphics[width=0.8\textwidth, keepaspectratio]{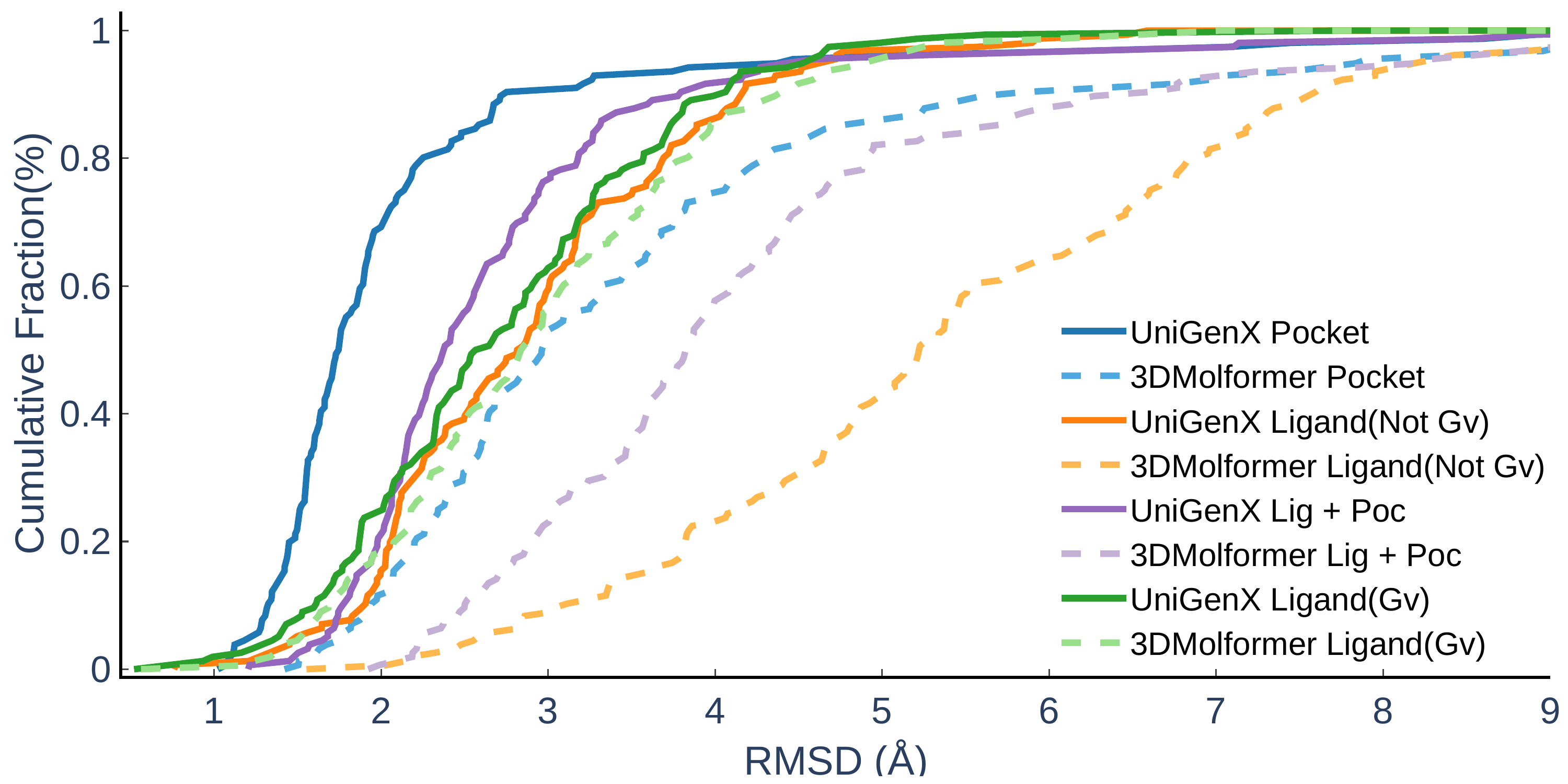} 
    \caption{Cumulative Distribution of RMSD Values for Predicted Ligand and Pocket Structures.}
    \label{fig:docking_rmsd}
\end{figure}

To complement these aggregate metrics with a more nuanced view of performance, Figure~\ref{fig:docking_rmsd} shows the cumulative distribution functions (CDFs) of RMSD values for predicted ligand and pocket structures across different modeling setups. These curves reveal the proportion of samples below varying RMSD thresholds, offering a finer-grained perspective than summary statistics such as mean or median.

The first task evaluates the ability to predict ligand structures given both the apo and holo pocket conformations. \ourM{} exhibits clear superiority across all metrics: the percentage of ligand predictions within 2Å RMSD improves from 1.27\% to 21.66\%, representing a 17-fold increase, while the median RMSD drops from 3.82Å to 2.37Å. Consistent gains are also observed at the 25th and 75th percentiles, as well as in mean RMSD, indicating not only improved accuracy but also greater robustness across diverse samples.

The second task, which requires predicting both the holo pocket and ligand coordinates from the apo structure alone, presents a more demanding setting that involves modeling pocket rearrangements—an essential aspect of induced fit. Here, the advantages of \ourM{} are even more pronounced. For pocket prediction, \ourM{} achieves a median RMSD of 1.75Å and a 2Å success rate of 69.43\%, compared to 2.96Å and 12.10\% for the baseline. For ligand prediction, the improvements are similarly substantial. When the pocket is provided (Pocket Gv), \ourM{} reaches 24.84\% within 2Å RMSD, outperforming the baseline’s 18.47\%. In the more challenging case where the pocket must also be inferred (Pocket Not Gv), \ourM{} achieves 15.92\% within 2Å—an over 23-fold improvement compared to the baseline’s 0.64\%.

Evaluation of the joint protein-ligand complex structure further underscores these gains: \ourM{} reduces the mean RMSD from 4.20Å to 2.72Å. Taken together, these results demonstrate that \ourM{} not only improves absolute prediction accuracy but also captures the coupled conformational dynamics of protein-ligand interactions. This ability to model joint structural distributions under dynamic settings is critical for advancing molecular design and docking beyond static approximations.

\begin{figure}[htbp]
\centering
\begin{subfigure}[t]{0.48\textwidth}
    \includegraphics[width=0.32\textwidth, keepaspectratio]{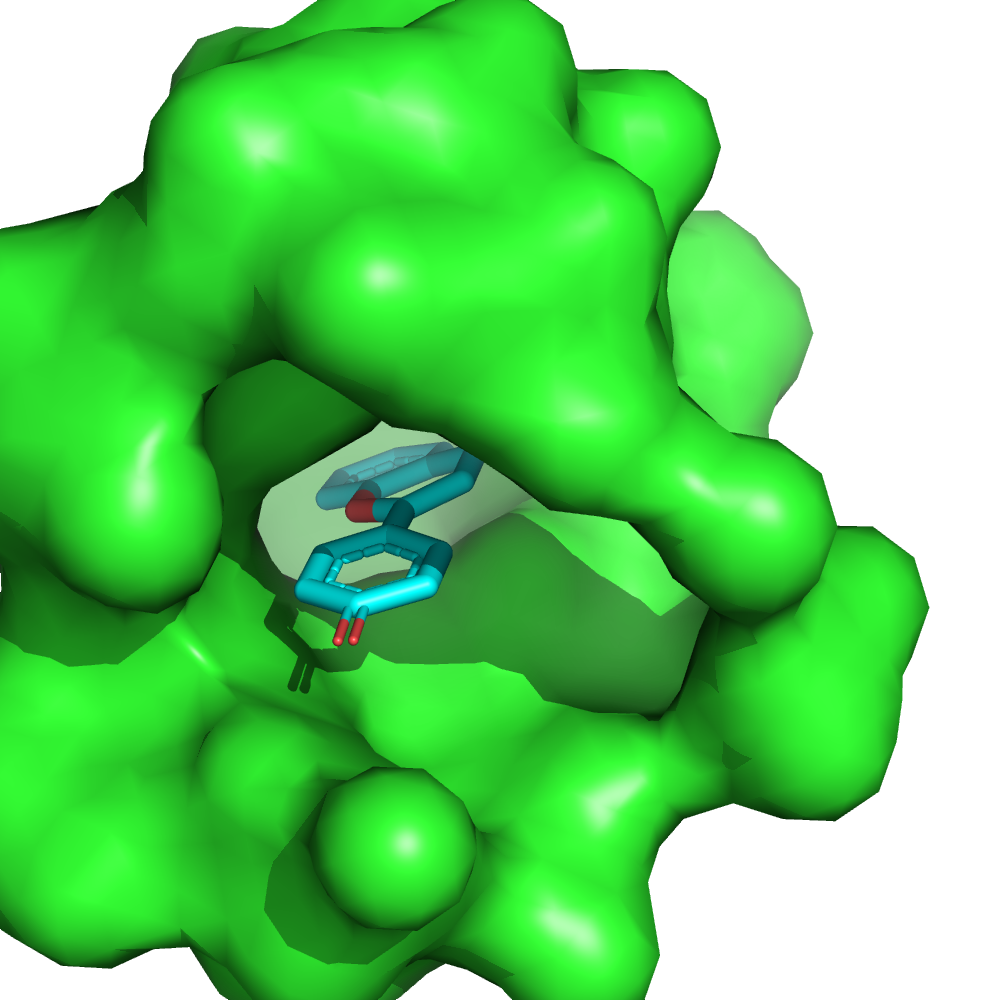}
    \includegraphics[width=0.32\textwidth, keepaspectratio]{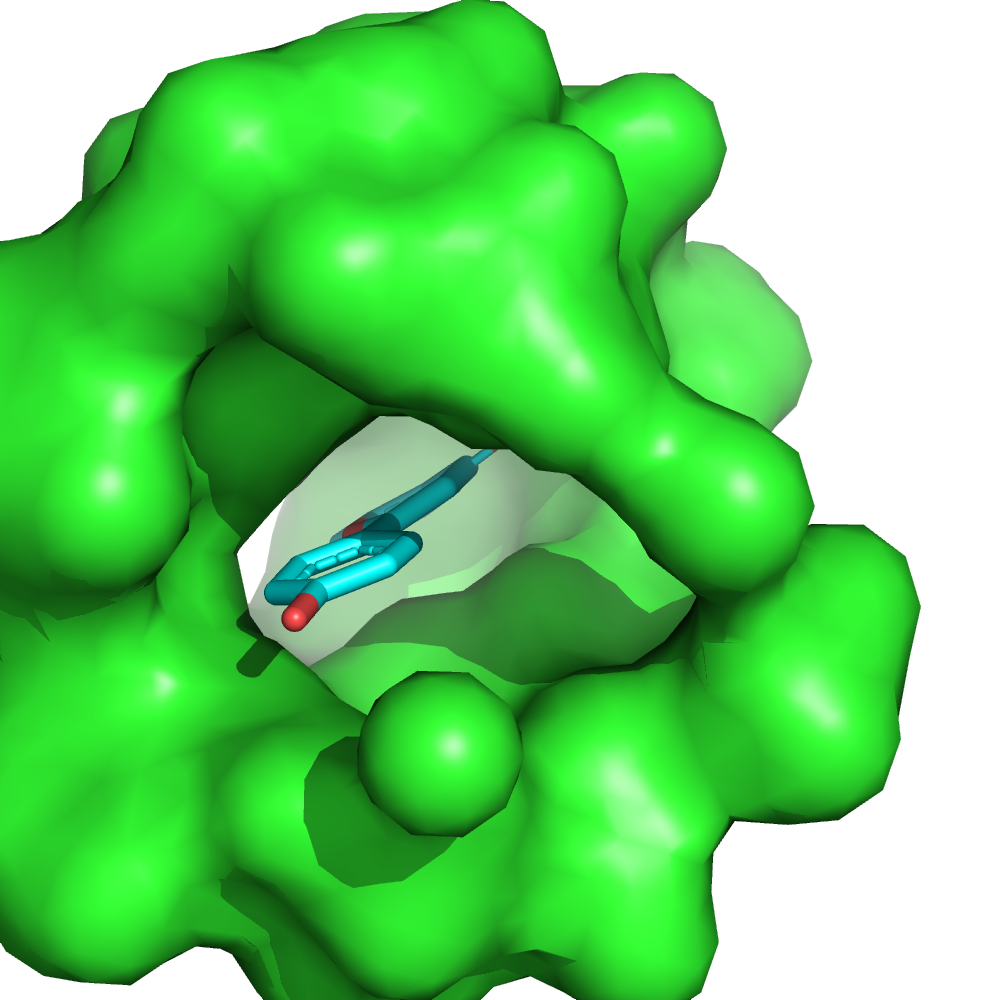}
    \includegraphics[width=0.32\textwidth, keepaspectratio]{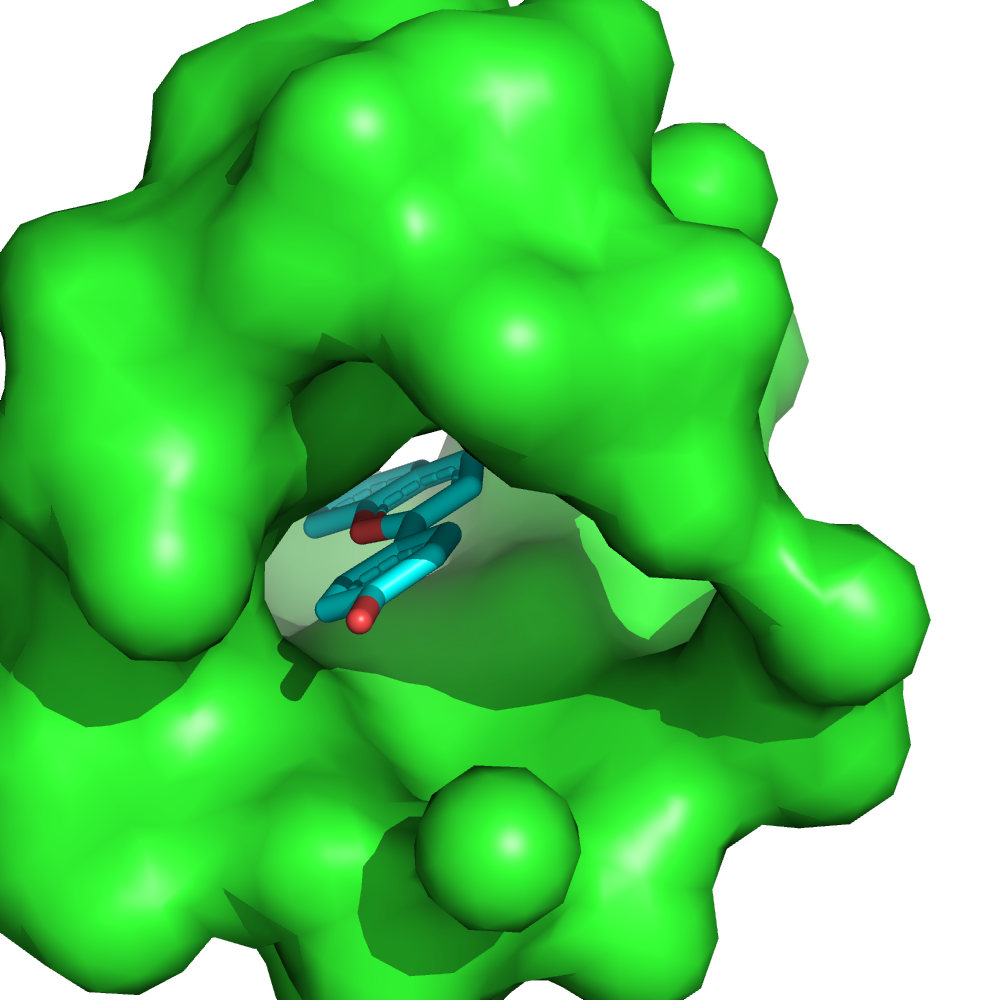}
    \caption{4KZQ Pocket given}
    \includegraphics[width=0.32\textwidth, keepaspectratio]{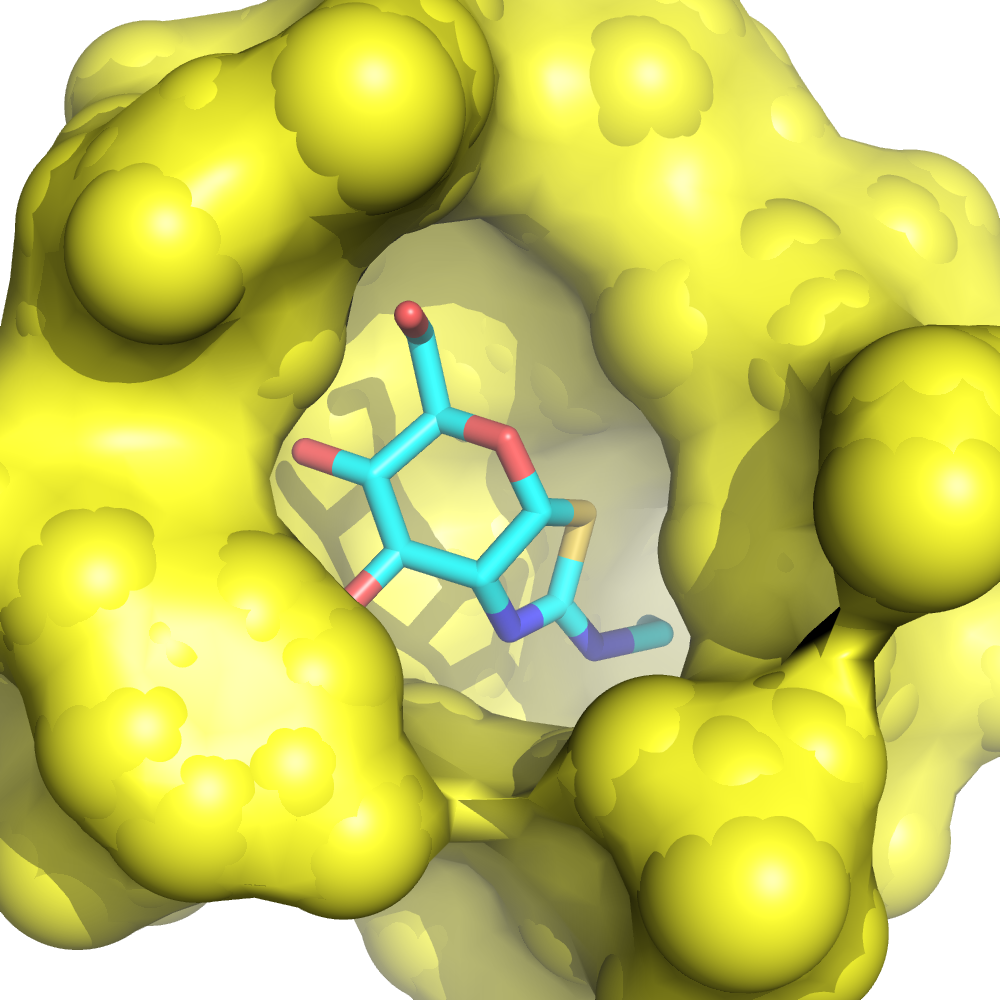}
    \includegraphics[width=0.32\textwidth, keepaspectratio]{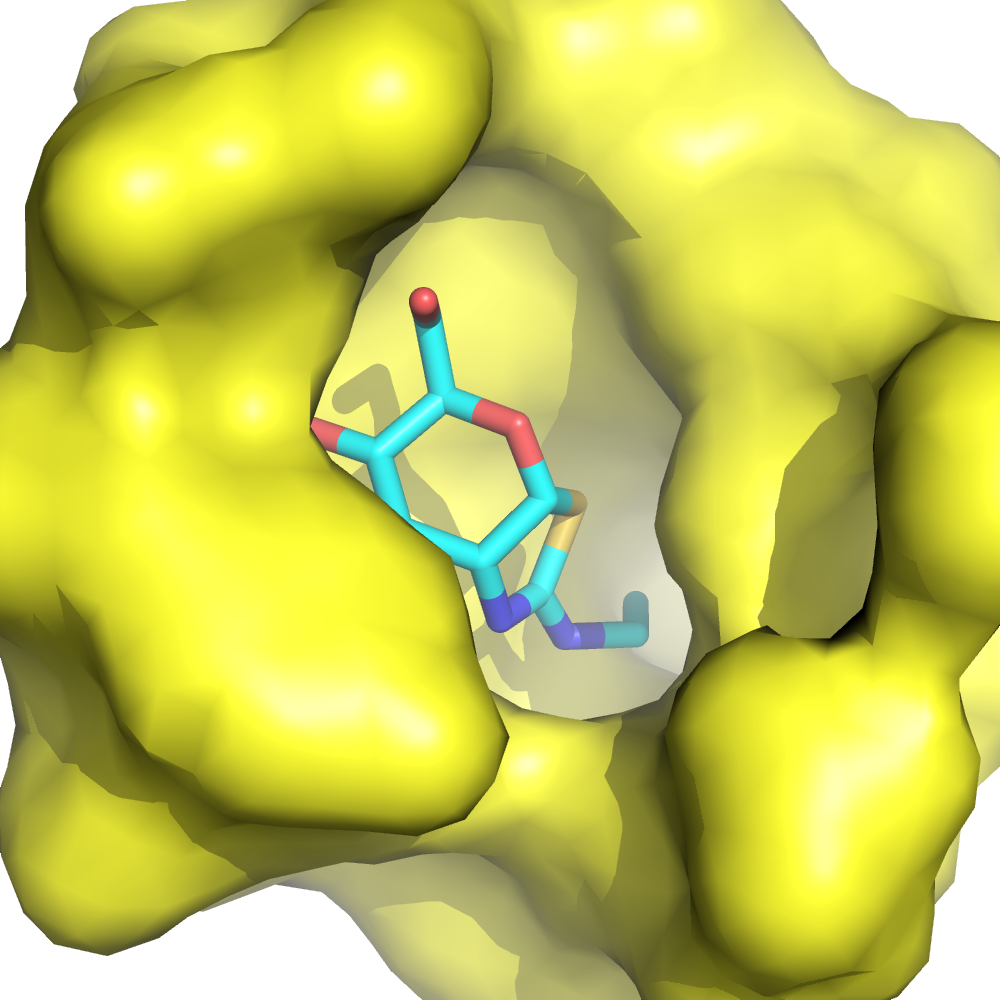}
    \includegraphics[width=0.32\textwidth, keepaspectratio]{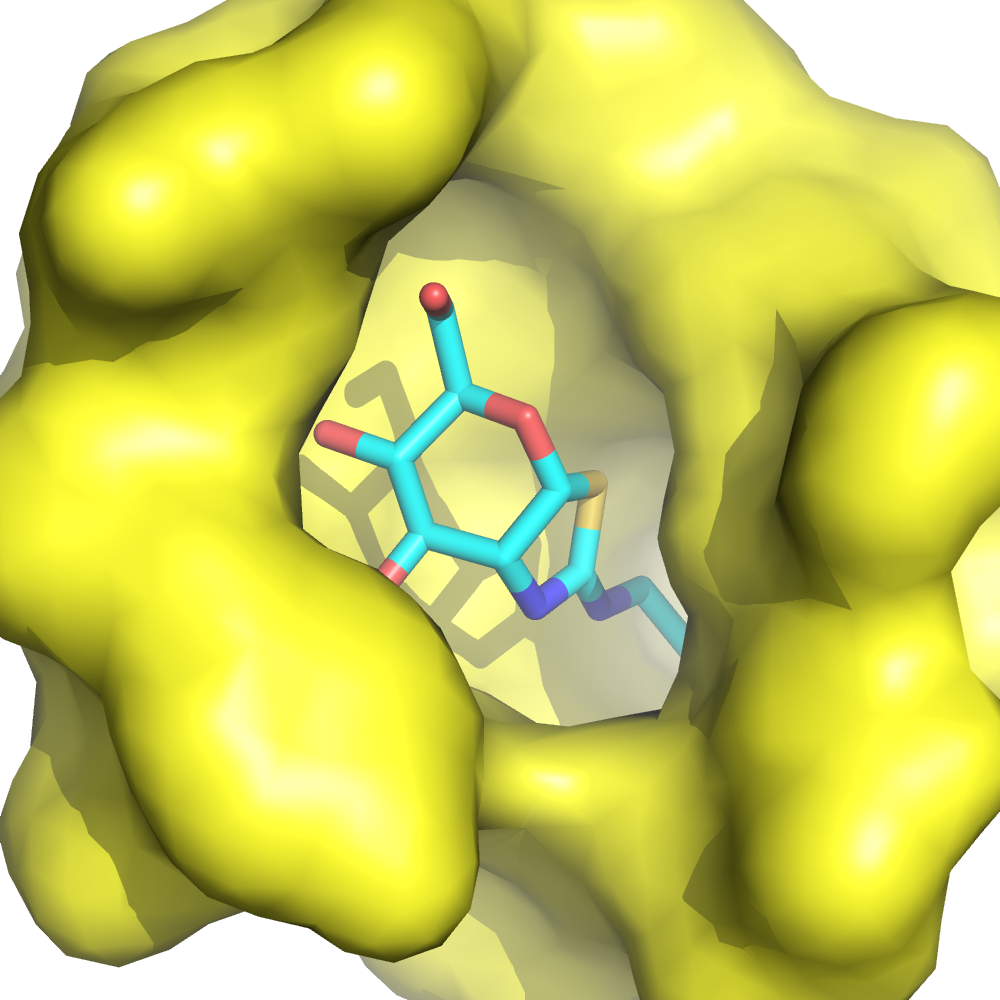}
    \caption{2VVN Pocket given}
\end{subfigure}
\hfill
\begin{subfigure}[t]{0.48\textwidth}
    \includegraphics[width=0.32\textwidth, keepaspectratio]{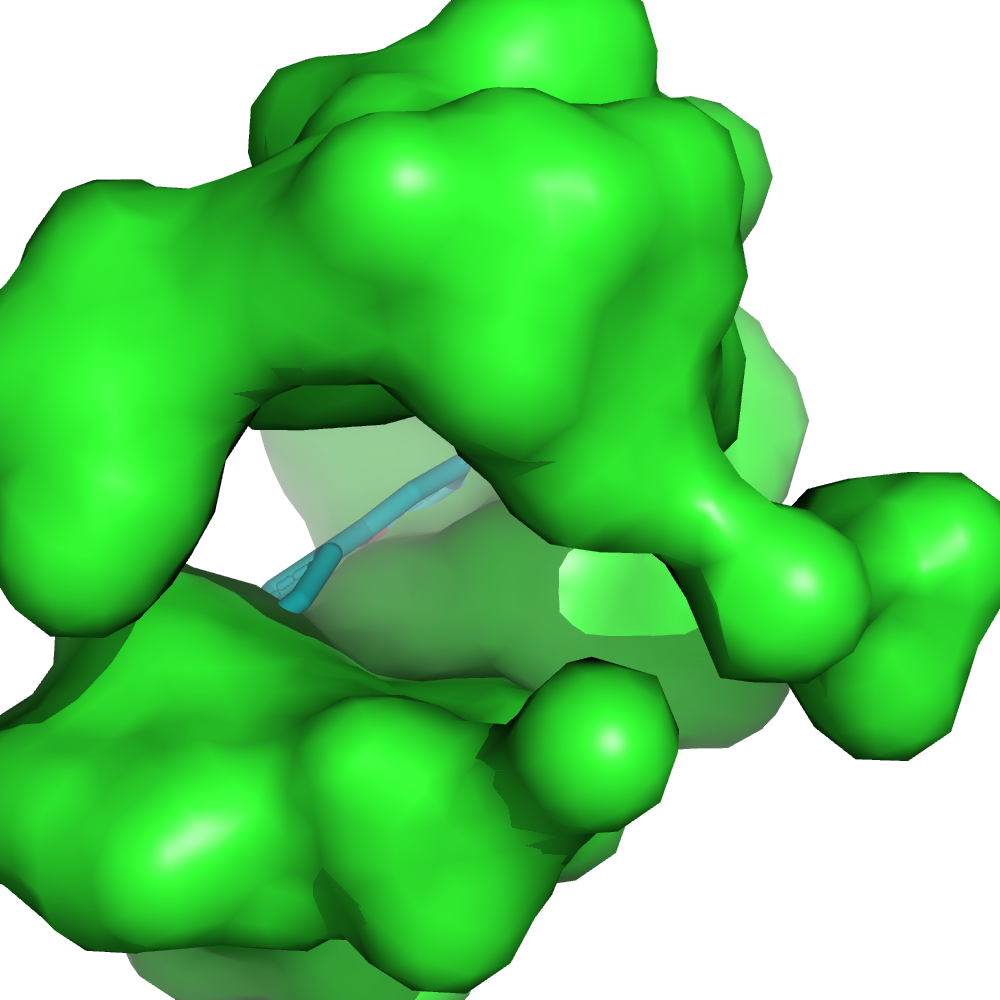}
    \includegraphics[width=0.32\textwidth, keepaspectratio]{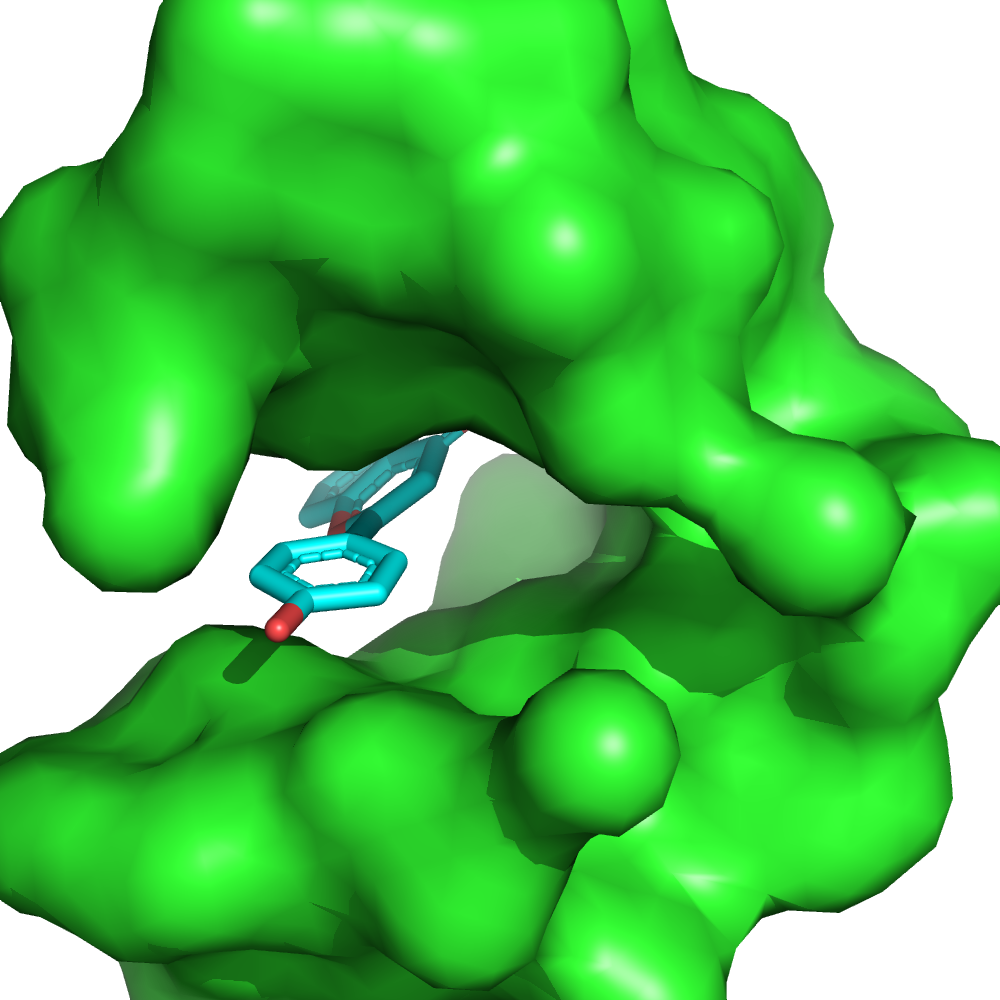}
    \includegraphics[width=0.32\textwidth, keepaspectratio]{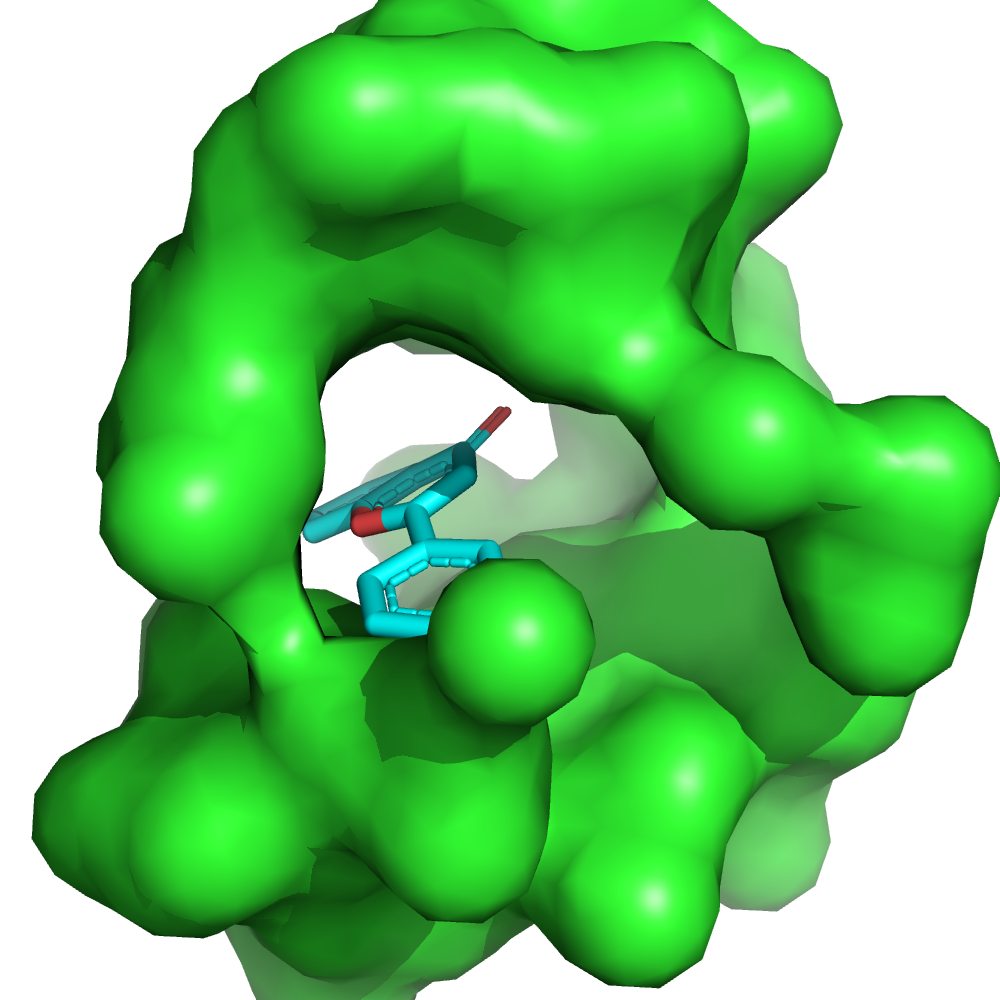}
    \caption{4KZQ Pocket not given}
    \includegraphics[width=0.32\textwidth, keepaspectratio]{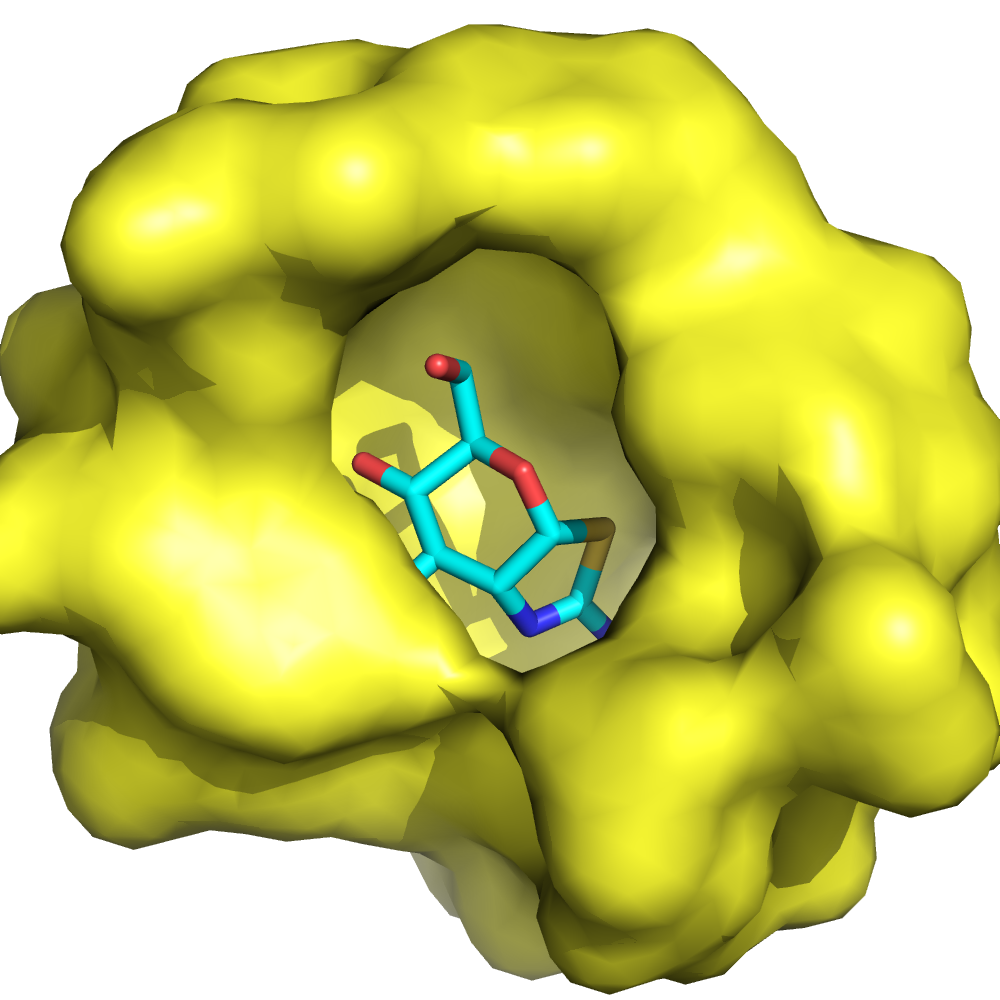}
    \includegraphics[width=0.32\textwidth, keepaspectratio]{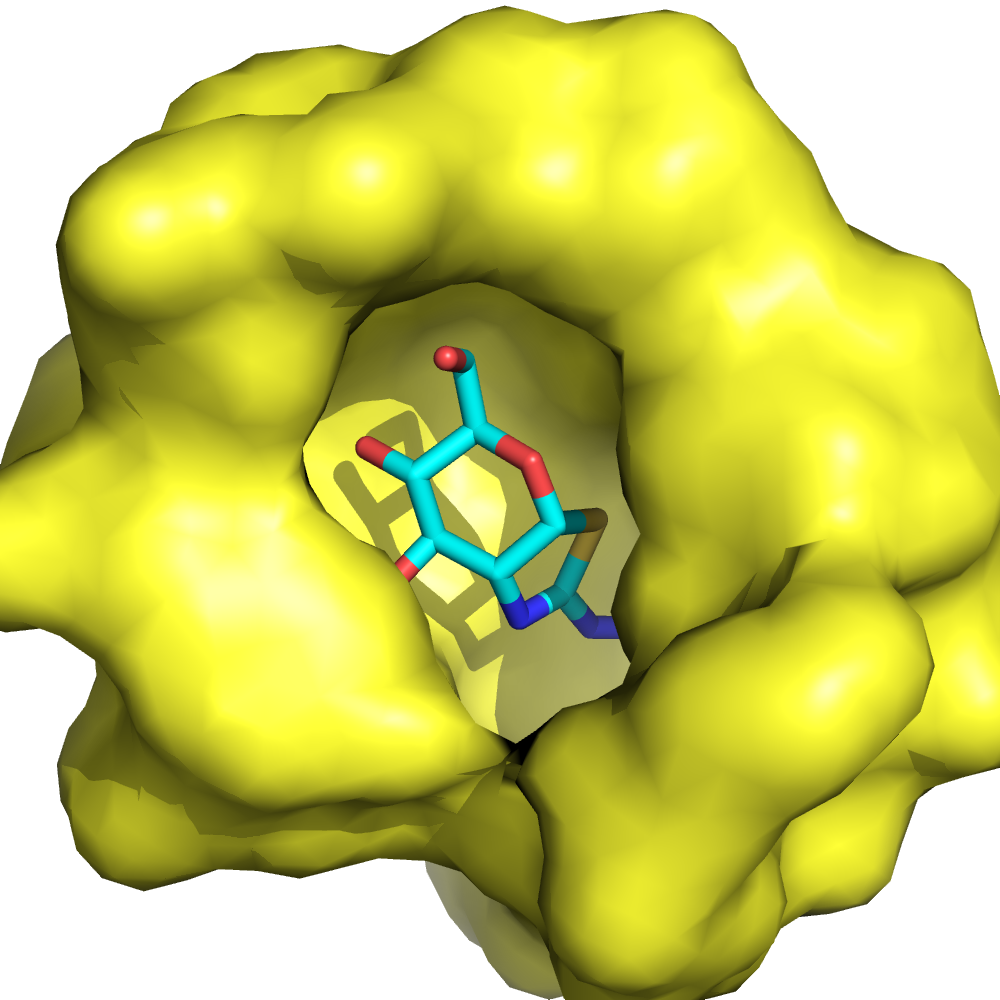}
    \includegraphics[width=0.32\textwidth, keepaspectratio]{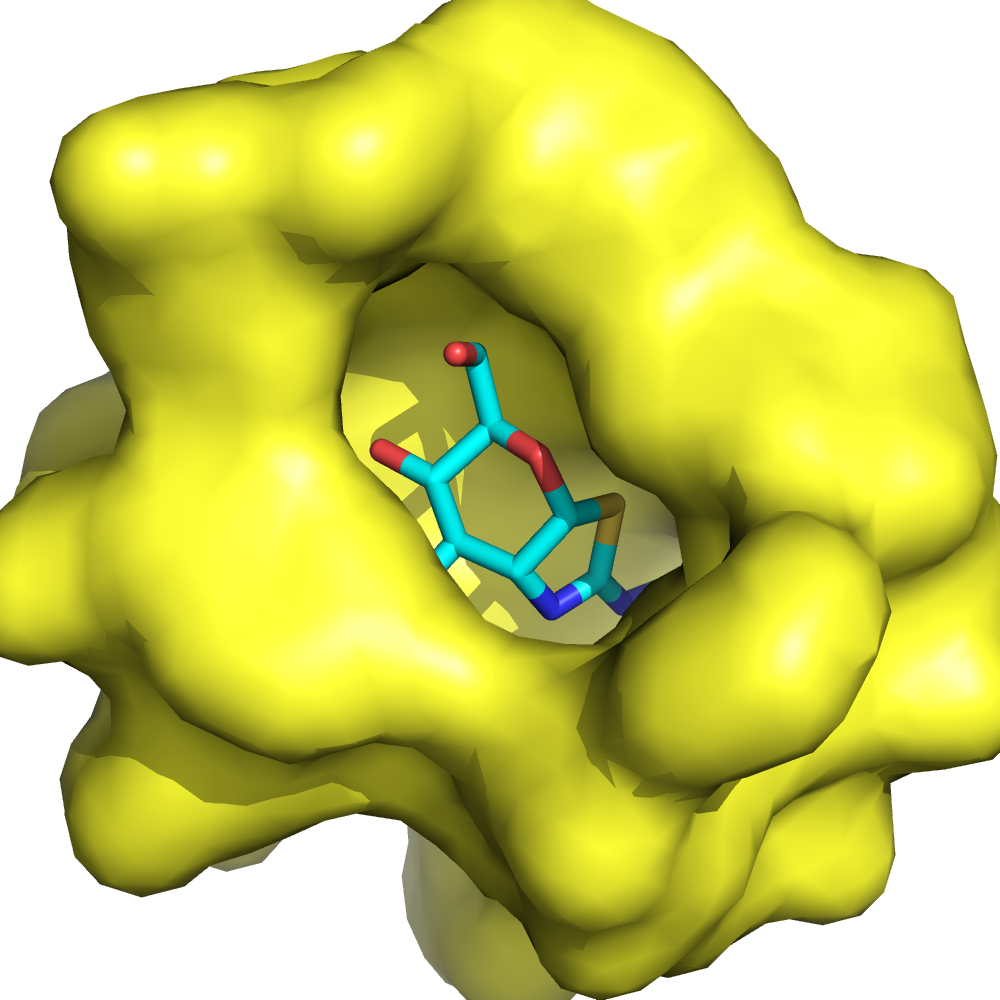}
    \caption{2VVN Pocket not given}
\end{subfigure}
\caption{Examples of the docking predictions.}
\label{fig:docking1}
\end{figure}

To further illustrate the qualitative fidelity of predicted structures, Figure~\ref{fig:docking1} shows representative examples for two protein-ligand complexes (PDB IDs: 4KZQ and 2VVN) under both modeling conditions—with and without the holo pocket provided. In both cases, the predicted ligand structures (shown in cyan) exhibit strong agreement with the ground truth conformations, even when the pocket geometry must be inferred. These visual results reinforce the model’s ability to recover realistic and physically plausible docking configurations, aligning well with the quantitative improvements described above.

\begin{table}[t]
    \caption{
        Evaluation of \ourM{} (100M) versus 3DMolFormer(100M) on dynamic pocket-ligand prediction. 
        RMSD percentiles, mean values, and the percentage of predictions within 2Å and 5Å are reported for pocket, ligand, and complex structures.
    }
    \vspace{-0.1cm}
    \centering
    \begin{tabular}{lcccccc}
    \toprule
    & \multicolumn{4}{c}{Percentiles $\downarrow$} & \multicolumn{2}{c}{\% Below $\uparrow$} \\
    \cmidrule(lr){2-5} \cmidrule(lr){6-7}
    \textbf{Method} & 25\% & 50\% & 75\% & Mean & 2\AA & 5\AA \\
    \midrule
    \multicolumn{7}{l}{\textit{3DMolFormer (100M)}\cite{hu20253dmolformer}} \\
    \quad Pocket + Ligand           & 3.03 & 3.82 & 4.66 & 4.20 & 1.27  & 82.17 \\
    \quad Pocket                    & 2.34 & 2.96 & 4.06 & 3.54 & 12.10 & 85.99 \\
    \quad Ligand (Pocket Gv)        & 2.18 & 2.84 & 3.63 & 2.97 & 18.47 & 95.54 \\
    \quad Ligand (Pocket Not Gv)    & 4.15 & 5.22 & 6.60 & 5.41 & 0.64  & 43.31 \\
    \midrule
    \multicolumn{7}{l}{\textit{\ourM{} (100M)}} \\
    \quad Pocket + Ligand           & 2.05 & 2.37 & 2.94 & 2.72 & 21.66 & 95.54 \\
    \quad Pocket                    & 1.51 & 1.75 & 2.14 & 2.12 & 69.43 & 95.54 \\
    \quad Ligand (Pocket Gv)        & 2.01 & 2.57 & 3.29 & 2.73 & 24.84 & 98.09 \\
    \quad Ligand (Pocket Not Gv)    & 2.10 & 2.81 & 3.51 & 2.90 & 15.92 & 96.82 \\
    \bottomrule
    \end{tabular}
    \label{tab:unigenx_vs_linear_eval}
\end{table}

\subsection{Unification Results of molecule and materials}

\begin{table}[htbp]
\centering
\small
\begin{tabular}{lllllll}
\hline & \multicolumn{2}{c}{ MP-20 } & \multicolumn{2}{c}{ Carbon-24 } & \multicolumn{2}{c}{ MPTS-52 } \\
\hline Methods & MR(\%)$\uparrow$ & RMSD(\textrm{\r{A}})  $\downarrow$ & MR(\%)$\uparrow$ & RMSD(\textrm{\r{A}})  $\downarrow$ & MR(\%)$\uparrow$ & RMSD(\textrm{\r{A}})  $\downarrow$ \\ 
\hline CDVAE\cite{xie2021crystal} & 33.90 & 0.1045 & 17.09 & 0.2969 & 5.34 & 0.2106 \\
\hline DiffCSP\cite{jiao2023crystal} & 51.49 & 0.063 & 17.54 & 0.2759 & 12.19 & 0.1786 \\
\hline FlowMM\cite{miller2024flowmm} & 61.39 & 0.057 & 23.47 & 0.4122 & 17.54 & 0.1726 \\
\hline \ourM{}(100M) & $\mathbf{64.74}$ & $\mathbf{0.042}$ & $\mathbf{29.46}$ & $\mathbf{0.2318}$ & $\mathbf{32.97}$ & $\mathbf{0.0838}$ \\
\hline
\end{tabular}
\caption{Results of the unification model on material tasks. MR represents Match Rate. \ourM{}'s performance surpasses the state-of-the-art, which proves its unified ability across diverse domains.}
\label{tab:unifmp}
\end{table}

\begin{table}[htbp]\centering\small  
\begin{tabular}{llllllllll}  
\hline  
Dataset & \multicolumn{8}{c}{Large-scale QM9} \\  
\hline  
Methods & \multicolumn{2}{c}{COV-P (\%) $\uparrow$} & \multicolumn{2}{c}{MAT-P $(\textrm{\AA}) \downarrow$} & \multicolumn{2}{c}{COV (\%) $\uparrow$} & \multicolumn{2}{c}{MAT $(\textrm{\AA}) \downarrow$} \\  
& Mean & Median & Mean & Median & Mean & Median & Mean & Median \\  
\hline  
ConfGF\cite{shi2021learning} & 46.23 & 44.87 & 0.5171 & 0.5133  & 89.21 & 95.12 & 0.2809 & 0.2837\\  
GeoMol\cite{ganea2021geomol} & 78.28 & 81.03 & 0.3790 & 0.3861 & 91.05 & 95.55 & 0.2970 & 0.2993 \\  
DMCG\cite{zhu2022direct} & 90.86 & 95.36 & 0.2305 & 0.2258 & $\mathbf{98.34}$ & $\mathbf{100}$ & 0.1486 & 0.1340 \\  
\ourM{}(100M)& $\mathbf{91.43}$ & $\mathbf{100.00}$ & $\mathbf{0.2186}$ & $\mathbf{0.0818}$  & 92.53 & $\mathbf{100}$ & $\mathbf{0.1441}$ & $\mathbf{0.0818}$ \\
\hline  
\end{tabular}  
\caption{Results of the unification model on the small molecular task of GEOM-QM9 using COV-P, MAT-P, COV, and MAT metrics. COV-P, MAT-P, COV, and MAT evaluate molecule coverage and accuracy, as defined by Equations~\eqref{cov}--\eqref{mat-p}.}  
\label{tab:unifqm9}  
\end{table}

The finetuned results for MP-20, Carbon 24, MPTS-52 and GEOM-QM9 results can be found in Table~\ref{tab:unifmp} and Table~\ref{tab:unifqm9}. The results show that a 100M unification model \ourM{}(100M) still performed better results than the SOTA methods on most indicators. With larger models and more data, our model can scale to achieve better results due to the scalability of the auto-aggressive and diffusion models. Additionally, due to the inclusion of a word loss head, our model is adaptable for language training, making it flexible for general tasks such as language-based design and generation.

\subsection{Intergrating with LLM Results}
\begin{table}[htbp]
\centering
\small
\begin{tabular}{lll}
\hline & \multicolumn{2}{c}{ MP-20 }  \\
\hline Methods & MR(\%)$\uparrow$ & RMSD(\textrm{\r{A}})  $\downarrow$ \\
\hline NatureLM\cite{xia2025naturelm} & 61.78&0.044\\
\hline \ourM{}-NL(1B) & $\mathbf{62.82}$ & $\mathbf{0.042}$\\
\hline 
\end{tabular}
\caption{Results of \ourM{} with capability of natural langauge on MP-20.}
\label{tab:nl}
\end{table}
As shown in Table~\ref{tab:nl}, our model \ourM{}-NL(1B), which incorporates natural language understanding capabilities, not only preserves but also improves upon the performance of prior models such as NatureLM. Specifically, it achieves a higher matching rate (MR) of 63.10\% and a lower RMSD of 0.040 \r{A}, outperforming NatureLM on both metrics.

Table~\ref{tab:gate_qm9} show that comparing with \ourM{}(100M), \ourM{}-NL(1B) also performed better on QM9.

These results highlight the effectiveness of our approach in bridging linguistic understanding with structural prediction tasks. This also means that we can greatly enhance the model's flexibility while maintaining its performance.

\begin{table}[htbp]\centering\small  
\begin{tabular}{llllllllll}  
\hline  
Dataset & \multicolumn{8}{c}{Large-scale QM9} \\  
\hline  
Methods & \multicolumn{2}{c}{COV-P (\%) $\uparrow$} & \multicolumn{2}{c}{MAT-P $(\textrm{\AA}) \downarrow$} & \multicolumn{2}{c}{COV (\%) $\uparrow$} & \multicolumn{2}{c}{MAT $(\textrm{\AA}) \downarrow$} \\  
& Mean & Median & Mean & Median & Mean & Median & Mean & Median \\  
\hline   
\ourM{}(100M)&91.43 & $\mathbf{100.00}$ & 0.2186 & $\mathbf{0.0818}$  & 92.53 & $\mathbf{100}$ & 0.1441 & 0.0818 \\
\hline  
\ourM{}--NL(1B)&$\mathbf{92.30}$&$\mathbf{100.00}$&$\mathbf{0.1518}$&0.0886&$\mathbf{92.88}$&$\mathbf{100}$&$\mathbf{0.1368}$&$\mathbf{0.0751}$ \\
\hline
\end{tabular}  
\caption{Comparison between 100M model and Results of \ourM{} with capability of natural langauge on QM9.}  
\label{tab:gate_qm9}  
\end{table}

The results demonstrate that our model retains its linguistic capabilities without compromising accuracy. The synergistic effect of combining linguistic and numerical reasoning opens the door for developing unified models that can jointly process textual descriptions and scientific data. This paves the way for future applications such as text-guided molecule design, multimodal scientific assistants, and interactive generative systems in computational chemistry and materranh1

\section{Conclusion}
This work introduces \ourM{}, a novel framework for unified generation of symbolic and numerical scientific data. \ourM{} integrates an autoregressive model with a conditional diffusion-based generative head, leveraging the autoregressive model's sequence flexibility and the diffusion model's numerical precision. A key innovation is a sequentialization scheme that represents diverse scientific data---formulas, coordinates, energies, forces---as a single token sequence. This enables next-token prediction with the autoregressive component, while the diffusion head addresses numerical precision limitations by operating in continuous space and training efficiently on low-dimensional variables. \ourM{} achieves significant performance improvements over state-of-the-art diffusion/flow matching models in material generation (10--120\% gains on MP-20, Carbon-24, and MPTS-52), surpasses NatureLM, and establishes new state-of-the-art results on most targets in GEOM-QM9 prediction, de novo design on materials, and conditional generation on QM9. Demonstrating high capacity for unified training across domains and natural language prompt generation, \ourM{} avoids explicit inductive biases for equivariance/invariance, relying on data augmentation to learn these properties, thus preserving transformer scalability. Future work will extend \ourM{} to other scientific data types---proteins, DNA---and tasks---energy/force prediction, aiming towards a unified foundation model for scientific generation, potentially integrating with general domain models.

\clearpage
\bibliography{reference}
\bibliographystyle{unsrt}

\clearpage
\appendix

\section{Auto-regressive model, diffusion model and flow matching}

\subsection{Autoregressive Model}
Auto-regressive models have become a cornerstone in deep learning for sequential data modeling. These models operate by factorizing the joint probability distribution of a sequence into a product of conditional probabilities, allowing each token or data point to be predicted based on the preceding ones. Formally, given a sequence $$\mathbf{x} = {x_1, x_2, \ldots, x_T},$$ the auto-regressive approach models the probability as
$$P(\mathbf{x}) = \prod_{t=1}^{T} P(x_t \mid x_1, x_2, \ldots, x_{t-1}), $$
where $x_i$ is the $i$-th token or element in the sequence. This property makes them highly effective for tasks such as language modeling, where predicting the next word in a sentence requires contextual understanding of the preceding words. Popular examples include models like the GPT series~\cite{achiam2023gpt} and LLaMA~\cite{touvron2023llama,touvron2023llama2,grattafiori2024llama}, which primarily utilize multi-head attention-based Transformer~\cite{vaswani2017attention} decoder architectures to capture long-range dependencies in data.

\subsection{Diffusion Models}
\subsubsection{Diffusion Process}
A diffusion model is defined by a diffusion process $\{x_t\}_{t=0}^1$, starts with $x \sim p_0$, and evolves to $x_1 \sim p_1$, where $p_1$ has a tractable form for efficient sample generation. The dataset of independent samples from the data distribution is available. Let $p(x_t)$ denote the probability density of $x_t$, and $p\left(x_t \mid x\right)$ represent the transition kernel from $x$ to $x_t$, with $0 < t \leq 1$. The diffusion process is modeled as the solution to a stochastic differential equation (SDE)~\cite{song2020denoising}:
\begin{equation}
\mathrm{d} x_t = f(x_t, t) \mathrm{d}t + g(t) \mathrm{d} \boldsymbol{B}_t,
\label{eq:sde}
\end{equation}
where $\boldsymbol{B}_t$ is a standard Wiener process, $f(\cdot, t): \mathbb{R}^d \rightarrow \mathbb{R}^d$ is the drift coefficient of $x(t)$, and $g(\cdot): \mathbb{R} \rightarrow \mathbb{R}$ is the diffusion coefficient of $x(t)$.

By reversing the diffusion process, starting from samples $x_1 \sim p_1$, where $p_1$ is usually a Gaussian distribution, it is possible to recover samples $x \sim p_0$. The reverse of a diffusion process is also a diffusion process, governed by the reverse-time SDE~\cite{song2020score}:
\begin{equation}
\mathrm{d} x_t = \left[f(x_t, t) - g(t)^2 \nabla_{x_t} \log p_t(x_t)\right] \mathrm{d}t + g(t) \mathrm{d} \overline{\boldsymbol{B}},  
\label{eq:inverse}
\end{equation}
where $\overline{\boldsymbol{B}}$ is another standard Wiener process, i.e., Brownian motion from $t=1$ to $t=0$. 

As the diffusion process is a stochastic process with the Brownian motion stochastic term, the SDE \eqref{eq:sde} of the forward process implies the following relationship 
\begin{equation}
    p(x_t) = \int_{x} p(x_t|x)p(x)d x,
\end{equation}
where the conditional probability $p(x_t|x) = C \exp(-\frac{\|x_t-\alpha_t x\|^2}{\sigma_t^2})$, and $C$ is the normalization constant satisfying $\int_{x_t}C \exp{-\frac{\|x_t-\alpha_t x\|^2}{\sigma_t^2}}dx_t=1$.
Equivalently, we have
\begin{equation}
\label{eq:noise}
    x_t = \alpha_t x + \sigma_t \epsilon.
\end{equation}
Here $\alpha_t$ and $\sigma_t$ are linear and variance variable, depending on the time $t$; $\epsilon$ is a random variable. In the diffusion model, $\alpha_t$ and $\sigma_t$ are given by designed schemes for VE and VP cases~\cite{song2020score}; the random variable $\epsilon$ is a standard Gaussian noise variable, i.e., $\epsilon\sim N(0, I)$.  In our work, we use the Denoising Diffusion Probabilistic Model (DDPM) setting for training and the VP scheme for sampling~\cite{ho2020denoising, song2020denoising}. We also test the diffusion choice of Elucidated Diffusion Model (EDM)~\cite{karras2022elucidating}. We obtained similar performance leveraging DDPM and EDM respectively, and listed the results as an ablation study in Appendix~\ref{sec:edm}.

\subsubsection{Diffusion Targets}
Different from conventional supervised learning, the training targets in the training loss of the diffusion model are not obvious most of the time. With the same noising process, the training targets differ between different models, such as $\epsilon$ model~\cite{ho2020denoising}, denoising model~\cite{kingma2021variational}, and score model~\cite{song2019generative}. 

Here we would show that all of these targets have connected forms.
Recall that adding noise by $x_t = \alpha_t x + \sigma_t \epsilon$, i.e.,
\begin{equation}\label{eq.cond}
    p(x_t|x) = Ce^{-\frac{-\|x_t-\alpha_t x\|^2}{2\sigma_t^2}}.
\end{equation}
By the definition of the score function $s(x_t) \equiv \nabla_{x_t}\log p(x_t)$ and through a direct calculation, we have
\begin{equation}
s(x_t) = \frac{\int_x \frac{\nabla_{x_t} p(x_t|x)}{p(x_t|x)}p(x_t|x) p(x)dx}{p(x_t)} = \int_x \nabla_{x_t} \log p(x_t|x)p(x|x_t)dx, 
\end{equation}
where we finally obtain
\begin{equation}
s(x_t) = \mathbb{E}_x[\nabla_{x_t} \log p(x_t|x)| x_t]=\mathbb{E}_x[-\frac{x_t-\alpha_t x}{\sigma_t^2}| x_t] = -\frac{x_t-\alpha_t \mathbb{E}[x|x_t]}{\sigma_t^2}.
\label{eq:connect}
\end{equation}

By Theorem~\ref{thm:conditional_expect}, the score of a distribution, $\nabla_{x_t} \log p_t(x_t)$, can be estimated by training a score-based model $s_{\boldsymbol{\theta}}$ using score matching~\cite{song2019generative} with the training target $\nabla_{x_t} \log p(x_t|x)$. To approximate $\nabla_{x_t} \log p(x_t)$, a time-dependent score-based model $s_{\theta}(x_t, t)$ can be trained using a continuous generalization of the score matching objective~\cite{song2020score}:
\[
\begin{aligned}
{\theta}^* = \underset{{\theta}}{\arg \min} \, \mathbb{E}_{t\sim U[0, 1]} \bigg\{ \lambda(t) \, \mathbb{E}_{x} \, \mathbb{E}_{x_t \mid x} \big[\|{s}_{{\theta}}(x_t, t) - \nabla_{x_t} \log p(x_t \mid x)\|_2^2\big] \bigg\},
\end{aligned}
\]
where $\lambda: [0, 1] \rightarrow \mathbb{R}^+$ is a weighting function, $U[0, 1]$ is a uniform distribution in $[0, 1]$, $x \sim p_0$ and $p_0$ is the data distribution. 

From Eq.~\eqref{eq:connect}, one can also approximate the training target with a different model, such as the denoising model $x_\theta$ with the target $x$~\cite{kingma2021variational}, and the $\epsilon$ model $\epsilon_\theta$~\cite{ho2020denoising} with the target $\frac{x_t-\alpha_t x}{\sigma_t} = \epsilon$, with a similar training loss as the score model. Hence, the optimal solution of the models (denoted as $\theta^*$) approaches:
\begin{equation}
    s_{\theta^*} \to \mathbb{E}_x[\nabla_{x_t} \log p(x_t|x)| x_t]; \quad \epsilon_{\theta^*}\to \mathbb{E}_x[\frac{x_t-\alpha_t x}{\sigma_t}| x_t]; \quad x_{\theta^*}\to \mathbb{E}_x[x|x_t].
\end{equation}
These models connect to each other in the optimal results given by Eq. \eqref{eq:connect}, and have the following forms
\begin{equation}\label{eq:scorec}
\begin{aligned}
    s_{\theta^*}(x_t) &= -\frac{x_t-\alpha_t x_{\theta^*} (x_t)}{\sigma_t^2}\\
    s_{\theta^*}(x_t) &= -\frac{\epsilon_{\theta^*}(x_t)}{\sigma_t}.
\end{aligned}
\end{equation}
Therefore, when training diffusion models using an $\epsilon$ model or denoising model, the sampling process can be aligned with score-based models via Eq.~\eqref{eq:scorec} and utilize the reverse stochastic differential equation~\eqref{eq:inverse}. In this work, we adopt the $\epsilon$ model proposed in DDPM~\cite{ho2020denoising} for training and employ the VP scheme~\cite{song2020score} for sampling.

\subsection{Connections of the Training Targets}\label{sec:target}
When the random variable $z$ is a standard Gaussian random variable, i.e., $z = \epsilon$, we can express the connection between the diffusion model training targets and the training target $v$ in flow matching.
From the property of conditional expectation, $x_t = \mathbb{E}[x_t|x_t]$, one can obtain
\begin{equation}
\label{eq:xt}
    x_t = \mathbb{E}[x_t|x_t] = \mathbb{E}[\alpha_t x+ \sigma_t \epsilon|x_t] = \alpha_t \mathbb{E}[x|x_t] + \sigma_t \mathbb{E}[\epsilon|x_t].
\end{equation}
From diffusion models, $x_{\theta^*}=\mathbb{E}[x|x_t]$ gives the training target of the denoising model, and $\epsilon_{\theta^*}=\mathbb{E}[\epsilon|x_t]$ gives the training target of the $\epsilon$ model. From Eq. \eqref{eq:connect}, we have
\begin{equation}
    \label{eq:score}
    s(x_t) = -\frac{x_t-\alpha_t \mathbb{E}[x|x_t]}{\sigma_t^2}.
\end{equation}
A direct calculation according to Eq. \eqref{eq:v}, \eqref{eq:xt}, and \eqref{eq:score} yields the connections of the diffusion targets $s_{\theta^*}$, $\epsilon_{\theta^*} = \mathbb{E}[\epsilon|x_t]$, and $x_{\theta^*} = \mathbb{E}[x|x_t]$ with the flow matching target $v$ as
\begin{equation}
    \begin{aligned}
        s_{\theta^*}&=\frac{\alpha_t v-\dot{\alpha}_t x_t}{\dot{\alpha}_t \sigma_t-\alpha_t \dot{\sigma}_t}\sigma_t^{-1};\\
        \epsilon_{\theta^*} &= \frac{\dot{\alpha}_t x_t - \alpha_t v}{\dot{\alpha}_t \sigma_t - \alpha_t\dot{\sigma}_t};\\
        x_{\theta^*} &= \frac{\dot{\sigma}_t x_t - \sigma_t v}{\dot{\sigma}_t\alpha_t - \sigma_t\dot{\alpha}_t}.
    \end{aligned}
\end{equation}

\subsection{Diffusion model and flow matching}

The training targets for diffusion and flow matching models share the same mathematical structure, as stated by Theorem \ref{thm:conditional_expect}~\cite{evans2012introduction}.
\begin{theorem}
\label{thm:conditional_expect}
    Let $X$ be an integrable random variable. Then, for each $\sigma$-algebra $\mathcal{V}$ and $Y \in \mathcal{V}$, $Z = \mathbb{E}(X| \mathcal{V})$ solves the least squares problem
    $$\min_{Z_\theta\in \mathcal{V}} \| Z_\theta - X\|\,,$$
    where $\| X\| = \left( \int X^2 \, dP \right)^{\frac{1}{2}}$.
\end{theorem}

Diffusion models and flow matching share the same noising procedure, given by Eq. \eqref{eq:noise}, i.e.,
\begin{equation}
    x_t = \alpha_t x + \sigma_t z,
\end{equation}
where $x \sim p_0$, $p_0$ is the data distribution, $z\sim p_{prior}$, $p_{prior}$ is a prior distribution from which samples can be easily drawn. In diffusion models, the prior random variable is a standard Gaussian distribution, i.e., $z = \epsilon \sim N(0, I)$, thus limiting the diffusion process to approach Gaussian noise from the data variable. However, flow matching extends the prior to any distribution, making it a flow from samples of the given distribution to the generated data samples~\cite{lipman2022flow, lipman2024flow}. Rectified flow and stochastic interpolation can also be obtained through similar approaches~\cite{liu2022flow, ma2024sit, albergo2023stochastic}. A typical training target is to find the velocity that connects the prior and data distributions, $\frac{dx_t}{dt} = v$. Hence, the training target is directly derived from Eq. \eqref{eq:noise} as
\begin{equation}
    \frac{dx_t}{dt} = \dot{\alpha}_t x + \dot{\sigma}_t z,
\end{equation}
with the neural network $v_\theta(x_t)$ approximating the velocity $v$ of the flow by
\begin{equation}
\label{eq:v}
    v = \mathbb{E}\left[\frac{dx_t}{dt}\middle|x_t\right]=\mathbb{E}\left[\dot{\alpha}_t x + \dot{\sigma}_t z \middle| x_t\right] = \dot{\alpha}_t \mathbb{E}\left[ x \middle|x_t\right] + \dot{\sigma}_t \mathbb{E}\left[ z \middle| x_t\right].
\end{equation}
The score target, denoising model target, and $\epsilon$ target for flow matching with the Gaussian noise can also be derived from Eq. \eqref{eq:connect} and direct calculation, as discussed in Appendix \ref{sec:target}. In this work, we primarily focus on the combination of autoregressive models and diffusion models, as the combination with flow matching is similar to that with diffusion models. Therefore, this investigation is left for future discussion.

\section{Pseudocode}
We provide the overall algorithm pseudocode, which includes the forward and sample processes for the model, as well as the forward and sample processes for diffloss. The algorithms are listed as follows.

\begin{algorithm}[htbp]
   \caption{\ourM{} Forward Procedure}
   \label{alg:\ourM{}train}
    \begin{algorithmic}[1]
        \Require    
            \Statex $w_{gt}$: Ids of input tokens in the input sentences
            \Statex $v_{gt}$: Numeral values in the input sentences
            \Statex $m_{val}$: Masks for values
            \Statex $m_{pad}$: Padding Masks
            \Statex $n$: Multiplication times of DiffLoss
            \Statex $w$: Loss weight
        \State{Initialize $x_{input}$ in the shape of $w_{gt}$}
        \State \textbf{\# Embed the inputs}
        \State $x_{input}[\neg m_v] \leftarrow$ \texttt{Embedding}($w$)
        \State $x_{input}[m_v] \leftarrow$ \texttt{Linear}($v$)
        \State \textbf{\# Pass through the AR Model}
        \State $h_{rep} \leftarrow$ \texttt{Transformer\_Decoder}($x_{input}, m_{pad}$)
        \State $w_{logits} \leftarrow$ \texttt{Linear}($h_{rep}[\neg m_{val}]$)
        \State \textbf{\# Compute the diffloss and wordloss}
        \State $l_d \leftarrow $ \texttt{\hyperref[alg:difflosstrain]{DiffLoss}}($v_{gt},h_{rep}[:-1][m_{val}], n$) 
        \State  $l_w \leftarrow $ \texttt{CrossEntropyLoss}($w_{gt}, w_{logits}[:-1] $)
        \State {\bfseries return} $l_d + w * l_w$
    \end{algorithmic}
\end{algorithm}

\begin{algorithm}[htbp]
   \caption{DiffLoss Forward Procedure}
   \label{alg:difflosstrain}
    \begin{algorithmic}[1]
        \Require    
            \Statex $v_{gt}$: Ground truth numeral values 
            \Statex $h_{rep}$: Hidden representations for conditional diffusion
            \Statex $M$: Times of Multiplication of DiffLoss
        \State {Initialize \texttt{net}$ \leftarrow$ \texttt{SimpleMLPAdaLN()}\texttt{, diff}$ \leftarrow$ \texttt{GaussianDiffusion()}}
        \State \textbf{\# Apply DiffMul}
        \State $v \leftarrow v_{gt}.$\texttt{repeat($M$)}
        \State $h \leftarrow h_{rep}.$\texttt{repeat($M$)}
        \State \textbf{\# Sample Different Timesteps for Diff}
        \State $T \leftarrow \texttt{diff.num\_timesteps}$
        \State $t \leftarrow \texttt{Randint}(0, T, (h_{rep}\texttt{.shape[0]},))$ 
        \State \textbf{\# Apply Diffusion }
        \State $l \leftarrow \texttt{diff.\hyperref[alg:diffloss]{trainingloss}}(\texttt{net}, v_{gt}, t, h_{rep})$
        \State{\bfseries return} $l$
    \end{algorithmic}
\end{algorithm}

\begin{algorithm}[htbp]
    \caption{GaussianDiffusion Training Loss}
    \label{alg:diffloss}
     \begin{algorithmic}[1]
         \Require    
             \Statex \texttt{net}: Neural network for predicting noise and variance
             \Statex $x_0$: Hidden representations for conditional diffusion
             \Statex $t$: Timesteps for Diffusion
             \Statex $c$: Condition for Diffusion
        \State {Sample $\epsilon$ from $\mathcal{N} (\bf{0}, \bf{I} )$}
        \State $x_t \leftarrow \sqrt{\overline{\alpha}_t}x_0 + \sqrt{1-{\overline{\alpha}_t}}\epsilon$
        \State \textbf{\# Predict Noise and Variance using }
        \State $\hat{\epsilon}, \hat{\sigma} = \texttt{net}(x_t, t, c).\texttt{split}()$
        \State \textbf{\# Compute L2 loss and VLB loss }
        \State $L_{mse} \leftarrow \texttt{L2Loss}(\hat{\epsilon},\epsilon)$
        \State $L_{vlb} \leftarrow \texttt{VLBLoss}(\hat{\epsilon}.\texttt{detatch()}, \hat{\sigma}, x_t, t)$
        \State{\bfseries return} $\texttt{mean\_flat}(L_2 + L_{vlb})$
     \end{algorithmic}
\end{algorithm}

\begin{algorithm}[htbp]
    \caption{\ourM{} Sampling Procedure}
    \label{alg:\ourM{}sample}
    \begin{algorithmic}[1]
        \Require    
            \Statex $w_{in}$: Ids of input tokens in the initial sentences 
            \Statex $v_{in}$: Numeral values in the initial sentences 
            \Statex $m_{val}$: Masks for values 
            \Statex $m_{pad}$: Padding Masks
        \State $\texttt{Initialize}~ w \leftarrow w_{in}, v \leftarrow v_{in}$
        \While{$w[-1] \neq \texttt{<eos>}$}
            \State \textbf{\# Prepare for next word/value prediction}
            \State $n \leftarrow \texttt{prepare\_inputs\_for\_generation}(w, v, m_{val}, m_{pad})$
            \State \textbf{\# Embed the inputs}
            \State $x_{input}[\neg m_v] \leftarrow$ \texttt{Embedding}($n_w$)
            \State $x_{input}[m_v] \leftarrow$ \texttt{Linear}($n_v$)
            \State \textbf{\# Pass through the AR Model}
            \State $h_{rep} \leftarrow$ \texttt{Transformer\_Decoder}($x_{input}, m_{pad}$)
            \State $w_{logits} \leftarrow$ \texttt{Linear}($h_{rep}[\neg m_{val}]$)
            \State \textbf{\# Compute the word/value}
            \State $scores \leftarrow$ \texttt{logits\_warper}(\texttt{logits\_processor}($w_{gt}, w_{logits[:-1]}), -1)$
            \State $y_w \leftarrow \texttt{multinomial}( \texttt{softmax}(scores))$
            \State \textbf{\# Use Diffloss Sampling Procedure in Alg\ref{alg:difflosssample}}
            \State $y_t \leftarrow$ \texttt{diff.\hyperref[alg:difflosssample]{sample}}($h_{rep}[:-1]$)
            \If{$n_{nxt}$ is a value}
                \State $v.\texttt{append}(y_t)$
            \Else
                \State $w.\texttt{append}(y_w)$
            \EndIf
        \EndWhile
        \State {\bfseries return} $w, v$
    \end{algorithmic}
\end{algorithm}

\begin{algorithm}[htbp]
    \caption{Diffloss Sampling Procedure}
    \label{alg:difflosssample}
    \begin{algorithmic}[1]
        \Require    
            \Statex $h_{rep}$: Hidden representations for conditional diffusion
            \Statex $T=200$: Number of sampling steps
        \State \textbf{\# Initialize noise, model kwargs and diffusion}
        \State $ y \leftarrow \texttt{randn}(h_{rep}.\texttt{shape}[0], \texttt{net}.in\_channels)$
        \State $model\_kwargs \leftarrow \texttt{dict}(c=h_{rep})$
        \State $\texttt{diff} \leftarrow$ \texttt{GaussianDiffusion(T)}
        \State $r \leftarrow \texttt{diff}.\texttt{p\_sample\_loop}(\texttt{net}, y, model\_kwargs) \textbf{ \# Predicted denoised sample}$
        \State {\bfseries return} $r$
    \end{algorithmic}
\end{algorithm}

\newpage

\setcounter{figure}{0}  
\renewcommand{\thefigure}{S\arabic{figure}}
\setcounter{table}{0}  
\renewcommand{\thetable}{S\arabic{table}} 

\section{More Details about Dataset Processing}

\subsection{Dictionary for Diverse Domains and Tasks}\label{app:dict}

In \ourM{}, we utilize a straightforward vocabulary dictionary for representing SMILES strings and material formulas, without any specialized design. Each atom type from the periodic table and special characters in SMILES, such as `C', `Cu', `(', `)', `=', are treated as individual tokens. This approach enables \ourM{} to accurately parse atoms and infer two-dimensional SMILES information through character-wise attention on special characters. Consequently, \ourM{} can be readily extended to other scientific domains by simply expanding the dictionary to include additional special characters, such as amino acid names for proteins and nucleic acid names for DNAs and RNAs.

The vocabulary dictionary also incorporates special tokens, which are crucial for \ourM{}'s flexibility as a next-token prediction model across diverse domains. These tokens serve as domain indicators, such as $<$bof$>$ for material formulas and $<$bos$>$ for molecule SMILES. Extending \ourM{} to other domains, like proteins and DNA, requires adding corresponding tokens, such as $<$boa$>$ and $<$bod$>$. This design empowers \ourM{} to handle various domains within a unified sequence representation. Thus, these domain-specific special tokens are integral to the vocabulary dictionary.

Furthermore, special tokens facilitate different tasks. For structure generation, $<$boc$>$ denotes coordinates. For conditional generation of sequences and structures, property tokens like $<$bulk$>$ for bulk modulus and $<$band$>$ for band gap are used. These tokens, followed by property values, precede the molecule sequence and coordinates. This allows \ourM{} to generate both sequences and structures conditioned on specific properties. Therefore, the special tokens in the vocabulary dictionary provide \ourM{} with flexibility across domains and tasks.

\subsection{Materials}

The pretraining dataset comprises structures from the Materials Project~\cite{jain2013commentary}, NOMAD~\cite{nomad}, and OQMD~\cite{oqmd2013,oqmd2015}, widely used open-source databases for materials. We eliminated duplicates in the merged dataset, filtered out data with chemical formulas failing SMACT validation, and removed structures with formulas matching those in the MP-20, Carbon-24, and MPTS-52 test sets used for evaluation. The final pretraining dataset contained approximately 5.5M samples.

We evaluated \ourM{} on three common material benchmark datasets: MP-20, Carbon-24, and MPTS-52. MP-20 consists of 45,231 stable inorganic materials from the Materials Project~\cite{jain2013commentary}, representing most experimentally synthesized materials with up to 20 atoms per unit cell. Carbon-24~\cite{Pickard2020AIRSSDF} includes 10,153 carbon materials, each with 6 to 24 atoms per unit cell. MPTS-52, an extension of MP-20, is a more challenging dataset comprising 40,476 structures with up to 52 atoms per unit cell, ordered by their earliest publication year. The dataset splits followed those described in \cite{xie2021crystal} and \cite{jiao2023crystal}.

To ensure data consistency, we sorted atoms and coordinates according to predefined rules. Coordinates were represented as fractional coordinates.

\begin{table}[htbp]
    \centering
    \renewcommand{\arraystretch}{1.2}
    \begin{tabular}{cccc}
        \toprule
        \textbf{} & \textbf{MP-20} & \textbf{MPTS-52} & \textbf{Carbon-24}\\
        \midrule
        \textbf{Train} & 27,136 & 32,380 & 6,091 \\
        \textbf{Valid} & 9,047 & \textendash & 2,032 \\
        \textbf{Test} & 9,046 & 8,096 & 2,030 \\
        \bottomrule
    \end{tabular}
    \caption{Size of Material Dataset}
    \label{tab:mat_data_size}
\end{table}

The parsed data format is as follows:
\begin{tcolorbox}
\texttt{<bos> [n * sites] <coord> [3 * lattice] [n * frac\_coords] <eos>}
\end{tcolorbox}

\subsection{Molecules Conformation Generation}
For the small molecule conformation generation task, we utilized the GEOM single-molecule multi-conformation dataset. Following DMCG, we partitioned the dataset into training, validation, and test sets with an 8:1:1 ratio, distributing conformations for each molecule into their respective sets to ensure no overlap. This resulted in 1,374,737, 165,204, and 174,162 conformation data points for QM9 in the training, validation, and test sets, respectively. For the larger GEOM-Drugs dataset, we selected 2,000,024, 100,104, and 100,106 conformation data points for training, test, and validation.

During data processing, we converted molecules to Canonical SMILES and renumbered atoms to align their indices with the SMILES order. Special SMILES characters were directly tokenized. In \ourM{}, we represent SMILES information as a sequence, rather than a graph, leveraging the transformer decoder's attention mechanism to capture all inherent information. To address potential mismatches between atom order in SMILES and coordinates, we used Canonical SMILES to align these orders. All information is then represented as a linear sequence input to the model, enabling \ourM{} to jointly train SMILES and coordinates, improving molecule understanding compared to sequence-to-structure-only training. The parsed data format is as follows:

\begin{tcolorbox}
\texttt{<bos> [cano\_smiles] <coord> [n * coords] <eos>}
\end{tcolorbox}

\subsection{Unification task}
We use the mixture of material pretraining dataset and QM9 mentioned above to pretrain \ourM. By adjusting the ratio of the two datasets, the material dataset loops twice when the small molecule dataset loops once. 

The parsed data format is as follows:
\begin{tcolorbox}
\texttt{<bos> <mat> [n * sites] <coord> [3 * lattice] [n * frac\_coords] <eos>}

\texttt{<bos> <mol> [cano\_smiles] <coord> [n * coords] <eos>}
\end{tcolorbox}

\subsection{Conditional Generation of Molecules}\label{app:cond}
For the Conditional Molecule Generation task, we adhered to two settings, as described in \cite{you2024latent} and \cite{gao2024tokenizing3dmoleculestructure}. Both settings utilize the QM9 dataset (approximately 137k samples), partitioning the training data into two halves: one for generator training and the other for classifier training. Consistent with \cite{you2024latent}, we employed a random seed of 42 for data splitting and utilized their classifier for testing, ensuring no overlap. Unlike \cite{gao2024tokenizing3dmoleculestructure}, we did not perform property normalization. Instead, property values were treated as numeric values and transformed into three-dimensional vectors, which were directly appended to the input sequence. Notably, we trained a unified model, using the same architecture for all properties, a departure from previous methods. Furthermore, unlike graph-based models, our approach does not require pre-sampling atom numbers, as the special token \texttt{<coord>} controls SMILES generation. Finally, our model employs a simple vocabulary dictionary, treating all SMILES characters, including special characters, as individual tokens (see Appendix \ref{app:dict} for details).


The parsed data format is as follows:
\begin{tcolorbox}
\texttt{<bos> <prop> [prop\_val] [cano\_smiles] <coord> [n * coords] <eos>}
\end{tcolorbox}

\subsection{Conditional Generation of Materials}
For conditional generation in the materials domain, we fine-tune a model that has been pretrained on large-scale materials data. The single-property datasets for bulk modulus and magnetization are taken directly from the \cite{zeni2025generative}. For multi-property conditional generation, we curate a new dataset by filtering materials from the Materials Project and computing their properties using \cite{yang2024mattersim}.

Each training sentence encodes a set of target properties through dedicated property tokens. Specifically, for a sample with $k$ conditioning attributes, the input sequence includes $k$ corresponding special tokens. The input sentence format is illustrated below:

\begin{tcolorbox}
\texttt{<prop1> [prop1\_val] <prop2> [prop2\_val] <prop3> [prop3\_val] <bos> [n * sites] <coord> [3 * lattice] [n * frac\_coords] <eos>}
\end{tcolorbox}

\subsection{Protein and Emulating MD equilibrium distributions}
To train our foundational protein model, we constructed a sequence of amino acid chains (represented by their single-letter abbreviations) for each protein. For the rare unconventional amino acids in the 'Emulating MD equilibrium distributions' section, we used the letter 'X' as a substitute. After generation, a proprietary post-processing procedure was applied to ensure their consistency and validity.

The full input sequence is organized as follows:

\begin{tcolorbox}
\texttt{<bos> [n * Amino\_acid] <coord> [n * C$\alpha$\_coords] <eos>}
\end{tcolorbox}

For the evaluation metrics, we first partitioned the 2D TICA projections into a grid of $num\_bins \times num\_bins$ along the x and y axes. We then counted the number of points falling into each bin, which can be interpreted as an unnormalized probability distribution $P(x,y)$. The free energy was subsequently computed using the Boltzmann relation:

\begin{equation}
    F(x,y)=-k_BT\cdot ln(P(x,y)+\epsilon).
\end{equation}

To improve visualization, we offset the minimum free energy to zero and truncated extremely high energy values. This procedure yielded a more interpretable representation of the Free Energy Landscape.

\subsection{Protein-Ligand Docking}
For the protein-ligand docking task, we adopt a similar strategy to that used in the small molecule generation setting. Each small molecule is represented using its canonical SMILES string, while protein pockets are described at the full-atom level, with atom order aligned to the residue indexing in the original MD.hdf5 files provided by \cite{siebenmorgen2024misato}. The corresponding coordinate tensors are arranged to follow this atom sequence exactly.

To enable a single model to handle both the ligand-only and joint pocket-ligand generation tasks, we design a unified input sequence structure that encodes all relevant components. The full input sequence is organized as follows:
\begin{tcolorbox}
\texttt{<bos> <bopo> [n * pocket\_atoms] <eopo> <boapc> [n * apo\_coords] <eoapc> <bom> [cano\_smiles] <eom> <bohpc> [n * holo\_coords] <eohpc> <bomc> [m * lig\_docked\_coords] <eomc> <eos>}
\end{tcolorbox}
During inference, task-specific tokens are used to control which parts of the structure the model is expected to generate. For the first task—ligand generation conditioned on both apo and holo pockets—the input sequence is provided up to the \texttt{<bomc>} token, allowing the model to autoregressively infer the subsequent ligand coordinates. For the second task—joint prediction of the holo pocket and ligand given only the apo pocket and SMILES—the input sequence is truncated at \texttt{<bohpc>}, prompting the model to predict both the holo pocket and downstream ligand coordinates.

Since the primary learning objective is to recover 3D coordinates, training is supervised using only the diffusion loss, applied to the predicted atomic positions. Following the design of 3D-MolFormer, we apply coordinate scaling as part of preprocessing. All 3D coordinates are scaled down by a factor of 5 to facilitate more stable and effective learning.

\subsection{EC Number-guided Protein Generation}

For training data, we used all protein sequences in UniProt that have an assigned EC number
and corresponding structural information available in the AlphaFold Database (AFDB), and with
sequence lengths below 512 residues. This results in a dataset of approximately 20 million samples.

The first three digits of the EC number are decomposed into three separate numbers. For example, "2.6.1.x" would be converted to 2 (ec number 1), 6 (ec number 2), and 1 (ec number 3).

\begin{tcolorbox}
\texttt{<bos> <ec1> [ec number 1] <ec2> [ec number 2] <ec3> [ec number 3]  [n * amino\_acids] <coord> [n * coords] <eos>}
\end{tcolorbox}

\subsection{Instruction Tuning with Natural Language Model}\label{app:instruct}

To train our model to generate crystal structures from natural language instructions, we constructed a dataset using the following format:

\begin{tcolorbox}
\texttt{<bos>\textbf{Instruction:} Generate a structure based on the provided composition , [n * sites]. \\
\textbf{Response:} [3 * lattice] [n * frac\_coords]<eos>}
\end{tcolorbox}

The \texttt{<bos>} and \texttt{<eos>} tokens delineated the start and end of the sequence, respectively, while the \textbf{Instruction:} and \textbf{Response:} labels clearly separated the input and output. By training on this structured data, the model learned to associate textual instructions with corresponding crystal structure representations, effectively bridging the gap between natural language and structural data.

\section{Computing Resource Usage}

Table~\ref{tab:gpu_usage} details the GPU computational resource usage for various tasks, including task category, dataset, model size, GPU specifications, and total computational cost measured in GPU days. All experiments were conducted using either NVIDIA A100 (80G) or AMD MI300X (192G) GPUs. Unless otherwise specified, training was performed with \underline{BF16} precision for computational efficiency. The reported GPU days range from 0.9 (Carbon-24) to 191 (Unified Pretraining).

It is worth noting that during fine-tuning phases, smaller batch sizes were often used due to limited dataset sizes or task-specific constraints. As a result, GPU utilization was suboptimal, and the actual effective GPU compute required may be significantly lower than the reported GPU days if fewer GPUs were used in parallel.

\begin{table}[htbp]
    \centering
    \renewcommand{\arraystretch}{1.2}
    \begin{tabular}{lcccc}
        \toprule
        \textbf{Task} & \textbf{Data} & \textbf{Model} & \textbf{GPU} & \textbf{Cost (GPU Days)} \\
        \midrule
        \multirow{4}{*}{\textbf{Material}} 
        & Pretrain & \ourM{}(400M) & MI300X (192G) & 109 \\
        & MP-20 & \ourM{}(400M) & A100 (80G) & 2.2 \\
        & MPTS-52 & \ourM{}(400M) & A100 (80G) & 2.6 \\
        & Carbon-24 & \ourM{}(400M) & A100 (80G) & 0.9 \\
        \midrule
        \multirow{2}{*}{\textbf{Molecule}} 
        & Large-QM9 & \ourM{}(400M) & A100 (80G) & 26.6 \\
        & Large-Drugs & \ourM{}(100M) & MI300X (192G) & 128 \\
        \midrule
        \textbf{Conditional Mol} 
        & QM9 & \ourM{}(100M) & A100 (80G) & 12 \\
        \midrule
        \multirow{3}{*}{\textbf{Conditional Mat}}
        & Mag & \ourM{}(400M) & A100 (80G) & \underline{55.8} \\
        & Bulk & \ourM{}(400M) & A100 (80G) & \underline{1.08} \\
        & Multiple & \ourM{}(400M) & MI300X (192G) & \underline{4.72} \\
        \midrule
        \textbf{Unified} 
        & Pretrain & \ourM{}(100M) & MI300X (192G) & 191 \\
        \midrule
        \multirow{2}{*}{\textbf{Protein}}
        & Pretrain & \ourM{}(400M) & MI300X (192G) & \underline{853} \\
        & MD & \ourM{}(400M) & MI300X (192G) & \underline{11} \\
        & EC Number & \ourM{}(400M) & MI300X (192G) & \underline{41} \\
        \midrule
        \textbf{Docking}
        & MISATO & \ourM{}(100M) & MI300X (192G) & \underline{67.7} \\
        \bottomrule
    \end{tabular}
    \caption{GPU Training Cost per Task. Models trained with \underline{BF16} precision are underlined.}
    \label{tab:gpu_usage}
\end{table}

\newpage

\section{Ablation Study}

To elucidate the contribution of each component within our model, we performed a series of ablation studies. By systematically removing or altering key components, we aimed to evaluate their individual and combined impact on overall model performance. The core components examined were the data parser, backbone, and diffloss, responsible for data preparation, word and condition generation, and structure generation, respectively. In our ablation studies, we selectively removed or modified these components to analyze their contribution to the final performance.

\subsection{Data Parser}

In this section, we investigate the impact of modeling the space group (sg) on the performance of the model. Sg train refers to whether the space group information is modeled during the training process, with the aim of evaluating its contribution to the model's prediction accuracy.

We kept the diff mul and res blocks parameters constant (at 16 and 12, respectively) and conducted a comparative experiment on whether to model the space group.

\begin{figure}[htbp]
    \centering
    \includegraphics[width=0.8\textwidth, keepaspectratio]{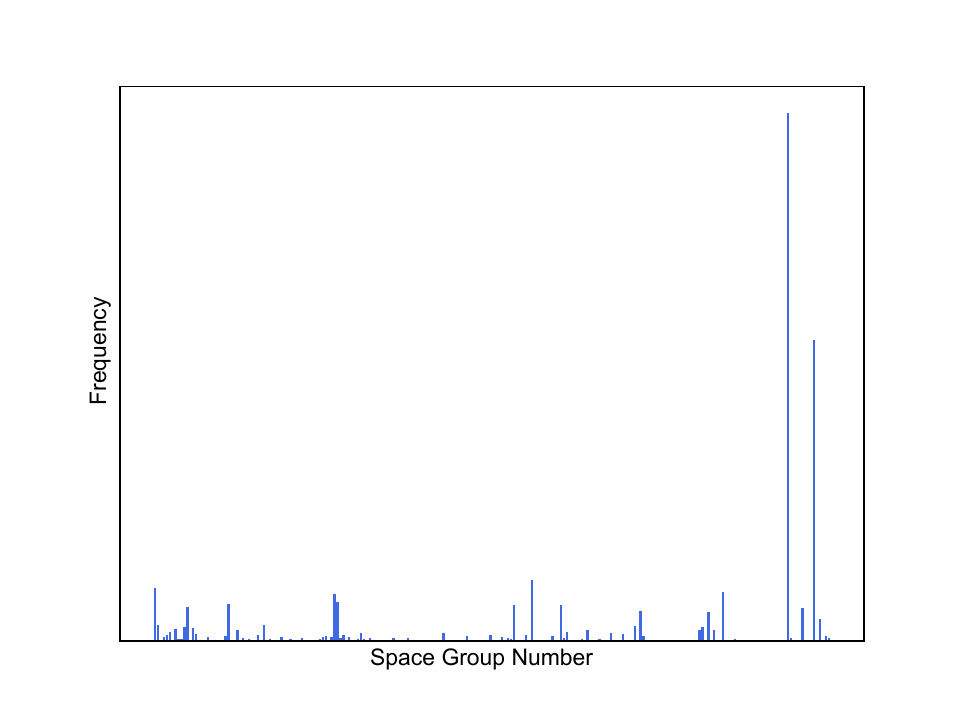} 
    \caption{The distribution of the frequency of the space group in training data is imbalanced.}
    \label{fig:sg_frq}
\end{figure}

The experimental results shown in~\ref{tab:sg_train_comparison} indicate that the model performs slightly better without modeling the space group than when the space group is modeled, due to the imbalanced space group distribution as shown in~\ref{fig:sg_frq}. This may suggest that in this particular task, the space group information does not significantly help to improve the match precision and may even introduce some unnecessary complexity, limiting further enhancement of model performance. 

Hence, we can conclude from this ablation study that although space group information might be beneficial for model performance in certain tasks, disabling sg allows the model to learn spatial features more freely in the current task, thereby improving overall performance. In subsequent ablation studies, unless specifically mentioned, the space group will not be modeled.

\begin{table}[htbp]
\centering
\begin{tabular}{|c|c|c|c|c|}
\hline
\textbf{Model} & \textbf{Diff mul} & \textbf{Res blocks} & \textbf{Sg} & \textbf{Match rate \%} \\ \hline
\ourM{}(100M)                & 16                & 12                  & +                 & 54.62          \\ \hline
\ourM{}(100M)                & 16                & 12                  & -                 & 56             \\ \hline
\end{tabular}
\caption{Comparison of Model Performance with and without sg}
\label{tab:sg_train_comparison}
\end{table}

\subsection{Model Size}
In this section, we conducted detailed experiments to investigate the impact of model size on the performance of the backbone. Specifically, we compared the performance of \ourM{}(100M) and \ourM{}(400M), and the experimental results are presented in~\ref{tab:llama_perf}. Meanwhile, different model size configurations are in~\ref{tab:llama_configurations}

In the experiments, the main differences between \ourM{}(100M) and \ourM{}(400M) lie in the number of hidden layers and the number of attention heads. The experimental results show that as the model size increased from 100M to 400M parameters, there was an improvement in model performance. Specifically, the Match score for \ourM{}(100M) was 57, while the Match score for \ourM{}(400M) increased to 59.35. This indicates that increasing the number of hidden layers (from 6 to 24) enhanced the model's feature representation capability and its ability to capture complex structures, under the same attention head configuration.

These experimental findings validate the advantages of larger models in complex tasks, particularly in terms of their ability to better extract and integrate information within multi-layered deep structures and rich contextual representations. However, it is important to note that while performance improvements are achieved, the increase in model size also brings higher computational costs. Therefore, in practical applications, balancing performance against computational expense is an important consideration.

The experiments consistently demonstrate that increasing model parameters, especially the number of hidden layers, has a positive impact on match precision.

\begin{table}[htbp]
\centering
\begin{tabular}{|c|c|c|c|c|}
\hline
\textbf{Model} & \textbf{Diff mul} & \textbf{Res blocks} & \textbf{Match rate \%} 
\\ \hline
\ourM{}(100M)                & 16                & 12                  & 57             
\\ \hline
\ourM{}(400M)                & 16                & 12                  & 59.35          
\\ \hline
\end{tabular}
\caption{Backbone parameters and performance.}
\label{tab:llama_perf}
\end{table}

\begin{table}[htbp]
\centering
\begin{tabular}{|c|c|c|}
\hline
\textbf{Model}         & \textbf{\ourM{}(100M) } & \textbf{\ourM{}(400M) } \\ \hline
hidden size                & 1024                & 1024                \\ \hline
intermediate size          & 4096                & 4096                \\ \hline
num hidden layers          & 6                   & 24                  \\ \hline
num attention heads        & 16                  & 16                  \\ \hline
num key value heads        & 16                  & 16                  \\ \hline
\end{tabular}
\caption{Comparison of \ourM{}(100M) and \ourM{}(400M) model configurations}
\label{tab:llama_configurations}
\end{table}

\subsection{Diffloss}

\subsubsection{Diffmul \& Resblocks}
In the ablation study of the diff module, we experimented with two key parameters in the module, the diffmul and the resblocks, to assess their impact on model performance.

The diffmul, short for diffusion multiplier, represents the number of time points selected along the path where the model pushes the noise distribution forward to the data distribution. It controls the model's ability to extract features along the diffusion path. We chose two different values, 4 and 16, to observe the impact of smaller and larger feature extraction capabilities on the final results. As shown in \ref{tab:ablation_study_diffres}, it is obvious that the model performs better with a diffmul value of 16 compared to a value of 4. For example, with res blocks fixed at 3, increasing the diffmul from 4 to 16 improved the Match score from 53.31 to 56.06. Similarly, with res blocks at 12, the Match score for a diffmul of 16 (54.62) was higher than that for a diffmul of 4 (48.98). This suggests that a greater feature extraction capability performs better in handling complex structures.

Resblocks refer to the number of residual blocks, which are used to enhance the model's expressive power. We compared the performance between models with 3 and 12 Resblocks, observing mixed effects on performance. With space group modeling, a model size of 100M, and a Diffmul of 4, increasing the number of Resblocks from 3 to 12 led to a decrease in Match Rate from 53.31\% to 48.98\%. However, with a Diffmul of 16, the performance difference between 3 and 12 Resblocks was marginal (56.06\% vs. 54.62\%). On the other hand, when increasing the model size from 100M to 400M and not modeling the space group, increasing Resblocks from 3 to 12 improved performance from 54.78\% to 59.35\%. These results suggest that while increasing the number of Resblocks helps the model handle more complex input data, the performance gains may plateau in certain conditions.

In summary, the experimental results demonstrate that increasing the values of diffmul and resblocks significantly contributes to enhancing the overall performance of the model.

\begin{table}[htbp]
\centering
\begin{tabular}{|c|c|c|c|c|}
\hline
\textbf{Model} & \textbf{Diffmul} & \textbf{Resblocks} & \textbf{Sg} & \textbf{Match rate \%} \\ \hline
\ourM{}(100M)                & 4                 & 3                   & +           & 53.31             \\ \hline
\ourM{}(100M)                & 16                & 3                   & +           & 56.06             \\ \hline
\ourM{}(100M)                & 4                 & 12                  & +           & 48.98             \\ \hline
\ourM{}(100M)                & 16                & 12                  & +           & 54.62             \\ \hline
\ourM{}(400M)                & 16                & 3                   & -           & 54.78             \\ \hline
\ourM{}(400M)                & 16                & 12                  & -           & 59.35             \\ \hline
\end{tabular}
\caption{Experimental Results of Diffloss Module Configurations}
\label{tab:ablation_study_diffres}
\end{table}

\subsubsection{Scheduler}\label{sec:edm}
In this section, we explore the impact of two schedulers on the effectiveness of the model. The two schedulers are the Diffusion scheduler used in this paper and the EDM scheduler, which was first introduced in \cite{karras2022elucidating}.

The EDM scheduler utilizes a second-order Heun ODE solver for improved sampling accuracy and efficiency. The time step scaling follows a specific schedule based on $(\sigma_{max}^{\frac{1}{\rho}} + \frac{1}{N-i}(\sigma_{\text{min}}^{\frac{1}{\rho}} - \sigma_{\text{max}}^{\frac{1}{\rho}}))^{\frac{1}{\rho}}$, allowing for smooth transitions during the diffusion process. EDM employs a flexible network architecture, enabling compatibility with various models. Key preconditioning techniques include skip scaling, defined as $\sigma \cdot \sigma_{\text{data}} / (\sigma_{\text{data}}^2 + \sigma^2)$, along with tailored input and output scaling formulas that optimize noise and signal propagation. The noise distribution is modeled as $\ln(\sigma) \sim \mathcal{N}(P_{\text{mean}}, P_{\text{std}}^2)$. Loss weighting is also balanced with the formula $(\sigma_{\text{data}}^2 / (\sigma_{\text{data}}^2 + \sigma^2))^2$. More details can be found in \cite{karras2022elucidating}.

\paragraph{EDM parameters}

\begin{figure}[htbp]
    \centering
    \includegraphics[width=1\textwidth, keepaspectratio]{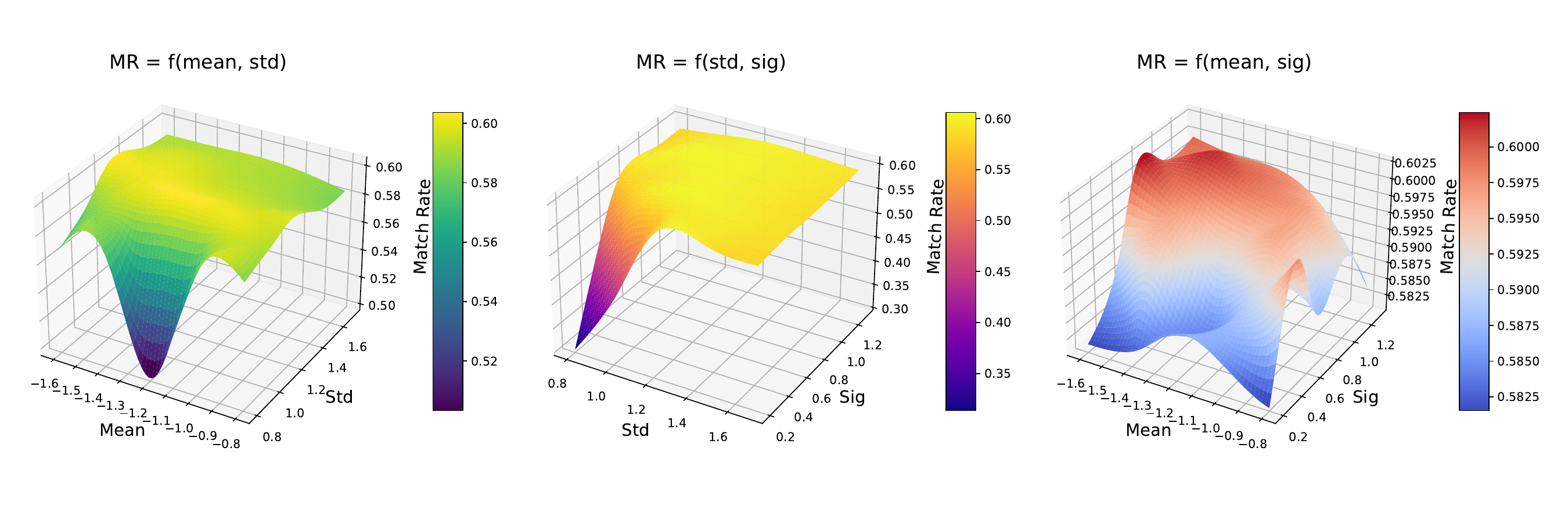} 
    \caption{Comparison between mean, std and sigma in EDM}
    \label{fig:edmparam}
\end{figure}

We provide three images showing the Match rate under different \texttt{std}, \texttt{sig}, and \texttt{mean} values in \ref{fig:edmparam}. The default model setting is \ourM{}(100M) with 12 resblocks and 16 diffmul. Here is a summary of the main trends:

\subparagraph{Impact of the mean value:}

\begin{itemize}
  \item The match rate tends to be higher, typically maintaining above 0.58 to 0.6, when the mean value fluctuates between -1.4 and -0.8. A relatively lower mean value seems to correlate with a higher match rate.
  \item When the mean value decreases to -1.6, the match rate sometimes drops slightly but the overall performance remains relatively stable.
  \item Larger mean values (e.g., -0.8) usually perform better in terms of match rate, and as the mean value becomes smaller (e.g., -1.6 or lower), the match rate may gradually decrease.
\end{itemize}

\subparagraph{Impact of std (standard deviation):}

\begin{itemize}
  \item The match rate usually remains stable when std fluctuates between 1 and 1.7. The smaller the std value (e.g., 1), the match rate tends to be slightly higher.
  \item When the std variation is small (e.g., from 1 to 1.5), fluctuations in the match rate are not significant, suggesting that std has a minimal impact on the match rate within this range.
  \item As std further increases to 1.7, the match rate begins to decline slightly, but not significantly.
\end{itemize}

\subparagraph{Impact of sigma value:}

\begin{itemize}
  \item The closer the sig value is to 1, the better the match rate typically performs, often fluctuating between 0.58 and 0.6 when the value is 1.
  \item When the sig value is 1.3, the match rate sometimes decreases slightly but still maintains a relatively high level.
  \item Lower sig values (e.g., 0.2 or 0.5) are usually accompanied by lower match rates, indicating that lower sig values may inhibit the model's matching ability.
\end{itemize}

\subparagraph{Multifactorial influence:}

\begin{itemize}
  \item For the combination of mean = -1.4, std = 1.2, and sig = 1, the match rate can reach a peak of 0.6024.
  \item As std and sig increase to higher values, the match rate typically declines, e.g., when sig increases to 1.5, there is a slight decrease in the match rate.
  \item A combination of particularly low sig and larger mean (e.g., mean = -1.6, sig = 0.2) results in a sharp decline in the match rate.
\end{itemize}

In summary, the data suggest that moderate mean and std values combined with a higher sig value can enhance the match rate, while more extreme combinations of mean and sig tend to lower the match rate.

\paragraph{DDPM vs. EDM}

We compare Diffloss and EDMloss in \ref{tab:model_comparison_edmaug}, and the analysis of the table is as follows.

Despite the fact that the EDM scheduler slightly outperforms the Diffusion loss in its best results, the values of key parameters such as \texttt{sig}, \texttt{mean}, and \texttt{aug} for conditional diffusion are challenging to determine accurately. This introduces a large number of parameter combinations in EDM, requiring extensive hyperparameter tuning. Moreover, after incorporating data augmentation, we observed that the performance gap between EDM and DDPM training strategy becomes minimal. In this work, we mainly focus on the combination of the auto-regressive model and the diffusion head in a unified generation across different tasks and different domains. Hence, here we adopt the DDPM training strategy for its simplicity, requiring fewer hyperparameters to be tuned. To achieve better performance, an investigation of combining different setups of other diffusion strategies or flow matching models is also supported but not included in this work. 

\begin{table}[htbp]
\centering
\begin{tabular}{|c|c|c|c|c|c|}
\hline
Model & Diffmul & Resblocks & Scheduler & Augmentation & MR \% \\ 
\hline
\ourM{}(400M)        & 16       & 12         & DDPM      & -            & 59.35 \\ 
\hline
\ourM{}(400M)        & 16       & 12         & EDM       & -            & 59.08 \\ 
\hline
\ourM{}(400M)        & 16       & 12         & DDPM      & +            & 63.88 \\ 
\hline
\ourM{}(400M)        & 16       & 12         & EDM       & +            & 63.00 \\ 
\hline
\end{tabular}
\caption{Ablation study on Scheduler and Augmentation.}
\label{tab:model_comparison_edmaug}
\end{table}

\subsection{Data Augmentation}

In this section, we investigate the impact of data augmentation on model performance. The augmentation techniques employed include unit cell translation and rotation. For unit cell translation, a random three-dimensional vector is added to the fractional coordinates of each atom, and the result is taken modulo 1 to maintain periodicity. Conversely, unit cell rotation involves randomly rotating the lattice coordinate system while keeping the atomic coordinates fixed. Algorithm \ref{alg:augmentation} presents the pseudocode for material data augmentation.

\begin{algorithm}[htbp]
   \caption{Rotation \& Translation Augmentation}
   \label{alg:augmentation}
    \begin{algorithmic}[1]
        \Require    
            \Statex $l$: Lattice
            \Statex $x$: Fractional Coordinates
        \State \textbf{\# Apply Rotation}
        \State $R \leftarrow \texttt{UniformRandomRotation()}$
        \State $l \leftarrow R \cdot l$
        \State \textbf{\# Apply Translation}
        \State $t \sim \mathcal{N}(\vec{0}, \bf{I}_3)$
        \State $x \leftarrow (x + t) \mod 1$
        \State{\bfseries return} $l, x$
    \end{algorithmic}
\end{algorithm}

As shown in Table \ref{tab:model_comparison_edmaug}, incorporating augmentation significantly enhances performance for both schedulers, resulting in an approximate 4-point increase in match score.

\subsection{Pretraining or Training from Scratch}

In this section, we present an ablation study comparing the impact of pretraining versus training from scratch across three datasets: MP-20, Carbon-24, and MPTS-52. Table~\ref{tab:csp_ablation} summarizes the results, detailing Match Rate (MR) and Root Mean Square Deviation (RMSD).

The pretrained base model exhibits moderate performance across the datasets, achieving a Match Rate (MR) of 60.48\% on MP-20, 3.15\% on Carbon-24, and 33.61\% on MPTS-52, with corresponding RMSD values of 0.0568, 0.1928, and 0.1086, respectively.

Training the model from scratch yields a slight improvement in MR for MP-20, reaching 63.88\%, but demonstrates mixed results on Carbon-24 and MPTS-52, with MR decreasing to 27.09\% and 29.09\%, respectively.

Conversely, finetuning the pretrained model on each dataset results in superior performance, with substantial enhancements in MR across all three datasets. On MP-20, the finetuned model achieves an MR of 67.01\%. Similarly, on Carbon-24 and MPTS-52, the finetuned model achieves MR values of 30.05\% and 38.65\%.

These findings suggest that pretraining followed by dataset-specific finetuning is a more effective strategy than training from scratch, particularly for complex datasets like MP-20 and MPTS-52. Notably, even when trained from scratch, \ourM{} outperforms FlowMM on these three benchmarks, demonstrating the inherent advantages of \ourM{}.

\begin{table}[htbp]
    \centering
    \begin{tabular}{lllllll}
\hline & \multicolumn{2}{c}{ MP-20 } & \multicolumn{2}{c}{ Carbon-24 } & \multicolumn{2}{c}{ MPTS-52 } \\
\hline Methods & MR(\%)$\uparrow$ & RMSD $\downarrow$ & MR(\%)$\uparrow$ & RMSD $\downarrow$ & MR(\%)$\uparrow$ & RMSD $\downarrow$ \\ 
\hline Base Model & 60.48 & 0.0568 & 3.15 & $\mathbf{0.1928}$ & 33.61 & 0.1086 \\
\hline Train from Scratch & 63.88 & 0.0598 & 27.09 & 0.2264 & 29.09 & 0.1256 \\
\hline Finetune on Each & $\mathbf{67.01}$ & $\mathbf{0.037}$ & $\mathbf{30.05}$ & 0.2286 & $\mathbf{38.65}$ & $\mathbf{0.0657}$ \\
\hline
\end{tabular}
    \caption{Pretraining or Training from Scratch.}
    \label{tab:csp_ablation}
\end{table}

\newpage
\section{Special Tokens Used in Sequence Modeling}\label{app:sp_tok}

In our model framework, we introduce a set of special tokens to indicate the beginning of different modalities. These tokens serve as explicit prompts for the model to distinguish between different input types, such as molecular SMILES, protein sequences, and geometric structures. Table~\ref{tab:special_tokens} provides a summary of the special tokens used in our vocabulary.

\begin{table}[htbp]
\centering
\begin{tabular}{ll}
\toprule
\textbf{Token} & \textbf{Description} \\
\midrule
\texttt{<bos>} & Beginning of sites/sentence/SMILES \\
\texttt{<eom>} & End of sentence \\
\texttt{<bof>} & Beginning of material formula \\
\texttt{<bol>} & Beginning of material lattice \\
\texttt{<bop>} & Beginning of protein amino-acid sequence \\
\texttt{<boc>} & Beginning of coordinates \\
\texttt{<bulk>} & Beginning of the property bulk modulus\\
\texttt{<band>} & Beginning of the property band gap \\
\texttt{<mag>} & Beginning of the property magnetic \\
\texttt{<heat\_capacity>} & Beginning of the property heat capacity\\
\texttt{<E\_hill>} & Beginning of the property energy above the convex hull \\
\texttt{<density>} & Beginning of the property density \\
\texttt{<bopo>} & Beginning of pocket \\
\texttt{<eopo>} & End of pocket \\
\texttt{<boapc>} & Beginning of pocket apo coords \\
\texttt{<eoapc>} & End of pocket apo coords \\
\texttt{<bom>} & Beginning of molecule SMILES sequence \\
\texttt{<eom>} & End of molecule SMILES sequence \\
\texttt{<bohpc>} & Beginning of pocket holo coords \\
\texttt{<eohpc>} & End of pocket holo coords \\
\texttt{<bomc>} & Beginning of molecule coords \\
\texttt{<eomc>} & End of the molecule coords \\
\texttt{<ec1>} & Beginning of the EC Number 1 \\
\texttt{<ec2>} & Beginning of the EC Number 2 \\
\texttt{<ec3>} & Beginning of the EC Number 3 \\
\bottomrule
\end{tabular}
\caption{Explanation of special tokens used in the unified vocabulary.}
\label{tab:special_tokens}
\end{table}


\end{document}